\documentclass[journal]{IEEEtran}
\usepackage{amsmath,amssymb,amsfonts}
\usepackage{epsfig,graphicx,float}
\usepackage[caption=false]{subfig}
\usepackage{booktabs,multirow,array} 

\usepackage[misc]{ifsym}
\usepackage{hyperref}

\newcommand{\raggedjustifyalign}{\leftskip=0pt \rightskip=0pt plus 0cm}

\usepackage{color}
\usepackage{ulem}
\newcommand{\rev}[1]{{ #1}}

\ifCLASSINFOpdf
\else
\fi
\begin{document}

%
\title{Empowering Things with Intelligence: A Survey of the Progress, Challenges, and Opportunities in Artificial Intelligence of Things}
%
%
%

\author{Jing Zhang,~\IEEEmembership{Member,~IEEE,}
        and Dacheng Tao,~\IEEEmembership{Fellow,~IEEE}
\thanks{This work was supported by the Australian Research Council Projects FL-170100117, DP-180103424, IH-180100002.}
\thanks{J. Zhang and D. Tao are with the School of Computer Science, in the Faculty of Engineering, at The University of Sydney, 6 Cleveland St, Darlington, NSW 2008, Australia (email: \{jing.zhang1; dacheng.tao\}@sydney.edu.au).}
\thanks{Copyright (c) 20xx IEEE. Personal use of this material is permitted. However, permission to use this material for any other purposes must be obtained from the IEEE by sending a request to pubs-permissions@ieee.org.}
}

%
%

\markboth{IEEE INTERNET OF THINGS JOURNAL,~Vol.~XX, No.~X, November~2020}%
{Shell \MakeLowercase{\textit{et al.}}: Bare Demo of IEEEtran.cls for IEEE Journals}
%



\maketitle

\begin{abstract}
In the Internet of Things (IoT) era, billions of sensors and devices collect and process data from the environment, transmit them to cloud centers, and receive feedback via the internet for connectivity and perception. However, transmitting massive amounts of heterogeneous data, perceiving complex environments from these data, and then making smart decisions in a timely manner are difficult. Artificial intelligence (AI), especially deep learning, is now a proven success in various areas including computer vision, speech recognition, and natural language processing. AI introduced into the IoT heralds the era of artificial intelligence of things (AIoT). This paper presents a comprehensive survey on AIoT to show how AI can empower the IoT to make it faster, smarter, greener, and safer. Specifically, we briefly present the AIoT architecture in the context of cloud computing, fog computing, and edge computing. Then, we present progress in AI research for IoT from four perspectives: perceiving, learning, reasoning, and behaving. Next, we summarize some promising applications of AIoT that are likely to profoundly reshape our world. Finally, we highlight the challenges facing AIoT and some potential research opportunities.
\end{abstract}


%
\IEEEpeerreviewmaketitle

\section{Introduction}
%
%
%
%
\IEEEPARstart{T}{he} Internet of Things (IoT), a term originally coined by Kevin Ashton at MIT's Auto-ID Center \cite{ashton2009internet}, refers to a global intelligent network that enables cyber-physical interactions by connecting numerous things with the capacity to perceive, compute, execute, and communicate with the internet; process and exchange information between things, data centers, and users; and deliver various smart services \cite{ccsa2011,itut2012}. From the radio-frequency identification (RFID) devices developed in the late 1990s to modern smart things including cameras, lights, bicycles, electricity meters, and wearable devices, the IoT has developed rapidly over the last twenty years in parallel with advances in networking technologies including Bluetooth, Wi-Fi, and long-term evolution (LTE). The IoT represents a key infrastructure for supporting various applications \cite{da2014internet}, e.g., smart homes \cite{li2011smart,stojkoska2017review}, smart transportation \cite{contreras2017internet,aslam2020internet}, smart grids \cite{guan2017achieving}, and smart healthcare \cite{catarinucci2015iot,rahmani2018exploiting}. According to McKinsey's report \cite{lamarre2019ten}, the IoT sector will contribute \$2.7 to \$6.2 trillion to the global economy by 2025.

A typical IoT architecture has three layers \cite{lin2017survey}: a perception layer, a network layer, and an application layer. The perception layer lies at the bottom of the IoT architecture and consists of various sensors, actuators, and devices that function to collect data and transmit them to the upper layers. The network layer lies at the center of the IoT architecture and comprises different networks (e.g., local area networks (LANs), cellular networks, the internet) and devices (e.g., hubs, routers, gateways) enabled by various communication technologies such as Bluetooth, Wi-Fi, LTE, and fifth-generation mobile networks (5G). The application layer is the top IoT layer and it is powered by cloud computing platforms, offering customized services to users, e.g., data storage and analysis. In conventional IoT solutions, data collected from sensors are transmitted to the cloud computing platform through the networks for further processing and analysis before delivering the results/commands to end devices/actuators.

However, this centralized architecture faces significant challenges in the context of the massive numbers of sensors used across various applications. Based on reports from Cisco \cite{macaulay2015internet} and IDC \cite{shirer2019growth}, 50 billion devices will be IoT connected by 2025, generating 79.4 zettabytes of data. Transmitting this huge amount of data requires massive bandwidth, and cloud processing and sending the results back to end devices leads to high latency. To address this issue, ``fog computing'', coined by Cisco \cite{bonomi2011connected}, aims to bring storage, computation, and networking capacity to the edge of the network (e.g., to distributed fog nodes such as routers) in proximity to the devices. Fog computing offers the advantages of low latency and high computational capacity for IoT applications \cite{chiang2016fog,ni2017securing}. ``Edge computing'' has also recently been proposed by further deploying computing capacity on edge devices in proximity to sensors and actuators \cite{abbas2017mobile,pan2017future}. Note that the terms fog computing and edge computing are interchangeable in some literature \cite{shi2016edge,abbas2017mobile} or the fog is treated as a part of the broader concept of edge computing \cite{tordera2016fog}. For clarity, here we treat them as different concepts, i.e., fog computing at the network side and edge computing at the thing side. Edge computing can process and analyze data on premises and make decisions instantly, thereby benefitting latency-sensitive IoT applications. The processed data from different devices can then be aggregated at the fog node or cloud center for further analysis to enable various services.

\begin{figure}
  \centering
  \includegraphics[width=\linewidth]{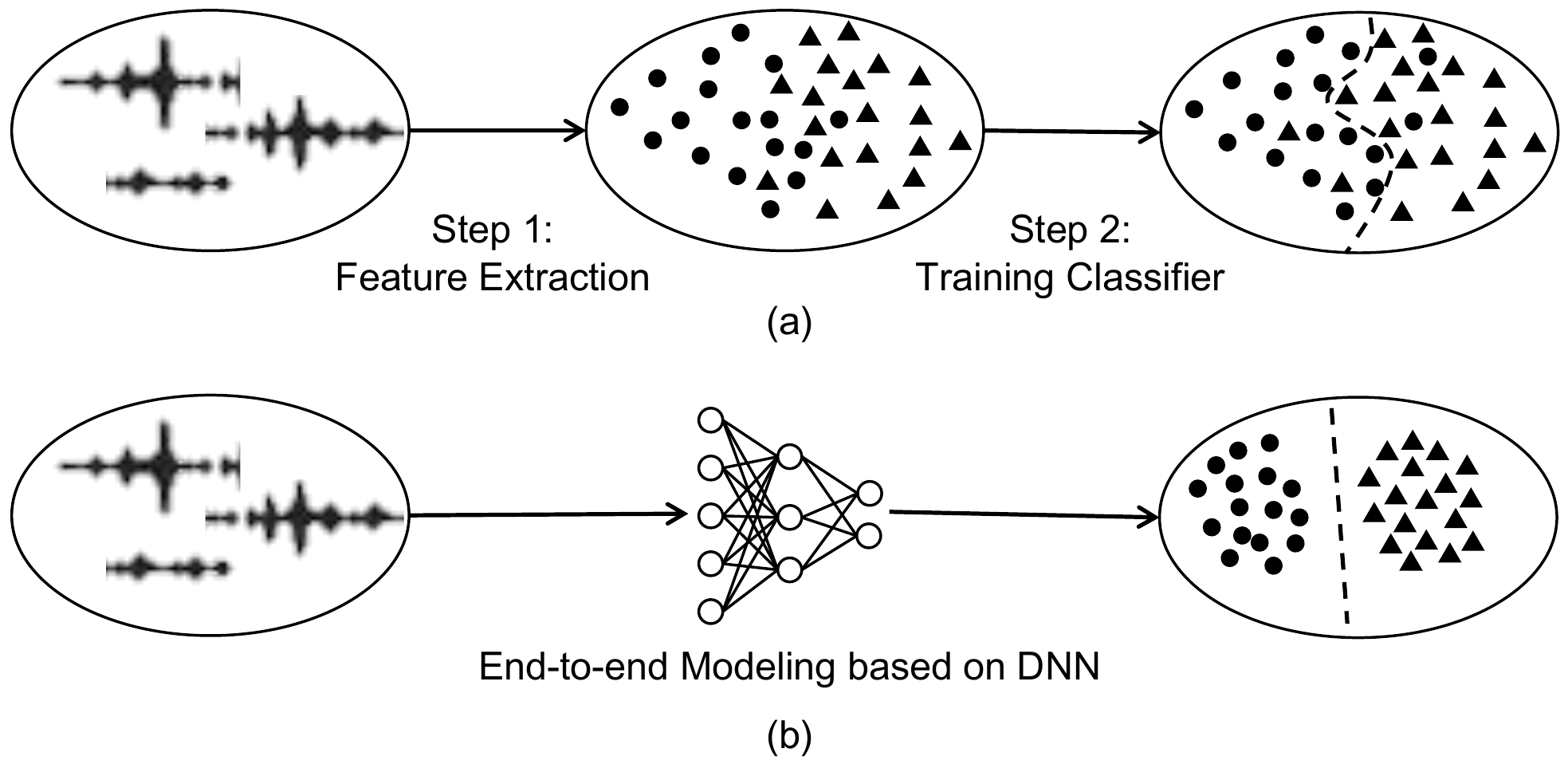}
  \caption{The schematic paradigm of (a) classical machine learning methods and (b) deep learning.}
  \label{fig:learningparadigm}
\end{figure}

In addition to these challenges created by massive numbers of sensors, another challenge arises through their heterogeneous nature \cite{chen2020internet} including scalar sensors, vector sensors, and multimedia sensors as summarized in Table~\ref{tab:sensor}. Perceiving and understanding dynamic and complex environments from sensor data is fundamental to IoT applications providing useful services to users. As a result, various intelligent algorithms have been proposed for certain applications with scalar and vector sensors, e.g., decision rules-based methods and data-driven methods. Typically, these methods use handcrafted features extracted from data for further prediction, classification, or decision (Figure~\ref{fig:learningparadigm}(a)). However, this paradigm of using handcrafted features and shallow models is unsuited to modern IoT applications with multimedia sensors. First, multimedia sensor data are high-dimensional and unstructured (semantics are unavailable without additional processing), so it is difficult to design handcrafted features for them without domain knowledge. Second, handcrafted features are usually vulnerable to noise and different types of variance (e.g., illumination, viewpoint) in data, limiting their representation and discrimination capacity. Third, feature design and model training are separate, without joint optimization.

\begin{table}[htbp]
  \centering
  \caption{Summary of exemplar AIoT sensors. A: Agriculture, C: Cities/Homes/Buildings, E: Education, G: Grids, H: Healthcare, I: Industry, S: Security, T: Transportation.}
    \begin{tabular}{p{1.4cm}<{\raggedjustifyalign}p{2.8cm}<{\raggedjustifyalign}p{1.2cm}<{\raggedjustifyalign}p{1.75cm}<{\raggedjustifyalign}}
    \toprule
     Sensor Type & Scalar & Vector & Multimedia\\
     \midrule
     Sensors & altimeter, ammeter, hygrometer, light meter, manometers, ohmmeter, tachometer, thermometer, voltmeter, wattmeter & anemometer, accelero-meter, gyroscope & microphone, camera, lidar, CT/MRI/ultra-sound scanner\\
     \midrule
     Data Type & scalar & vector &  2/3/4D tensor\\
     \midrule
     Applications & A,C,E,G,H,I,T  & A,C,G,I,T & A,C,E,G,H,I,S,T\\
    \bottomrule
    \end{tabular}%
  \label{tab:sensor}%
\end{table}%

The last few years has witnessed a renaissance in artificial intelligence (AI) assisted by deep learning. Deep neural networks (DNNs) have been widely used in many areas and have achieved excellent performance in many applications including speech recognition \cite{sak2014long}, face recognition \cite{taigman2014deepface}, image classification \cite{krizhevsky2012imagenet}, object detection \cite{ren2015faster}, semantic segmentation \cite{chen2017deeplab}, natural language processing \cite{devlin2019bert}, benefitting from their powerful capacity to feature learn and end-to-end model (Figure~\ref{fig:learningparadigm}(b)). Moreover, with modern computational devices, e.g., graphics processing units (GPUs) and tensor processing units (TPUs), DNNs can efficiently and automatically discover discriminative feature representations from large-scale labeled or unlabeled datasets in a supervised or unsupervised manner \cite{deng2009imagenet}. Deploying DNNs into cloud platforms, fog nodes, and edge devices in IoT systems enables the construction of an intelligent hybrid computing architecture capable of leveraging the power of deep learning to process massive quantities of data and extract structured semantic information with low latency. Therefore, advances in deep learning have paved a clear way for improving the perceiving ability of IoT systems with large numbers of heterogeneous sensors.

Although an IoT's perception system is a critical component of the architecture, simply adapting to and interacting with the dynamic and complex world is insufficient. For example, edge cases exist in the real world that may not be seen in the training set nor defined in the label set, resulting in degeneration of a pre-trained model. Another example is in industry, where the operating modes of machines may drift or change due to fatigue or wear and tear. Consequently, models trained for the initial mode cannot adapt to this variation, leading to a performance loss. These issues are related to some well-known machine learning research topics including few-shot learning \cite{wang2020generalizing}, zero-shot learning \cite{wang2019survey}, 
meta-learning \cite{finn2017model}, unsupervised learning \cite{chen2020simple}, semi-supervised learning \cite{xie2020self}, transfer learning \cite{zamir2018taskonomy}, and domain adaptation \cite{zhang2019category,yang2020mobileda}. Deep learning has facilitated progress in these areas, suggesting that deep learning can be similarly leveraged to improve IoT system learning. Furthermore, to interact with the environment and humans, an IoT system should be able to reason and behave. For example, a man parks his car in a parking lot every morning and leaves regularly on these days. Therefore, a smart parking system may infer that he probably works nearby. Then, it can recommend and introduce some parking offers, car maintenance, and nearby restaurants to him via an AI chatbot. These application scenarios could benefit from recent advances in causal inference and discovery \cite{zhang2019multi}, graph-based reasoning \cite{wang2019explainable}, reinforcement learning \cite{zhang2019hierarchical}, and speech recognition and synthesis \cite{sak2014long,arik2017deep}.

According to Cisco's white paper \cite{bradley2013embracing}, 99.4\% of physical objects are still unconnected. Advanced communication technologies such as Wi-Fi 6 (IEEE 802.11ax standard) and 5G and AI technologies will enable mass connection. This heralds the era of the artificial intelligence of things (AIoT), where AI encounters IoT. Both academia and industry have invested heavily in AIoT, and various AIoT applications have now been developed, providing services and creating value. Therefore, here we performed a survey of this emerging area to demonstrate how AI technologies empower things with intelligence and enhance applications.

\subsection{Contributions of this Survey}
There are several excellent existing surveys on IoT covering different perspectives, a detailed discussion and comparison of which is provided below. Here we specifically focus on AIoT and provide an overview of research advances, potential challenges, and future research directions through a comprehensive literature review and detailed discussion. The contributions of this survey can be summarized as follows:

$\bullet$ We discuss AIoT system architecture in the context of cloud computing, fog computing, and edge computing.

$\bullet$ We present progress in AI research for IoT, applying a new taxonomy: perceiving, learning, reasoning, and behaving.

$\bullet$ We summarize some promising applications of AIoT and discuss enabling AI technologies.

$\bullet$ We highlight challenges in AIoT and some potential research opportunities.

\begin{figure*}
  \centering
  \includegraphics[width=\linewidth]{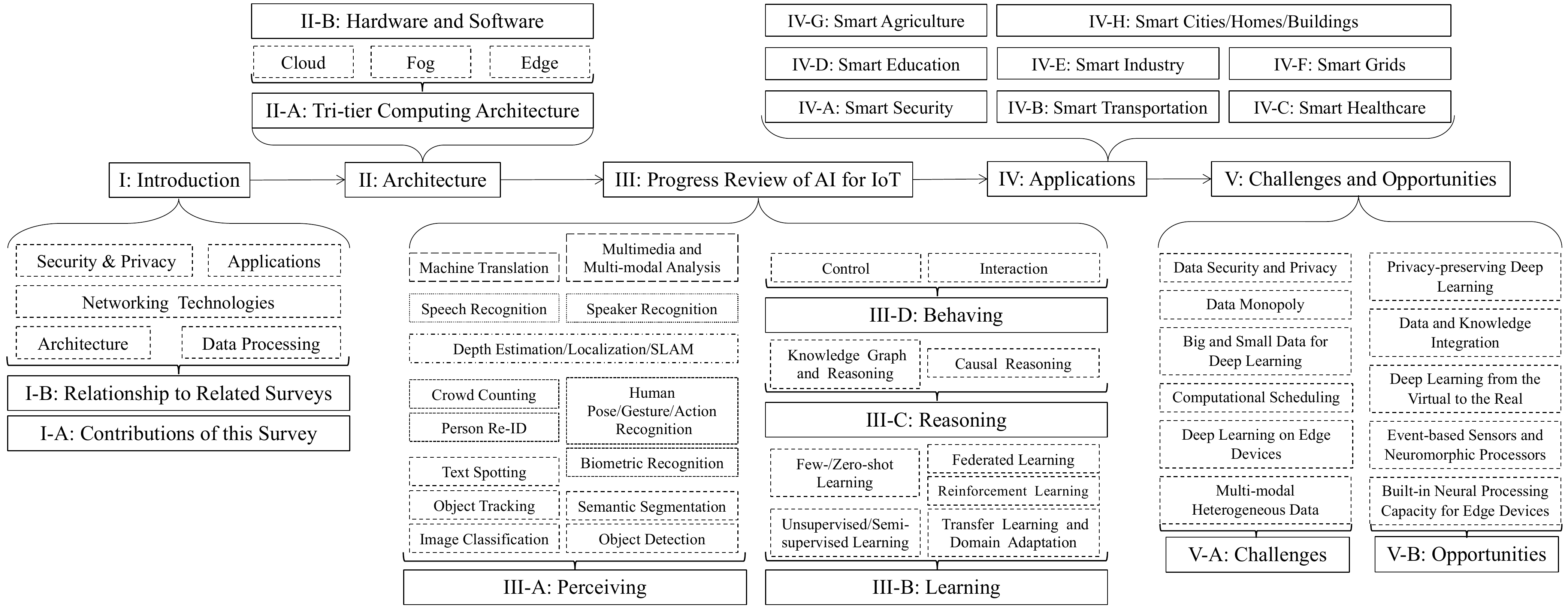}
  \caption{\rev{Diagram of the organization of this paper.}}
  \label{fig:organization}
\end{figure*}

\subsection{Relationship to Related Surveys}
We first review existing surveys related to IoT and contrast them with our work. Since the IoT is related to many topics such as computing architectures, networking technologies, applications, security, and privacy, surveys have tended to focus on one or some of these topics. For example, Atzori et al. \cite{atzori2010internet} described the IoT paradigm from three perspectives: ``things''-oriented, ``internet''-oriented, and ``semantic''-oriented, corresponding to sensors and devices, networks, and data processing and analysis, respectively. They reviewed enabling technologies and IoT applications in different domains and also analyzed some remaining challenges with respect to security and privacy. In \cite{whitmore2015internet}, Whitmore et al. presented a comprehensive survey on IoT and identified recent trends and challenges. We review the other surveys according to the specific topic covered.

\subsubsection{Architecture}
In \cite{yaqoob2017internet}, several typical IoT architectures were reviewed including software-defined network-based architectures, the MobilityFirst architecture, and the CloudThings architecture. They argued that future IoT architectures should be scalable, flexible, interoperable, energy efficient, and secure, such that the IoT system can integrate and handle huge numbers of connected devices. \cite{lin2017survey} discussed two typical architectures: the three-layer architecture (i.e., with a perception layer, network layer, and application layer) and the service-oriented architecture. For the IoT computing architecture, integrating cloud computing \cite{gubbi2013internet} with fog/edge computing \cite{lin2017survey} has attracted increasing attention. \cite{chiang2016fog,pan2017future} provided a detailed review of fog computing and edge computing for IoT. Since we focus on AI-empowered IoT, we are also interested in the cloud/fog/edge computing architectures of IoT systems, especially those tailored for deep learning. More detail will be presented in Section~\ref{sec:architecture}.

\subsubsection{Networking Technologies}
Connecting massive numbers of things to data centers and transmitting data at scale relies on various networking technologies. In \cite{verma2017survey}, Verma et al. presented a comprehensive survey of network methodologies including data center networks, hyper-converged networks, massively-parallel mining networks, and edge analytics networks, which support real-time analytics of massive IoT data. Wireless sensor networks have also been widely used in IoT to monitor physical or environmental conditions \cite{mainetti2011evolution}. The recently developed 5G mobile networks can provide very high data rates at extremely low latency and a manifold increase in base station capacity. 5G is expected to boost the number of connected things and drive the growth of IoT applications \cite{chettri2019comprehensive}. Due to the massive numbers of sensors and network traffic, resource management in IoT networks has become a topic of interest, with advanced deep learning technologies showing promising results \cite{hussain2020machine}. Although we also focus on deep learning for IoT, we are more interested in its role in IoT data processing rather than networking, which is therefore beyond the scope of this survey.

\subsubsection{Data Processing}
Massive sensor data must be processed to extract useful information before being used for further analysis and decision-making. Data mining and machine learning approaches have been used for IoT data processing and analysis \cite{tsai2013data,mahdavinejad2018machine}. Moreover, the context of IoT sensors can provide auxiliary information to help understand sensor data. Therefore, various context-aware computing methods have been proposed for IoT \cite{perera2013context}. There has recently been rapid progress in deep learning, with these positive effects also impacting IoT data processing, e.g., streaming data analysis \cite{mohammadi2018deep}, mobile multimedia processing \cite{ota2017deep}, manufacturing inspection \cite{li2018deep}, and health monitoring. By contrast, we conduct this survey on deep learning for IoT data processing using a new taxonomy, i.e., how deep learning improves the ability of IoT systems to perceive, learn, reason, and behave. Since deep learning is itself a rapidly developing area, our survey covers the latest progress in deep learning in various IoT application domains.

\subsubsection{Security and Privacy}
Massive user data are collected via ubiquitous connected sensors, which may be transmitted and stored in the cloud through IoT networks. These data may contain some biometric information such as faces, voice, or fingerprints. Cyberattacks on IoT systems may result in data leakage, so data security and privacy have become a critical concern in IoT applications \cite{ni2017securing}. Recently, access control \cite{qiu2020survey} and trust management \cite{yan2014survey} approaches have been reviewed to protect the security and privacy of IoT. We also analyze this issue and review progress advanced by AI, such as federated learning \cite{yang2019federated}.

\subsubsection{Applications}
Almost all surveys refer to various IoT application domains including smart cities \cite{zanella2014internet}, smart homes \cite{stojkoska2017review}, smart healthcare \cite{tokognon2017structural}, smart agriculture \cite{elijah2018overview}, and smart industry \cite{da2014internet}. Furthermore, IoT applications based on specific things, e.g., the Internet of Vehicles (IoV) \cite{contreras2017internet} and Internet of Video Things (IoVT) \cite{chen2020internet} have also been rapidly developed. We also summarize some promising applications of AIoT and demonstrate how AI enables them to be faster, smarter, greener, and safer.

\subsection{Organization}
The organization of this paper is shown in Figure~\ref{fig:organization}. We first discuss AIoT computing architecture in Section~\ref{sec:architecture}. Then, we present a comprehensive survey of enabling AI technologies for AIoT in Section~\ref{sec:progress}, followed by a summary of AIoT applications in Section~\ref{sec:applications}. The challenges faced by AIoT and research opportunities are discussed in Section~\ref{sec:challengesOpportunities}, followed by conclusions in Section~\ref{sec:conclusion}.

\section{Architecture}
\label{sec:architecture}
\begin{figure}
  \centering
  \includegraphics[width=1\linewidth]{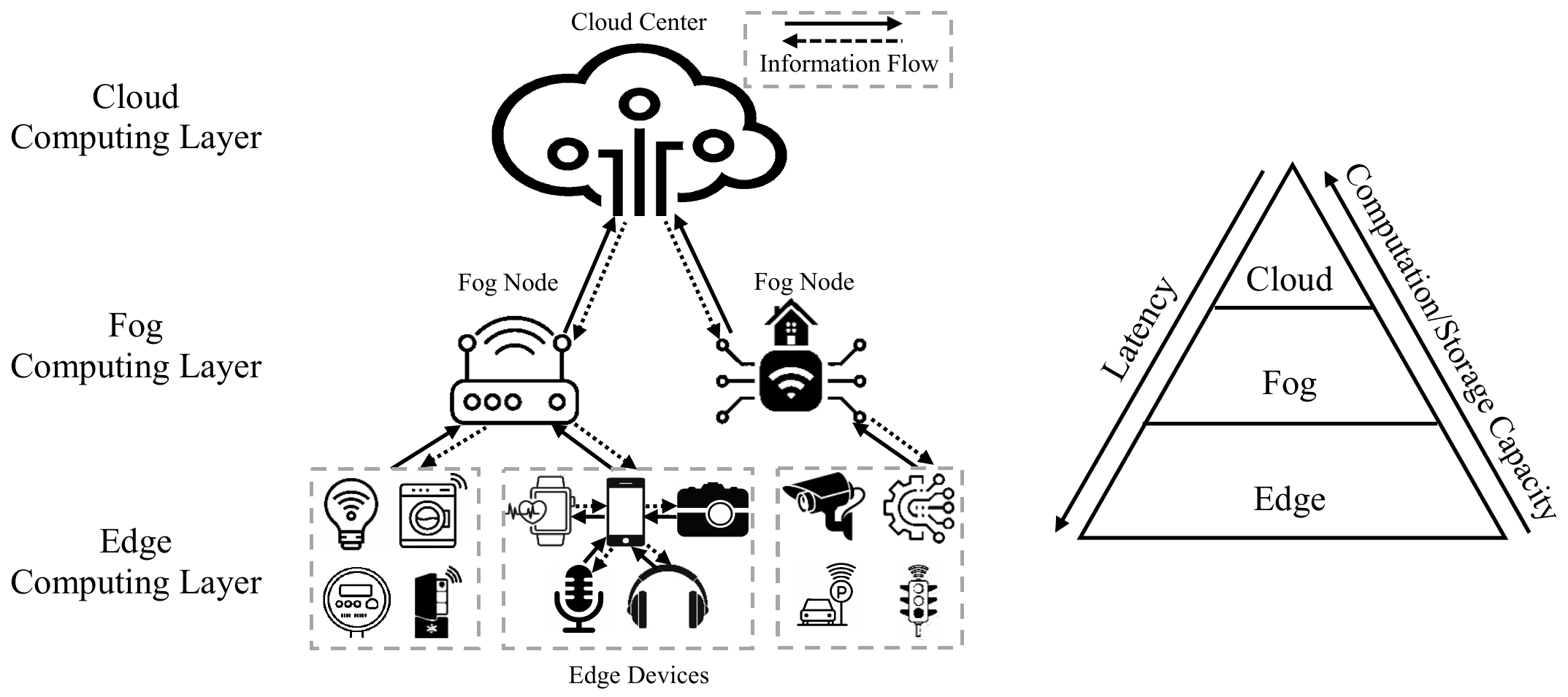}
  \caption{Diagram of the tri-tier computing architecture of AIoT.}
  \label{fig:architecture}
\end{figure}

In this section, we discuss the architecture for AIoT applications. Similar to \cite{lin2017survey,chen2020internet}, we also adopt a tri-tier architecture but from the perspective of computing. For simplicity, we term the three layers as the cloud/fog/edge computing layer, as shown in Figure~\ref{fig:architecture}. The edge computing layer may function like the perception layer in \cite{lin2017survey} and smart visual sensing block in \cite{chen2020internet}. It also supports control and execution over sensors and actuators. Thereby, this layer aims to empower AIoT systems with the ability to perceive and behave. The fog computing layer is embodied in the fog nodes within the networks, like hubs, routers, gateways. The cloud computing layer supports various application services, functioning similarly to the application layer \cite{lin2017survey} and intelligent integration block in \cite{chen2020internet}. The fog and cloud computing layers mainly aim to empower AIoT systems with the ability of learning and reasoning since they can access massive amounts of data and have vast computation resources. It is noteworthy that the edge things and fog nodes are always distributed while the cloud is centralized in the AIoT network topology.

\subsection{Tri-tier Computing Architecture}
\label{subsec:tritierArch}
\subsubsection{Cloud Computing Layer}
The cloud enables AIoT enterprises to use computing resources virtually via the Internet instead of building their physical infrastructure on premises. It can provide flexible, scalable, and reliable resources including computation, storage, and network for enabling various AIoT applications. Typically, real-time data streams from massive distributed sensors and devices are transmitted to the remote cloud center through the Internet, where they are further integrated, processed, and stored. With the off-the-shelf deep learning tools and scalable computing hardware, it is easy to set up the production environment on the cloud, where deep neural networks are trained and deployed to process the massive amounts of data. An important feature of cloud computing is that it provides elastic computing resources in the pay-as-you-go way, which is useful for the AIoT services with fluctuant traffic loads. Another feature is that it can leverage all the data from the registered devices in an AIoT application, which is useful for training deep models with better representation and generalization ability.

\subsubsection{Fog Computing Layer}
Fog computing brings storage, computation, and networking capacity to the edge of the network that is in the proximity of devices. The facilities or infrastructures that provide fog computing service are called fog nodes, e.g., routers, switches, gateways, wireless access points. Although functioning similarly to cloud computing, fog computing offers a key advantage, i.e., low latency, since it is closer to devices. Besides, fog computing can provide continuity of service without the need for the Internet, which is important for specific AIoT applications with an unstable Internet connection, e.g., in agriculture, mining, and shipping domains. The other advantage of fog computing is the protection of data security and privacy since data can be held within the LAN. Fog nodes are better suited for deploying DNNs rather than training since they are designed to store data from local devices, which are incomplete compared with those on the cloud. Nevertheless, model training can still be scheduled on fog nodes by leveraging federated learning \cite{yang2019federated}.

\subsubsection{Edge Computing Layer}
The term of edge computing is interchangeable with fog computing in some literature \cite{shi2016edge,abbas2017mobile} or denotes a broader concept that the fog can be treated as a part of it \cite{tordera2016fog}. Nevertheless, we treat them as different concepts for clarity in this paper. Specifically, we distinguish them based on their locations within the LAN, i.e., fog computing at the network side and edge computing at the thing side. In this sense, edge computing refers to deploying computing capacity on edge devices in proximity to sensors and actuators. A great advantage of edge computing over fog and cloud computing is the reduction of latency and network bandwidth since it can process data into compact structured information on-site before transmission, which is especially useful for AIoT applications using multimedia sensors. However, due to its limited computation capacity, only lightweight DNNs can run on edge devices. Therefore, research topics including neural network architecture design or search for mobile setting and network pruning/compression/quantization have attracted increasing attention recently.

In practice, it is common to deploy multiple different models into cloud platforms, fog nodes, and edge devices in an AIoT system to build an intelligent hybrid computing architecture. By intelligently offloading part of the computation workload from edge devices to the fog nodes and cloud, it is expected to achieve low latency while leveraging deep learning capacities for processing massive amounts of data. For example, a lightweight model can be deployed on edge devices to detect cars in a video stream. It can act as a trigger to transmit keyframes to fog nodes or the cloud for further processing.

\subsection{Hardware and Software}
\subsubsection{Hardware}
While GPU is initially developed for accelerating image rendering on display devices, the general-purpose GPU turns the massive computational power of its shader pipeline into general-purpose computing power (e.g., for massive vector operations), which has sparked the deep learning revolution along with DNN and big data. Lots of operations in the neural network such as convolution can be computed in parallel on GPU, significantly reducing the training and inference time. Recently, an application-specific integrated circuit (ASIC) named TPU is designed by Google specifically for neural network machine learning. Besides, Field-Programmable Gate Arrays (FPGA) have also been used for DNN acceleration due to their low power consumption and high throughput. Several machine learning processors have also been developed for fog and edge computing, e.g., Google Edge TPU and NVIDIA Jetson Nano.

\subsubsection{Software}
Researchers and engineers must design, implement, train, and deploy DNNs easily and quickly. To this end, different open-source deep learning frameworks have been developed, from the beginners like Caffe\footnote{\url{https://github.com/BVLC/caffe}} and MatConvNet\footnote{\url{https://github.com/vlfeat/matconvnet}} to the popular TensorFlow\footnote{\url{https://github.com/tensorflow/tensorflow}} and PyTorch\footnote{\url{https://github.com/pytorch/pytorch}}. MatConvNet is a MATLAB toolbox for implementing Convolutional Neural Networks (CNNs). Caffe is implemented in C++ with Python and Matlab interfaces and well-known for its speed but does not support distributed computation and mobile deployment. Caffe2 improves it accordingly, which has been later merged into PyTorch. The features like dynamic computation graphs and automatic computation of gradients in TensorFlow and PyTorch have made them easy to use and popular. They also support for deploying models into mobile devices by enabling model compression/quantization and hardware acceleration. Porting models among different frameworks is necessary and useful. The Open Neural Network Exchange (ONNX)\footnote{\url{https://github.com/onnx/onnx}} offers this feature by defining an open format built to represent machine learning models, which has been supported by TensorFlow and Pytorch. There are other deep learning frameworks like MXNet\footnote{\url{https://github.com/apache/incubator-mxnet}}, Theano\footnote{\url{https://github.com/Theano/Theano}}, PaddlePaddle\footnote{\url{https://github.com/PaddlePaddle}}, and neural network inference computing framework for mobile devices like ncnn\footnote{\url{https://github.com/Tencent/ncnn}}.

\section{Progress Review of AI for IoT}
\label{sec:progress}
In this section, we comprehensively review the progress of enabling AI technologies for AIoT applications, especially deep learning. We conduct the survey by applying a new taxonomy, i.e., how deep learning improves the ability of AIoT systems for perceiving, learning, reasoning, and behaving. To prevent it from being a survey on deep learning, we carefully select the topics and technologies that are closely related to and useful for various AIoT applications. Moreover, we only outline the trend of the research progress and highlight state-of-the-art technologies rather than diving into the details. We specifically discuss their potentials for AIoT applications. We hope this survey can draw an overall picture of AI technologies for AIoT and provide insights into their utility.

\subsection{Perceiving}
\label{subsec:perceiving}
\begin{figure*}
  \centering
  \includegraphics[width=1\linewidth]{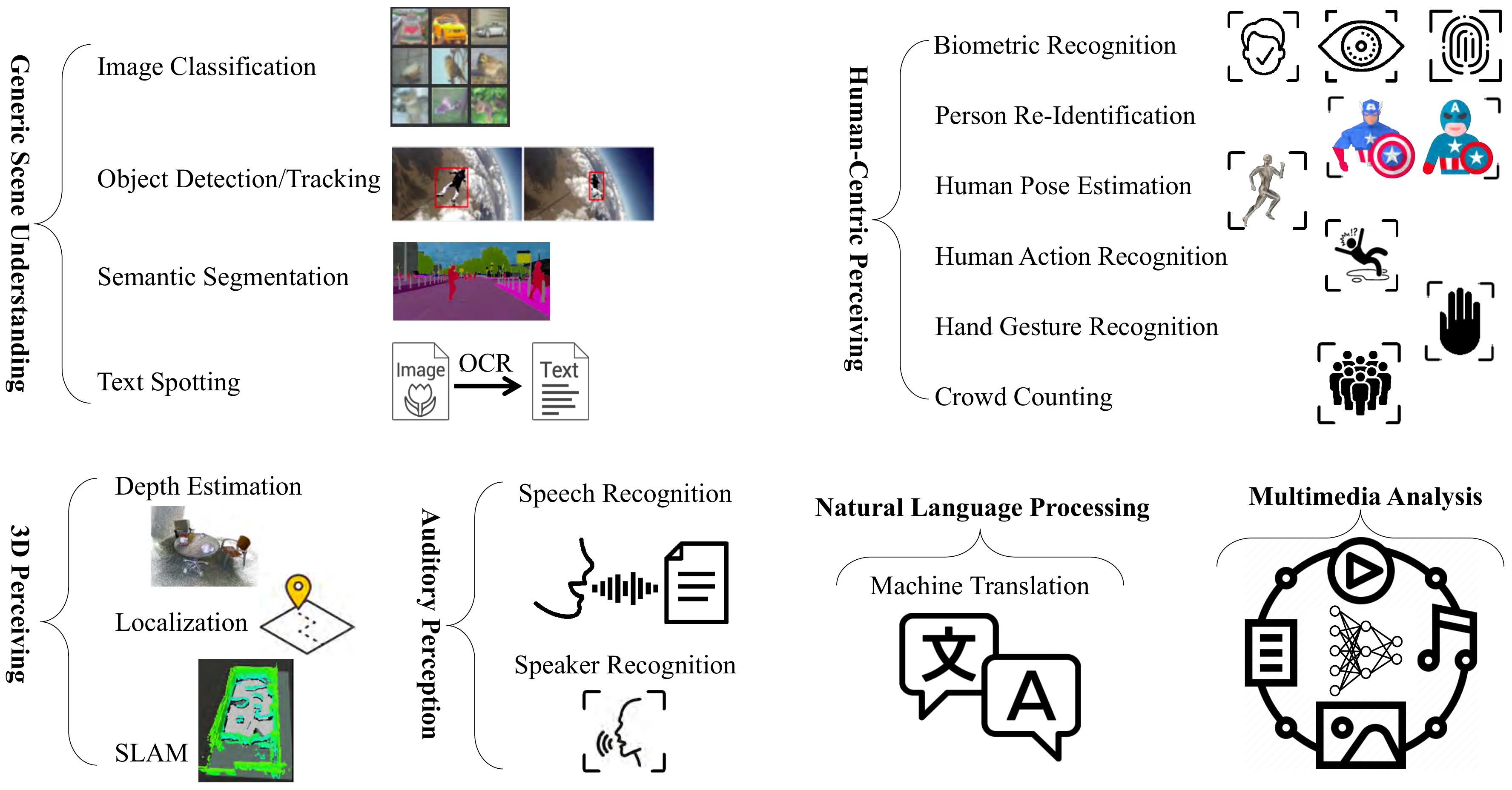}
  \caption{Diagram of the perceiving-related topics in AIoT.}
  \label{fig:perceiving}
\end{figure*}

Empowering things with the perceiving ability, i.e., understanding the environment using various sensors, is fundamental for AIoT systems. In this part, we will focus on several related topics as diagrammed in Figure~\ref{fig:perceiving}.

\textit{First, we present a review of the progress in generic scene understanding including image classification, object detection, and tracking, semantic segmentation, and text spotting.}

\subsubsection{Image Classification}
\label{subsubsec:classification}
\begin{figure*}
  \centering
  \includegraphics[width=1\linewidth]{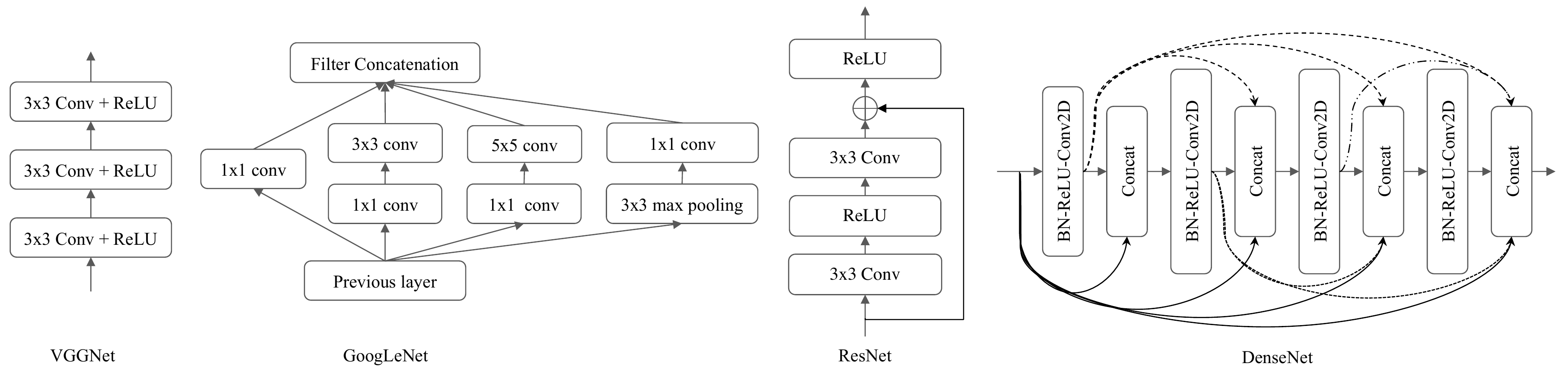}
  \caption{\rev{Basic blocks of representative deep CNNs.}}
  \label{fig:networkblocks}
\end{figure*}

Image classification refers to recognizing the category of an image. Classical machine learning methods based on hand-crafted features have been surpassed by DNNs \cite{krizhevsky2012imagenet} on large-scale benchmark datasets like ImageNet \cite{deng2009imagenet}, sparking a wave of research on the architecture of DNNs. \rev{From AlexNet \cite{krizhevsky2012imagenet} to ResNet \cite{he2016deep}, more and more advanced network architectures have been devised by leveraging stacked 3$\times$3 convolutional layer for reducing network parameters and increasing network depth, 1$\times$1 convolutional layer for feature dimension reduction, residual connections for preventing gradient vanishing and increasing network capacity, and dense connections for reusing features from previous layers as shown in Figure~\ref{fig:networkblocks}. A brief summary of representative deep CNNs is listed in Table~\ref{tab:networksummary}. As can be seen, with the increase of network depth and the number of parameters, the representation capacity also increases, leading to lower top1 classification error on the ImageNet dataset. Besides, the architecture of the network matters. Even with fewer model parameters and computational complexity, the recently proposed networks such as ResNet and DenseNet outperform previous ones such as VGGNet.} Lightweight networks are appealing to AIoT applications where DNNs are deployed on edge devices. Recently, some computationally efficient networks like MobileNet have been proposed by leveraging depth-wise convolutions \cite{howard2017mobilenets}, point-wise convolutions \cite{zhang2018fully}, 
or binary operations \cite{rastegari2016xnor}. \rev{Besides, network compression such as pruning and quantization can be used to obtain lightweight models from heavy models, which will be reviewed in Section~\ref{subsubsec:nas}.} Image recognition can be very useful in many AIoT applications, such as smart education tools or toys, which help and teach children to explore the world with cameras. Besides, some popular applications in smartphones also benefit from the advances in this area for recognizing flowers and birds, and food items and calories.

\begin{table}[htbp]
  \centering
  \caption{\rev{A summary of representative deep CNNs. Param.: number of parameters; Comp.: computational complexity (MACs).}}
    \begin{tabular}{llllll}
    \toprule
    Network & Year  & Depth & Param. (M) & Comp. (G) & top1-err \\
    \midrule
    AlexNet & 2012  & 8     & 61.10  & 0.72  & 43.45 \\
    \hline
    \multicolumn{1}{l}{\multirow{4}[0]{*}{VGGNet}} & \multicolumn{1}{l}{\multirow{4}[0]{*}{2014}}  & 11    & 132.86 & 7.63  & 30.98 \\
          &       & 13    & 133.05 & 11.34 & 30.07 \\
          &       & 16    & 138.36 & 15.5  & 28.41\\
          &       & 19    & 143.67 & 19.67 & 27.62 \\
    \hline
    GoogLeNet & 2014  & 22    & 6.62  & 1.51  & 30.22 \\
    \hline
    Inception v3 & 2015  & 48    & 27.16 & 2.85  & 22.55 \\
    \hline
    \multicolumn{1}{l}{\multirow{5}[0]{*}{ResNet}} & \multicolumn{1}{l}{\multirow{5}[0]{*}{2015}}  & 18    & 11.69 & 1.82  & 30.24 \\
          &       & 34    & 21.80  & 3.68  & 26.70 \\
          &       & 50    & 25.56 & 4.12  & 23.85 \\
          &       & 101   & 44.55 & 7.85  & 22.63  \\
          &       & 152   & 60.19 & 11.58 & 21.69 \\
    \hline
    \multicolumn{1}{l}{\multirow{4}[0]{*}{DenseNet}} & \multicolumn{1}{l}{\multirow{4}[0]{*}{2016}}  & 121   & 7.98  & 2.88  & 25.35 \\
          &       & 169   & 14.15 & 3.42  & 24.00  \\
          &       & 201   & 20.01 & 4.37  & 22.80  \\
          &       & 161   & 28.68 & 7.82  & 22.35 \\
    \bottomrule
    \end{tabular}%
  \label{tab:networksummary}%
\end{table}%

\subsubsection{Object Detection}
\label{subsubsec:detection}
Generic object detection refers to recognizing the category and location of an object, which is used as a prepositive step for many down-stream tasks including face recognition, person re-identification, pose estimation, behavior analysis, and human-machine interaction. The methods for object detection from images have been revolutionized by DNNs. State-of-the-art methods can be categorized into two groups: two-stage methods and one-stage methods. The former follows a typical ``proposal$\rightarrow$detection'' paradigm \cite{ren2015faster}, while the latter directly evaluates all the potential object candidates and outputs the detection results \cite{redmon2016you}. Recently, one-stage anchor-free detectors have been proposed by representing object location using points or regions rather than anchors \cite{law2018cornernet}, achieving a better trade-off between speed and accuracy, which is appealing to AIoT applications that require onboard detection. Detection of specific category of objects such as pedestrian, car, traffic-sign, and the license plate has been widely studied, which are useful for improving the perceiving ability of AIoT systems for traffic and public safety surveillance and autonomous driving \cite{chen2015deepdriving}. Besides, object detection is a crucial technique for video data structuring in many AIoT systems using visual sensors, which aims to extract and organize compact structured semantic information from video data for further retrieval, verification, statistics, and analysis at low transmission, storage, and computation cost.

\subsubsection{Object Tracking}
\label{subsubsec:tracking}
Classical object tracking methods include generative and discriminative methods where the former ones try to search the most similar regions to the target and the latter ones leverage both foreground target and background context information to train an online discriminative classifier \cite{li2013survey}. Later, different deep learning methods have been proposed to improve the classical methods by learning multi-resolution deep features \cite{danelljan2016beyond}, end-to-end representation learning \cite{valmadre2017end}, and leveraging siamese networks \cite{li2019siamrpn}. Object trackers usually run much faster than object detectors, which can be deployed on edge devices of AIoT applications such as video surveillance and autonomous driving for object trajectory generation and motion prediction. One possible solution is to leverage the hybrid computation architecture (see Section~\ref{subsec:tritierArch}) by deploying object tracker on edge devices while deploying object detectors on the fog nodes or cloud, i.e., tracking across all frames while detecting only on keyframes. In this way, only keyframes and the compact structured detection results should be transmitted via the network, thereby reducing the network bandwidth and processing latency.

\begin{table*}[htbp]
  \centering
  \caption{\rev{Representative benchmark datasets in generic scene understanding. BBox: Bounding Box; Mask: Pixel-level Semantic Mask.}}
    \begin{tabular}{lllll}
    \toprule
    Area  & Dataset & Link  & Volume & Label Type \\
    \midrule
    \multicolumn{1}{l}{\multirow{4}[0]{*}{Image Classification}} & ImageNet & \url{http://www.image-net.org/} & 1.2M  & Category, BBox \\
          & CIFAR-10/-100 & \url{https://www.cs.utoronto.ca/~kriz/cifar.html} & 6k  & Category \\
          & Caltech-UCSD Birds & \url{http://www.vision.caltech.edu/visipedia/CUB-200-2011.html} & 11,788 & Category, Attributes, BBox \\
          & Caltech 256 & \url{http://www.vision.caltech.edu/Image\_Datasets/Caltech256/} & 30,607 & Category \\
          \hline
    \multicolumn{1}{l}{\multirow{2}[0]{*}{Object Detection}} & COCO  & \url{https://cocodataset.org/\#detection-2020} & 200k & Category, BBox, Mask \\
          & Pascal VOC & \url{http://host.robots.ox.ac.uk/pascal/VOC/} & 10k & Category, BBox, Mask\\
          \hline
    \multicolumn{1}{l}{\multirow{3}[0]{*}{Object Tracking}} & MOT   & \url{https://motchallenge.net/} & 22 & BBox \\
          & KITTI-Tracking & \url{http://www.cvlibs.net/datasets/kitti/eval\_tracking.php} & 50 & 3D BBox \\
          & UA-DETRAC & \url{http://detrac-db.rit.albany.edu/} & 140k & BBox \\
          \hline
    \multicolumn{1}{l}{\multirow{3}[0]{*}{Semantic Segmentation}} & Cityscape & \url{https://www.cityscapes-dataset.com/} & 25k & Mask\\
          & ADE20K & \url{https://groups.csail.mit.edu/vision/datasets/ADE20K/} & 22,210 & Category, Attributes, Mask \\
          & PASCAL-Context & \url{https://cs.stanford.edu/~roozbeh/pascal-context/} & 19,740 & Mask\\
          \hline
     \multicolumn{1}{l}{\multirow{3}[0]{*}{Text Spotting}} & Total-text & \url{https://github.com/cs-chan/Total-Text-Dataset} & 1,555  & Polygon Box, Text\\
          & SCUT-CTW1500 & \url{https://github.com/Yuliang-Liu/Curve-Text-Detector} & 1,500  & BBox, Text \\
          & LSVT  & \url{https://ai.baidu.com/broad/introduction?dataset=lsvt} & 450k & Binary Mask, Text \\
    \bottomrule
    \end{tabular}%
  \label{tab:datasetgenericscene}%
\end{table*}%

\subsubsection{Semantic Segmentation}
\label{subsubsec:segmentation}
\rev{Semantic segmentation refers to predicting pixel-level category label for an image. The fully CNN with an encoder-decoder structure has become the \textit{de facto} paradigm for semantic segmentation \cite{ronneberger2015u,long2015fully}, since it can learn discriminative and multi-resolution features through cascaded convolution blocks while preserving spatial correspondence. Many deep models have been proposed to improve the representation capacity and prediction accuracy from the following three aspects: context embedding, resolution enlarging, and boundary refinement. Efficient modules are proposed to exploit context information and learn more representative feature representations such as global context pooling module in the ParseNet \cite{liu2015parsenet}, atrous spatial pyramid pooling in the DeepLab models \cite{chen2017deeplab}, and the pyramid pooling module in the PSPNet \cite{zhao2017pyramid}. Enlarging the resolution of feature maps is beneficial for improving prediction accuracy, especially for small objects. Typical techniques include using the deconvolutional layer, unpooling layer, and dilated convolutional layer. Boundary refinement aims to obtain sharp boundaries between different categories in the segmentation map, which can be achieved by using conditional random field as the post-processing technique on the predicted probability maps \cite{chen2017deeplab}.} 

\rev{There are two research topics related to semantic segmentation, i.e., instance segmentation and panoptic segmentation. Instance segmentation refers to detecting foreground objects as well as obtaining their masks. A well-known baseline model is Mask R-CNN which adopts an extra branch for object mask prediction in parallel with the existing one for bounding box regression \cite{he2017mask}. Its performance can be improved further by exploiting and enhancing the feature hierarchy of deep convolutional networks \cite{lin2017feature}, employing non-local attention \cite{wang2018non}, and leveraging the reciprocal relationship between detection and segmentation via hybrid task cascade \cite{chen2019hybrid}. Panoptic segmentation refers to simultaneously segmenting the masks of foreground objects as well as background stuff \cite{kirillov2019panoptic}, i.e., unifying both the semantic segmentation and instance segmentation tasks. A simple but strong baseline model is proposed in \cite{kirillov2019panopticfeature}, which adds a semantic segmentation branch into the Mask R-CNN framework and uses a shared feature pyramid network backbone \cite{lin2017feature}. Semantic segmentation in many sub-areas such as medical image segmentation \cite{ronneberger2015u}, road detection \cite{chen2019progressive}, and human parsing \cite{he2020grapy}, are useful in various AIoT applications. For example, it can be used to recognize the dense pixel-level drivable area and traffic participants like cars and pedestrians, which can be further combined with 3D measure information to get a comprehensive understanding of the driving context and make smart driving decisions accordingly. Moreover, obtaining the foreground mask or body parts matters for many AIoT applications, e.g., video editing for entertainment and computational advertising, virtual try-on, and augmented/virtual reality (AR/VR). Besides, the structured semantic mask is also useful for semantic-aware efficient and adaptive video coding.}

\subsubsection{Text Spotting}
\label{subsubsec:text}
Text spotting is a composite task including text detection and recognition. Although text detection is related to generic object detection, it is a different and challenging problem: 1) while generic objects have regular shapes, text may be in variable length and shape depending on the number of characters and their orientation; 2) the appearance of same text may change significantly due to fonts, styles as well as background context. Deep learning has advanced this area by learning more representative feature \cite{jaderberg2014deep}, devising better representation of text proposals \cite{liu2018fots}, and using large-scale synthetic dataset \cite{gupta2016synthetic}. Recently, end-to-end modeling of text detection and recognition has achieved impressive performance \cite{li2017towards,liu2020asts}. Each sub-task can benefit from the other by leveraging more supervisory signals and learning a shared feature representation. Moreover, rather than recognizing text at the character level, recognizing text at the word or sentence level can benefit from the word dictionary and language model. Specifically, the idea of sequence-to-sequence modeling and connectionist temporal classification (CTC) \cite{graves2006connectionist} from the areas of speech recognition and machine translation has also been explored. Since the text is very common in real-world scenes, e.g., traffic sign, nameplate, information board, text spotting can serve as a useful tool in many AIoT applications for ``reading'' text information from scene images, e.g., live camera translator for education, reading assistant for the visually impaired \cite{mancas2007multifunctional}, optical character recognition (OCR) for automatic document analysis, store nameplate recognition for self-localization and navigation. \rev{The representative benchmark datasets in the areas related to generic scene understanding are summarized in Table~\ref{tab:datasetgenericscene}.}

\textit{Next, we present a review of the progress in human-centric perceiving including biometric recognition such as face/fingerprint/iris recognition, person re-identification, pose/gesture/action estimation, and crowd density estimation.}

\subsubsection{Biometric Recognition}
\label{subsubsec:biometric}
Biometric recognition based on face, fingerprint, or iris, is a long-standing research topic. We first review the progress in face recognition. There are usually four key stages in a face recognition system, i.e., face detection, face alignment, face representation, and face classification/verification. Face detection, as a specific sub-area of object detection, benefits from the recent success of deep learning in generic object detection. Nevertheless, special effort should be made to address the following challenges, i.e., vast scale variance, severe imbalance of positive and negative proposals, profile and front face, occlusion, and motion blur. One of the most famous classical methods is the Viola-Jones algorithm, which sets up the fundamental face detection framework \cite{viola2001rapid}. The idea of using cascade classifiers inspires many deep learning methods such as cascade CNN \cite{li2015convolutional}. Recently, jointly modeling face detection with other auxiliary tasks including face alignment, pose estimation, and gender classification, can achieve improved performance, owing to the extra abundant supervisory signals for learning a shared discriminative feature representation \cite{zhang2016joint,ranjan2017hyperface}. Note that the all-in-one model is appealing to some AIoT application where multiple structured facial information could be extracted. 

Face alignment, a.k.a. facial landmark detection aims to detect facial landmarks from a face image, which is useful for front face alignment and face recognition. Typically, face facial landmark detectors are trained and deployed in a cascade manner that a shape increment is learned and used to update the current estimate at each level \cite{ren2014face}. Face landmark detectors are usually lightweight and run very fast, which are very useful for latency-sensitive AIoT applications. 

For face recognition, significant progress has been achieved in the last decade, mainly owing to deep representation learning and metric learning. The milestone work in \cite{taigman2014deepface} propose to learn discriminative deep bottleneck features using classification and verification losses. Nevertheless, they face a challenge to scale to orders of magnitude larger datasets with more identities. To address this issue, a representation learning method using triplet loss is proposed to directly learn discriminative and compact face embedding \cite{schroff2015facenet}. Face recognition is one of the most widely used perceiving techniques for identity verification and access control in various AIoT applications, e.g., smart cities and smart homes. Associating the facial identity with one's accounts can create vast business value, e.g., mobile payment, membership development and promotion, fast track in smart retail. Regarding the number of people to be recognized and privacy concerns, either offline or online solutions can be used, where models are deployed on edge devices, fog nodes, or cloud centers \cite{soyata2012cloud,amos2016openface}. A research topic related to practical face recognition applications is liveness detection and spoof detection. Different methods have been proposed based on action imitation, speech collaboration, and multi-modal sensors \cite{ramachandra2017presentation}.

\rev{In addition to face recognition, iris, fingerprint, and palmprint recognition have also been studied for a long period and are widely used in practical AIoT applications. Compared with fingerprint, palmprint has abundant features and can be captured using common built-in cameras of mobile phones rather than sensitive sensors. Typically, a palmprint recognition system is composed of a palmprint image acquisition system, a palmprint region of interest (ROI) extraction module, a feature extraction module, and a feature matching module for recognition or verification. Both hand-crafted features such as line features, orientation-based features, and the orthogonal line ordinal feature and deep learning-based feature representation have been studied in the literature \cite{fei2018feature,zhang2019pay}. For example, Zhang et al. propose a novel device to capture palmprint images in a contactless way, which can be used for access control, aviation security, and e-banking \cite{zhang2017towards}. It uses block-wise statistics of competitive code as features and the collaborative representation-based framework for classification. Besides, a DCNN-based palmprint verification system named DeepMPV is proposed for mobile payment in \cite{zhang2019pay}. It first extracts the palmprint ROI by using pre-trained detectors and then trains a siamese network to match palmprints. Recently, Amazon has announced its new payment system called Amazon One, which is a fast, convenient, contactless way for people to use their palms to make payments based on palmprint recognition. In practice, the choice of a specific biometric recognition solution depends on sensors, usage scenarios, latency, and power consumption. Although biometric recognition offers great utility, the concerns about data security and privacy have to be carefully addressed in practical AIoT systems.}

\begin{figure*}
\centering
\subfloat[]{\label{fig:coco_stats_train_occlusions}
\includegraphics[width=0.495\linewidth]{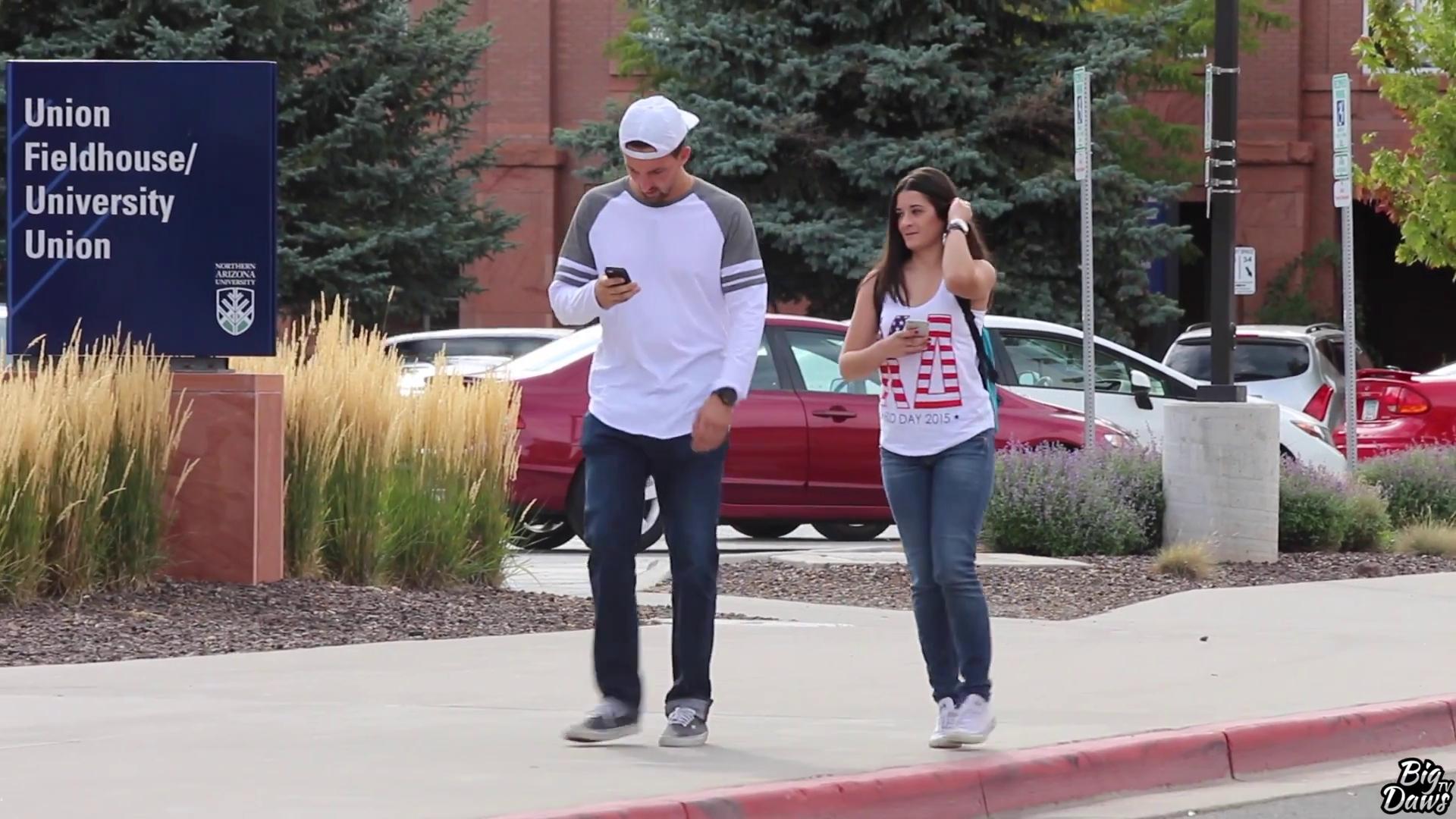}}
\vspace{0.001\linewidth}
\subfloat[]{\label{fig:coco_statas_train_vis_invis}
\includegraphics[width=0.495\linewidth]{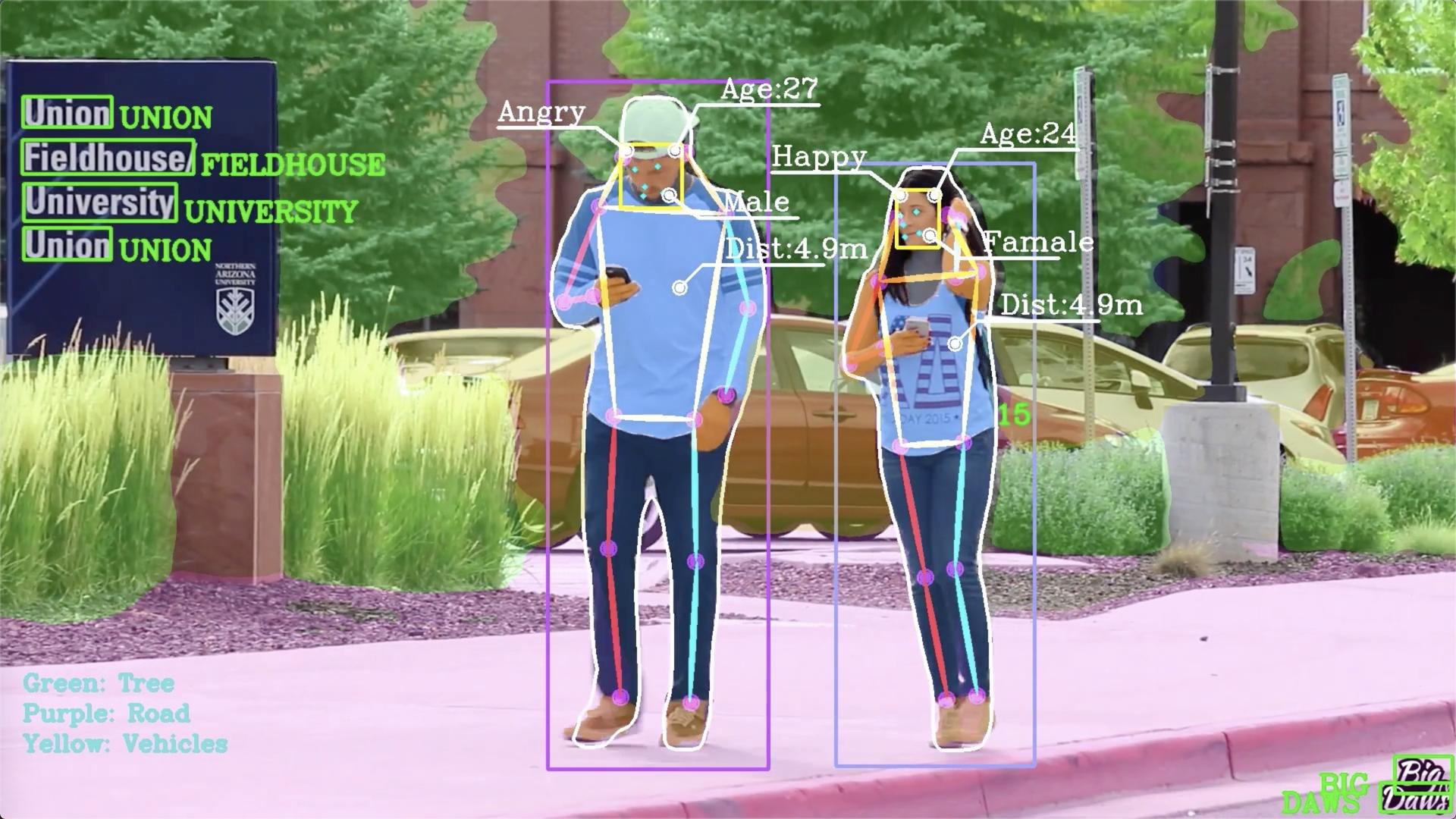}}

\caption{Demonstration of AI techniques for generic scene understanding, human-centric perceiving and 3D perceiving. (a) A frame from the video ``Walking Next to People''\protect\footnotemark[11]. (b) The processed result by using different perceiving methods, i.e., semantic segmentation \cite{chen2019progressive}; object detection \cite{chen2020recursive}; text spotting \cite{liu2020asts}; human parsing \cite{he2020grapy}; human pose estimation \cite{zhang2020towards}; face detection, alignment, and facial attribute analysis \cite{zhang2016joint,zeng2019soft}; depth estimation \cite{fu2018deep}.}
\label{fig:perceivingn_demo}
\end{figure*}

\subsubsection{Person Re-Identification}
\label{subsubsec:personreid}
Person re-identification, as a sub-area of image retrieval, refers to recognizing an individual captured in disjoint camera views. In contrast to face recognition in a controlled environment, person re-identification is more challenging due to the variations in the uncontrolled environment, e.g., viewpoint, resolution, clothing, and background context. To address these challenges, different methods have been proposed \cite{ye2020deep} including deep metric learning based on various losses, integration of local features and context, multi-task learning based on extra attribute annotations, and using human pose and parsing mask as guidance. Recently, generative adversarial networks (GAN) have been used to generate style-transferred images for bridging the domain gap between different datasets \cite{wei2018person}. Person re-identification has vast potential for AIoT applications such as smart security in an uncontrolled and non-contact environment, where other biometric recognition techniques are not applicable. Although extra efforts are needed to build practical person re-identification systems, one can leverage the idea of human-in-the-loop artificial intelligence to achieve high performance with low labor effort. For example, the person re-identification model can be used for initial proposal ranking and filtering, then human experts are involved to make final decisions.
 
\subsubsection{Human Pose Estimation and Gesture/Action Recognition}
\label{subsubsec:pose}
Human pose estimation, a.k.a. human keypoint detection refers to detecting body joints from a single image. There are two groups of human pose estimation methods, i.e., top-down methods and bottom-up methods. The former consists of two stages including person detection and keypoint detection, while the latter directly detects all keypoints from the image and associates them with corresponding person instances. Although top-down methods still dominate the leaderboard of public benchmark datasets like MS COCO\footnote{\url{http://cocodataset.org/index.htm\#keypoints-leaderboard}}, they are usually slower than bottom-up methods \cite{cao2017realtime}. Recent progress in this area can be summarized in the following aspects: 1) learning better feature representation from stronger backbone network, multi-scale feature fusion, or context modeling \cite{sun2019deep}; 2) effective training strategy including online hard keypoint mining, hard negative person detection mining, and harvesting extra data \cite{zhang2020towards}; 3) sub-pixel representation or post-processing techniques \cite{zhang2020towards,zhang2020distribution}. Recently, dealing with pose estimation in crowd scenes with severe occlusions also attracts much attention. The other related topic is 3D human pose estimation from a single image or multi-view images \cite{wang2019not}, aiming to estimate the 3D coordinate of each keypoint rather than the 2D coordinate on the image plane.

Once we detect the human keypoints for each frame given a video clip, the skeleton sequence for each person instance can be obtained, from which we can recognize the action. This process is known as skeleton-based action recognition. To model the long-term temporal dependencies and dynamics as well as spatial structures within the skeleton sequence, different neural networks have been exploited for action recognition such as the deep recurrent neural network (RNN) \cite{du2015hierarchical}, CNN \cite{ke2017new}, deep graph convolutional networks (GCN) \cite{yan2018spatial}. Besides, since some joints may be more relevant to specific actions than others, attention mechanism has been used to automatically discover informative joints and emphasize their importance for action recognition \cite{song2017end}. Estimation of human pose and recognition of action can be very useful in many real-world AIoT scenarios, such as rehabilitation exercises monitoring and assessment \cite{gonzalez2014kinect}, dangerous behavior monitoring \cite{kashevnik2019methodology}, and human-machine interaction (HMI).
 
\footnotetext[11]{\url{https://www.youtube.com/watch?v=__eLCXUKtec}}
\setcounter{footnote}{11}

\begin{table*}[htbp]
  \centering
  \caption{\rev{Representative benchmark datasets in human-centric perceiving. ID: Identity; BBox: Bounding Box.}}
    \begin{tabular}{lllll}
    \toprule
    Area  & Dataset & Link  & Volume & Label Type \\
    \midrule
    \multicolumn{1}{l}{\multirow{3}[0]{*}{Face Recognition}} & FFHQ & \url{https://github.com/NVlabs/ffhq-dataset} & 70k & ID \\
          & FDDB  & \url{http://vis-www.cs.umass.edu/fddb/} & 2,845  & ID, BBox \\
          & YouTube Faces DB & \url{https://www.cs.tau.ac.il/~wolf/ytfaces/} & 3,425 & ID, BBox \\
          \hline
    \multicolumn{1}{l}{\multirow{2}[0]{*}{Fingerprint Recognition}} & FVC2000 & \url{http://bias.csr.unibo.it/fvc2000/databases.asp} & 3,520  & ID \\
          & LivDet Databases & \url{http://livdet.org/registration.php} & 11k & ID \\
          \hline
    \multicolumn{1}{l}{\multirow{2}[0]{*}{Iris Recognition}} & LivDet Databases & \url{http://livdet.org/registration.php} & 7,223  & ID \\
          & IrisDisease & \url{http://zbum.ia.pw.edu.pl/AGREEMENTS/IrisDisease-v2\_1.pdf} & 2,996  & ID \\
          \hline
    \multicolumn{1}{l}{\multirow{3}[0]{*}{Person Re-ID}} & Market-1501 & \url{http://zheng-lab.cecs.anu.edu.au/Project/project\_reid.html} & 1,501 & ID, BBox \\
          & DukeMTMC-ReID & \url{https://github.com/sxzrt/DukeMTMC-reID\_evaluation} & 1,404 & ID, BBox \\
          & CUHK03 & \url{https://www.ee.cuhk.edu.hk/~xgwang/CUHK\_identification.html} & 1,360 & ID, BBox \\
          \hline
    \multicolumn{1}{l}{\multirow{3}[0]{*}{Pose Estimation}} & COCO  & \url{https://cocodataset.org/\#keypoints-2020} & 200k & Keypoints \\
          & MPII  & \url{http://human-pose.mpi-inf.mpg.de/} & 25k & Keypoints \\
          & DensePose-COCO & \url{http://densepose.org/} & 50k & Keypoints \\
          \hline
    \multicolumn{1}{l}{\multirow{2}[0]{*}{Gesture Recognition}} & DVS128 & \url{https://www.research.ibm.com/dvsgesture/} & 1,342 & Category \\
          & MS-ASL & \url{https://www.microsoft.com/en-us/research/project/ms-asl/} & 25k & Category \\
          \hline
    \multicolumn{1}{l}{\multirow{2}[0]{*}{Action Recognition}} & UCF101 & \url{https://www.crcv.ucf.edu/data/UCF101.php} & 13,320 & Category \\
          & ActivityNet & \url{http://activity-net.org/} & 19,994 & Category \\
          \hline
    \multicolumn{1}{l}{\multirow{3}[0]{*}{Crowd Counting}} & NWPU-Crowd & \url{https://gjy3035.github.io/NWPU-Crowd-Sample-Code/} & 5,109  & Dots, BBox \\
          & JHU-CROWD++ & \url{http://www.crowd-counting.com/} & 4,372  & Dots, BBox \\
          & UCF-QNRF & \url{https://www.crcv.ucf.edu/data/ucf-qnrf/} & 1,535  & Dots \\
    \bottomrule
    \end{tabular}%
  \label{tab:datasethuman}%
\end{table*}%

Hand gesture recognition is also a hot research topic and has many practical applications such as HMI and sign language recognition. Different sensors can be used in AIoT systems for gesture recognition, such as millimeter-wave radar and visual sensors like RGB camera, depth camera, and event camera \cite{rautaray2015vision,lien2016soli,amir2017low}. Nevertheless, due to the prevalence of cameras and great progress in deep learning and computer vision, visual hand gesture recognition has the vast potential, which can be categorized into two groups, i.e., static ones and dynamic ones. The former aims to match the gesture in a single image to some predefined gestures, while the latter tries to recognize the dynamic gesture from an image sequence, which is more useful. Usually, there are three phases in dynamic hand gesture recognition, i.e., hand detection, hand tracking, and gesture recognition. While hand detection and tracking can benefit from recent progress in generic object detection and tracking as described in Section~\ref{subsubsec:detection} and \ref{subsubsec:tracking}, hand gesture recognition can also borrow useful ideas from the area of action recognition, e.g., exploiting RNN and 3D CNN to capture the gesture dynamics from image sequences. Hand gesture recognition can be very useful for interactions with things in AIoT systems, e.g., non-contact control of television and car infotainment system, and communication with the speech and hearing impaired \cite{starner1998real}.

\subsubsection{Crowd Counting}
In the video surveillance scenario, it is necessary to count the crowd in both indoor and outdoor areas and prevent crowd congestion and accident. For practical AIoT applications with crowd counting ability, WI-FI, Bluetooth, and camera-based solutions have been proposed by estimating the connections between smartphones and WI-FI access points or Bluetooth beacons \cite{schauer2014estimating} or estimating the crowd density of a crowd image \cite{wang2020nwpu}. Although counting the detected faces or heads in a crowd image can be used for crowd counting intuitively, the person instance in a crowd image is always in relatively low resolution and blurry, which limits the performance of the detection model. Besides, detecting a vast amount of persons in a single shot is computationally inefficient. Therefore, most CNN-based methods directly regress the crowd density map, in which the ground truth is constructed by placing Gaussian density maps at the head regions. Since it is costly to collect and annotate crowd images, synthetic datasets can be used and have demonstrated its value for this task, i.e., either being used in the pretraining-finetuning scheme or by domain adaptation \cite{wang2019learning}. Despite the progress in this area, more efforts are needed to address real-world challenges for practical AIoT applications, e.g., designing lightweight and computational efficient crowd counting models, simultaneous crowd counting and crowd flow estimation, and integration of multi-modal sensors for more accurate crowd counting. \rev{The representative benchmark datasets in the aforementioned research areas related to human-centric perceiving are summarized in Table~\ref{tab:datasethuman}.}

\begin{table*}[htbp]
  \centering
  \caption{\rev{Representative benchmark datasets in 3D perceiving.}}
    \begin{tabular}{lllll}
    \toprule
    Area  & Dataset & Link  & Volume & Label Type \\
    \midrule
     \multicolumn{1}{l}{\multirow{3}[0]{*}{Depth Estimation}} & KITTI & \url{http://www.cvlibs.net/datasets/kitti/eval\_depth.php} & 93k   & Depth Maps \\
          & NYU Depth Dataset V2 & \url{https://cs.nyu.edu/~silberman/datasets/nyu\_depth\_v2.html} & 1,449  & Depth Maps \\
          & Make3D & \url{http://make3d.cs.cornell.edu/data.html\#object} & 534   & Depth Maps \\
          \hline
    \multicolumn{1}{l}{\multirow{3}[0]{*}{SLAM}}  & KITTI & \url{http://www.cvlibs.net/datasets/kitti/eval\_odometry.php} & 22 & Poses \\
          & EUROC MAV Dataset & \url{https://projects.asl.ethz.ch/datasets/doku.php?id=kmavvisualinertialdatasets} & 12 & Poses \\
          & TUM Visual-Inertial & \url{https://vision.in.tum.de/data/datasets/visual-inertial-dataset} & 43 & Poses \\
    \bottomrule
    \end{tabular}%
  \label{tab:dataset3d}%
\end{table*}%

\textit{In the following, we review several topics related to 3D perceiving including depth estimation, localization, and simultaneous localization and mapping (SLAM).}

\subsubsection{Depth Estimation/Localization/SLAM}
Estimating depth using cameras is a long-standing research topic \cite{hoiem2007recovering,wang2008region,mahjourian2018unsupervised,fu2018deep}. In real-world AIoT applications, there can be several configurations such as the monocular camera, stereo camera, multi-view camera system. Recently, depth estimation from monocular video together with camera pose estimation has attracted a lot of attention. In contrast to traditional matching and optimization-based methods, current research on this topic mainly focuses on deep learning in an unsupervised or self-supervised way \cite{gordon2019depth}. Nevertheless, they construct the self-supervisory signals based on the re-projection photometric loss w.r.t. depth and camera pose derived from the well-defined multi-view geometry, which is similar to the matching error or photometric error terms in the traditional optimization objective. Although CNN has powerful representation capacity, special effort has to be made to address the challenges including occlusions and dynamic objects, as well as the scale issue (per-frame ambiguity and temporally inconsistent). 

The aforementioned camera pose estimation is also related to visual odometry (VO) and visual-inertial odometry (VIO) \cite{yang2020d3vo,han2019deepvio}, which aim to calculate sequential camera poses of an agent based on the camera and Inertial Measurement Unit (IMU) sensors. VO and VIO are always used ad the front-end in a SLAM system, where the back-end refers to the nonlinear optimization of the pose graph, aiming to obtain globally consistent and drift-free pose estimation results. In traditional methods like ORB-SLAM \cite{mur2015orb}, the front-end and back-end are two separate modules. Recently, a differentiable architecture named neural graph optimizer is proposed for global pose graph optimization \cite{parisotto2018global}. Together with a local pose estimation model, it achieves a complete end-to-end neural network solution for SLAM. 

Depth estimation, pose estimation, VO/VIO, and SLAM constitute the important 3D perceiving ability of AIoT, which could be very useful in smart transportation \cite{shao2020tightly}, smart industry \cite{krug2016next}, smart agriculture \cite{cheein2011optimized,alsalam2017autonomous,liu2018robust}, smart cities and homes \cite{lee2016rgb,wang2017enabling,katzschmann2018safe}. For example, deploying multiple cameras at different viewpoints, one can construct a multi-view visual system for depth estimation and object or scene 3D reconstruction. In the autonomous driving scenario, depth estimation can be integrated into the object detection module and road detection module for forward collision warning. Besides, SLAM can be used for lane departure warning, lane-keeping, high-precision map construction and update \cite{shao2020tightly}. Other use cases may include self-localization and navigation for the agricultural robot, sweeper robot, service robot, and unmanned aerial vehicle \cite{cheein2011optimized,krug2016next,alsalam2017autonomous}. \rev{The representative benchmark datasets in the areas of 3D perceiving are summarized in Table~\ref{tab:dataset3d}.}

\textit{\rev{Due to sensor quality and imaging conditions, the captured image may need to be pre-processed to enhance illumination, increase contrast, and rectify distortions before being used in the aforementioned visual perception tasks. In the following, we briefly review the recent progress in the area of image enhancement as well as image rectification and stitching.}}

\subsubsection{Image Enhancement}
\label{subsubsec:imageenhancement}
\rev{Image enhancement is a task-oriented task that refers to enhancing specific property of a given image, such as illumination, contrast, sharpness. Images captured in a low-light environment are in low visibility and difficult to see details due to insufficient incident light or underexposure. An image can be decomposed into the reflectance map and illumination map based on the Retinex theory \cite{land1977retinex}. Then, the illumination map can be enhanced, thereby balancing the overall illumination of the original low-light image. However, it is a typical ill-posed problem to obtain the reflectance and illumination from a single image. To address this issue, different prior-based or learning-based low-light enhancement methods have been proposed in recent literature. For example, LIME leverages a structure prior of the illumination map to refine the initial estimation \cite{guo2016lime} while a piece-wise smoothness constraint is used in \cite{li2018structure}. Since low-light images usually contain noises that will be amplified after enhancement, some robust Retinex models have been proposed to account for noise and estimate reflectance, illumination, and noise simultaneously \cite{li2018structure, zhu2020zero}.}

\rev{Images captured in a haze environment are in low contrast due to the haze attenuation and scattering effects. Recovering the clear image from a single hazy input is also an ill-posed problem, which can be addressed by both prior-based and learning-based methods \cite{he2010single,cai2016dehazenet,zhang2019famed}. For example, He et al. propose a dark channel prior to estimate the haze transmission efficiently \cite{he2010single}. Cai et al. propose the first deep CNN model for image dehazing, which outperforms traditional prior-based methods by leveraging the powerful representation capacity of CNNs \cite{cai2016dehazenet}. Recently, Zhao et al. propose a real-world benchmark to evaluate dehazing methods according to visibility and realness \cite{zhao2020dehazing}. When images are captured in the low-light and haze environment, it comes to the more challenging case, i.e., nighttime image dehazing. Similarly, some methods have been proposed based on either statistical priors or deep learning, e.g., maximum reflectance prior \cite{zhang2017fast}, glow separation \cite{li2015nighttime}, and ND-Net \cite{zhang2020nighttime}.}

\begin{table*}[htbp]
  \centering
  \caption{\rev{Representative benchmark datasets in auditory perception. ID: Identity; BBox: bounding box of speakers.}}
    \begin{tabular}{lllll}
    \toprule
    Area  & Dataset & Link  & Volume & Label Type \\
    \midrule
    \multicolumn{1}{l}{\multirow{3}[0]{*}{Speech Recognition}} & CHiME-3 & \url{http://spandh.dcs.shef.ac.uk/chime\_challenge/chime2015/} & 14,658 utterances & Text, ID \\
          & VoxCeleb & \url{http://www.robots.ox.ac.uk/~vgg/data/voxceleb/} & 1M utterances& Text, BBox, Attributes\\
          & TIMIT APCSC & \url{https://catalog.ldc.upenn.edu/LDC93S1} & 630 speakers & Text, ID \\
          \hline
    \multicolumn{1}{l}{\multirow{3}[0]{*}{Speaker Verification}} & VoxCeleb & \url{http://www.robots.ox.ac.uk/~vgg/data/voxceleb/} & 1M utterances & Text, BBox, Attributes\\
          & TIMIT APCSC & \url{https://catalog.ldc.upenn.edu/LDC93S1} & 630 speakers & Text, ID \\
          & Common Voice & \url{https://commonvoice.mozilla.org/en/datasets} & 61,528 voices & ID, Attributes \\
    \bottomrule
    \end{tabular}%
  \label{tab:datasetspeech}%
\end{table*}%

\subsubsection{Image Rectification and Stitching}
\label{subsubsec:imagerectification}
\rev{Wide Field-of-View (FOV) cameras such as fisheye cameras have been widely used in different AIoT applications, e.g., video surveillance and autonomous driving since they can capture a larger scene area than narrow FOV cameras. However, the captured images contain distortions since they break the perspective transformation assumption. To facilitate downstream tasks, the distorted image should be rectified beforehand. The rectification methods can be categorized as camera calibration-based methods and distortion model-based methods. The former calibrate the intrinsic and extrinsic parameters of cameras and then rectify the distorted image by following perspective transformation. The widely used calibration method is proposed by Z. Zhang in \cite{zhang2000flexible} based on planar patterns at a few different orientations, where radial lens distortion is modelled. The latter directly estimate the distortion parameters of a distortion model and map the distorted image to the rectified image based on it accordingly. Different geometric cues have been exploited for formulating the optimization constraints in optimization-based methods or loss functions in learning-based methods \cite{xue2019learning}, such as lines and vanishing points. Given two or more fisheye cameras with calibrated parameters, a panorama image can be obtained from their images by image stitching. For example, Liu et al. propose an online camera pose optimization method for the surround view system \cite{liu2019online}, which is composed of several fisheye cameras around the vehicle. The surround view system can capture a 360$^\circ$ view around the vehicle, which is useful in IoV for the advanced driver assistant system and crowd-sourcing high-precision map update.}

\textit{In addition to the above reviewed visual perception methods, we then present a brief review of auditory perception, specifically speech perception. We include two topics in the following part, i.e., speech recognition and speaker verification.}

\subsubsection{Speech Recognition}
\label{subsubsec:speech}
Speech recognition, a.k.a automatic speech recognition (ASR), is a sub-field of computational linguistics that aims to recognizing and translating spoken language into text automatically. Traditional ASR models are based on hand-crafted features like cepstral coefficient and Hidden Markov Model (HMM) \cite{bahl1986maximum}, which have been revolutionized by the deep neural network for end-to-end modeling without the need of domain knowledge for feature engineering, HMM design, as well as explicit dependency assumption. For example, RNN, especially Long Short-Term Memory (LSTM), is used to model the long-range dependencies in the speech sequence and decode the text sequentially \cite{sak2014long}. However, for one thing, extra effort is needed to pre-segment training sequences so that the classification loss can be calculated at each point in the sequence independently, for another, RNN processes data in a sequential manner, which is parallel-unfriendly. To address the first issue, the connectionist temporal classification is proposed by directly maximizing the probabilities of the correct label sequence in a differentiable way \cite{graves2006connectionist}. To mitigate the other issue, the Transformer architecture is devised using scaled dot-product attention and multi-head attention \cite{dong2018speech}. 

Recently, real-time ASR systems have been developed either using on-device computing or in a cloud-assisted manner \cite{chang2011csr,he2019streaming}. ASR is very useful in many AIoT applications since speech is one of the most important non-contact interaction modes. For example, ASR can be used in the smart input system \cite{longo2019ubiquitous}, automatic transcription system, smart voice assistant \cite{owens2010road,tashev2009commute}, computer-assisted speech rehabilitation and language teaching \cite{angelov2014speech}. The computing paradigm could be on-device edge computing (e.g., off-line mode of smart voice assistant), fog computing with a powerful computing device and sound pickup system (e.g., automatic transcription system for conferences), as well as the cloud computing with acceptable latency (e.g., on-line mode of smart voice assistant). Besides, some related techniques for music and humming recognition and birdsong recognition could be useful to empower AIoT systems for music retrieval and recommendation and wild bird conservation.

\subsubsection{Speaker Recognition}
\label{subsubsec:speaker}
While face recognition aims to recognize an individual through one's unique facial patterns, speaker recognition achieves the same goal using one's voice characteristics. A speaker recognition system is composed of three modules, i.e., speech acquisition and production, feature representation and selection, and pattern matching and classification \cite{campbell1997speaker}. Previously speaker recognition methods are dominated by the i-vector representation and probabilistic linear discriminant analysis framework \cite{dehak2010front}, where i-vector refers to extracting low-dimensional speaker embeddings from sufficient statistics. Recently, several end-to-end deep speaker recognition models have been devised \cite{chung2018voxceleb2}, achieving better performance than i-vector baselines. Similar to the techniques in face recognition, speaker recognition also benefits from the advances in deep metric learning, i.e., leveraging the contrastive loss or triplet loss to learn discriminative speaker embeddings from large-scale datasets. Speaker recognition is one of the important means for identity identification, which has many applications in various AIoT domains, for example, automatic transcription system for multi-person meetings, personalized recommendation by smart voice assistants \cite{feng2017continuous}, and audio forensics \cite{ross2020security}. Besides, speaker recognition can be integrated with face recognition for access control. \rev{The representative benchmark datasets in the areas related to auditory perception are summarized in Table~\ref{tab:datasetspeech}.}

\begin{table*}[htbp]
  \centering
  \caption{\rev{Representative benchmark datasets in natural image processing and multimedia analysis. I: Image; T: Text; V: Video; A: Audio.}}
    \begin{tabular}{lllll}
    \toprule
    Area  & Dataset & Link  & Volume & Label Type \\
    \midrule
    \multicolumn{1}{l}{\multirow{3}[0]{*}{Machine Translation}} & WMT   & \url{http://www.statmt.org/wmt14/translation-task.html} & 50M words & Text \\
          & NIST 2008 & \url{https://catalog.ldc.upenn.edu/LDC2011S08} & 942 hours & Text, Attributes \\
          & TedTalks & \url{http://opus.nlpl.eu/TedTalks.php} & 2.81M tokens & Text \\
          \hline
    \multicolumn{1}{l}{\multirow{3}[0]{*}{Text-to-Image}} & Caltech-UCSD Birds & \url{http://www.vision.caltech.edu/visipedia/CUB-200-2011.html} & 11,788 & Category, Text \\
          & COCO  & \url{https://cocodataset.org/\#captions-2015} & 123,287 & Caption \\
          & Oxford-102 Flowers & \url{http://www.robots.ox.ac.uk/~vgg/data/flowers/102/} & 8,189  & Category, Text \\
          \hline
    \multicolumn{1}{l}{\multirow{3}[0]{*}{Image Captioning}} & COCO  & \url{https://cocodataset.org/\#captions-2015} & 123,287 & Caption \\
          & nocaps & \url{https://nocaps.org/} & 15,100 & Caption, Category \\
          & Flickr30k & \url{http://shannon.cs.illinois.edu/DenotationGraph/} & 31,783 & Caption \\
          \hline
    \multicolumn{1}{l}{\multirow{2}[0]{*}{Coss-Media Retrieval}} & Wikipedia & \url{https://en.wikipedia.org/wiki/Wikipedia:Featured\_articles} & 2,866 & I-T pairs \\
          & PKU XMediaNet  & \url{http://59.108.48.34/tiki/XMediaNet/} & 40k & I-T-V-A pairs \\
    \bottomrule
    \end{tabular}%
  \label{tab:datasetmultimedia}%
\end{table*}%

\textit{Next, we present a review of the progress in natural language processing (taking machine translation as an example) and multimedia and multi-modal analysis.}

\subsubsection{Machine Translation}
\label{subsubsec:translation}
Machine translation (MT) is also a sub-field of computational linguistics that aims to translate text from one language to another automatically. Neural machine translation (NMT) based on deep learning has made rapid progress in recent years, outperforming the traditional statistical MT methods or example-based MT methods by leveraging the powerful representation capacity and large-scale training data. The prevalent architecture for NMT is the encoder-decoder \cite{cho2014properties}. Later, attention mechanism is used to attend to all source words (i.e., global attention) or only part of them (i.e., local attention) when decoding at each step of RNN \cite{bahdanau2015neural,wu2016google,vaswani2017attention}. Attention can be useful for learning context features related to the target and achieve joint alignment and translation, showing better performance for long sentences. Unsupervised representation learning has shown promising performance for many down-stream language tasks by learning context-aware and informative embeddings, e.g., BERT \cite{devlin2019bert}. Recently, unsupervised NMT has also been studied, which could be trained on monolingual corpora. For example, leveraging BERT as contextual embedding has been proved useful for NMT by borrowing informative context from the pre-trained model \cite{zhu2019incorporating}. Together with speech recognition and speech synthesis, MT can be extended to translation speech from one language to another, which is very useful in many AIoT applications such as language education \cite{angelov2014speech}, automatic translation and transcription, and multilingual customer service (e.g., subway broadcast).

\subsubsection{Multimedia and Multi-modal Analysis}
\label{subsubsec:multimedia}
With the rapid growth of multimedia content (e.g., text, audio, image, and video) created in various Internet platforms, understanding the content becomes a hot research topic. Recent studies on cross-media matching and retrieval try to align both domains semantically by leveraging deep learning, especially adversarial learning \cite{wang2017adversarial}. However, the modality-exclusive information impedes representation learning. To address this issue, disentangled representation learning has been proposed \cite{guo2019learning}, which tries to maximize the mutual information between feature embeddings from different modalities and separate modality-exclusive features from them. Image/video captioning and text-to-image generation are two generative tasks related to cross-modal matching, where captioning refers to generating a piece of text description for a given image or video \cite{hossain2019comprehensive} while text-to-image generation aims to generate a realistic image that matches the given text description \cite{qiao2019mirrorgan}. 

\rev{In addition to the aforementioned multimedia content, there are other modalities of data that are also useful for scene understanding, e.g., depth image, Lidar point cloud, thermal infrared image. By using them with RGB images as input, cross-modal perceiving has attracted increasing attention in real-world applications, e.g., scene parsing for autonomous driving \cite{li2017traffic,chen2019progressive}, object detection and tracking in low-light scenarios \cite{zhang2019rgb,wang2020cross}, and action recognition \cite{liu2018multi}. There are three ways of fusing multi-modal data, i.e., at the input level \cite{li2017traffic}, at the feature level \cite{wang2015large,xu2020outdoor,chen2019progressive,zhang2019rgb,wang2020cross}, and at the output level \cite{liu2018multi}, respectively. Among them, fusing multi-modal data at the feature level is most prevalent, which can be further categorized into three groups, i.e., early fusion \cite{xu2020outdoor}, late fusion \cite{wang2015large}, and fusion at multiple levels \cite{chen2019progressive,zhang2019rgb}. For example, a multi-branch group fusion module is proposed to fuse features from RGB and thermal infrared images at different levels in \cite{zhang2019rgb}, since the semantic information and visual details differ at different levels. Besides, the authors in \cite{chen2019progressive} leverage the residual learning idea to fuse the multi-level RGB image features and Lidar features via a residual structure in a cascaded manner.}

\rev{Multi-media generation and cross-modal analysis are useful in some AIoT applications, e.g., television program retrieval/recommendation based on speech description \cite{tashev2009commute}, automatic (personalized) item description generation in e-commerce, a teaching assistant in education, multimedia content understanding and responding in a chatbot, nighttime object detection and tracking for smart security, and action recognition for rehabilitation monitoring and assessment. Another research topic that is close to AIoT is multimedia coding, which has also been advanced by deep learning \cite{liu2020deep}. It is noteworthy that a novel idea named video coding for machines is proposed recently \cite{duan2020video}, which attempts to bridge the gap between feature coding for machine vision and video coding for human vision. It can facilitate down-stream tasks given the compact coded features as well as support human-in-the-loop inspection and intervention, therefore having vast potential for supporting many AIoT applications. The representative benchmark datasets in the areas related to natural language processing and multimedia analysis are summarized in Table~\ref{tab:datasetmultimedia}.}

\textit{\rev{In the end, we briefly review the progress in network compression and neural architecture search (NAS).}}

\subsubsection{Network Compression and NAS}
\label{subsubsec:nas}
\rev{Network compression is an effective technique to improve the efficiency of DNNs for AIoT applications with limited computational budgets. It mainly included four kinds of techniques, i.e., network pruning, network quantization, low-rank factorization, and knowledge distillation. Typically, network pruning consists of three stages: 1) training a large network; 2) pruning the network according to a certain criterion; and 3) retraining the pruned network. Network pruning can be carried out at different levels of granularity, e.g., weight pruning, neuron pruning, filter pruning, and channel pruning based on the magnitude of weights or responses calculated by $L_1/L_2$ norm \cite{he2017channel}. Network quantization compresses the original network by reducing the number of bits required for each weight, which significantly reduces memory use and float point operations with a slight loss of accuracy. Usually, uniform precision quantization is adopted inside the whole network, where all layers share the same bit-width. Recently, a mixed-precision model quantization method has been proposed by leveraging the power of NAS \cite{cai2020rethinking}, where different bit-widths are assigned to different layers/channels. For other techniques, we recommend the comprehensive review in \cite{cheng2018model}.}

\rev{Instead of manually designing the network, NAS aims to automatically search the architecture from a predefined search space \cite{elsken2019neural}. Most NAS methods fall into three categories, i.e., evolutionary methods, reinforcement learning-based methods, and gradient-based methods. Evolutionary methods need to train a population of neural network architectures, which are then evolved with recombination and mutation operations. Reinforcement learning-based methods model the architecture generation process as a Markov decision process, treat the validation accuracy of the sampled network architecture as the reward and update the architecture generation model (e.g. RNN controller) via RL algorithms. The above two kinds of methods require rewards/fitness from the sampled neural architecture, which usually leads to a prohibitive computational cost. By contrast, gradient-based methods adopt a continuous relaxation of the architecture representation. Therefore, the optimization of neural architecture can be conducted in a continuous space using gradient descent, which is orders of magnitude faster.}

\subsection{Learning}
\label{subsec:learning}
\begin{figure}
  \centering
  \includegraphics[width=\linewidth]{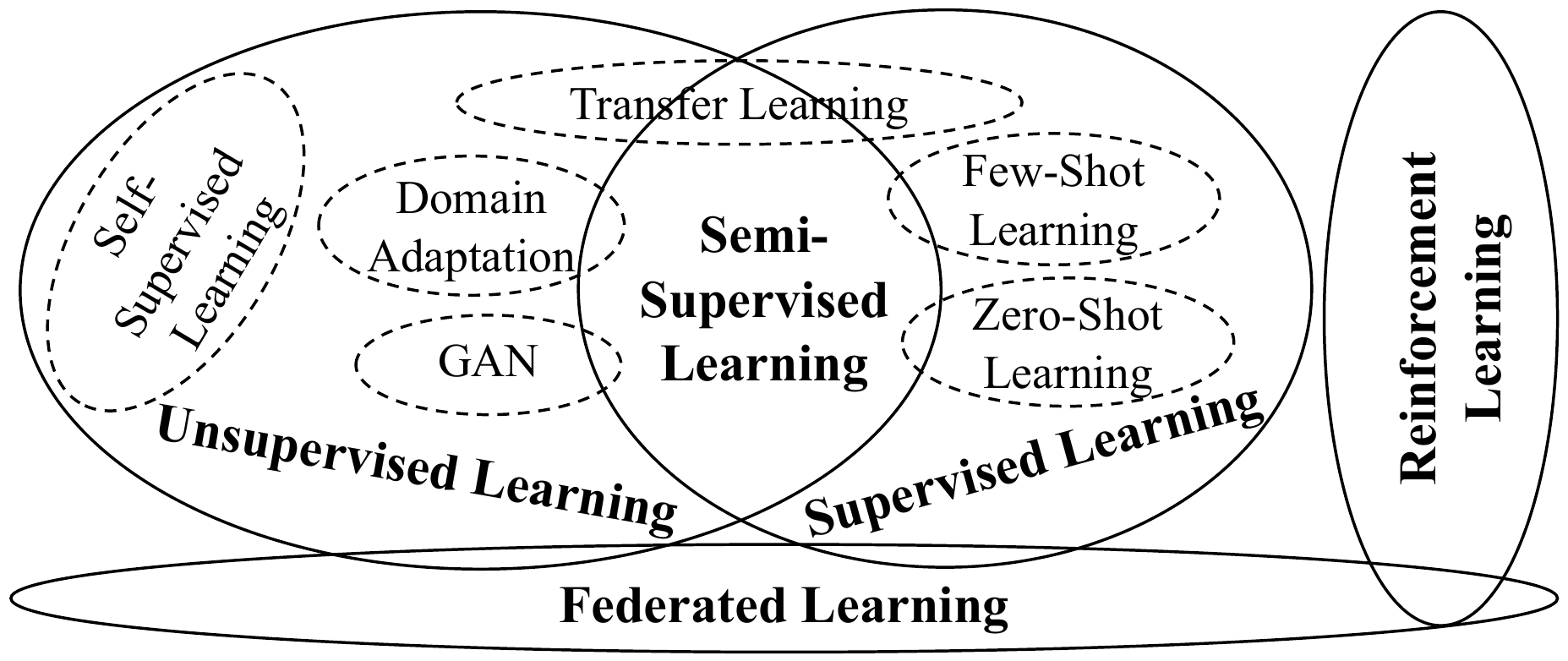}
  \caption{Diagram of the learning-related topics in AIoT.}
  \label{fig:learning}
\end{figure}

Since the real world is dynamic and complex, using a fixed model in AIoT systems can not adapt to the variations, probably leading to a performance loss. Thereby, empowering things with learning ability is important for AIoT so that it can update and evolve in response to the variations. Here, we briefly review the progress in several sub-areas of machine learning as diagrammed in Figure~\ref{fig:learning}.

\textit{First, we review some research topics in machine learning, where none or few data/annotations from the target task are available, i.e., unsupervised learning (USL), semi-supervised learning (SSL), transfer learning (TL), domain adaptation (DA), few-shot learning (FSL), and zero-shot learning (ZSL).}

\subsubsection{Unsupervised/Semi-supervised Learning}
\label{subsubsec:uslssl}
Deep unsupervised learning refers to learning from data without annotations based on deep neural networks, e.g., deep autoencoders, deep belief networks, and GAN, which can models the probability distribution of data. Recently, various GAN models have been proposed which can generate high-resolution and visually realistic images from random vectors. Accordingly, the models are expected to have learned a high-level understanding of the semantics of training data. For example, the recent BigBiGAN model can learn discriminative visual representation with good transferring performance on down-stream tasks, by devising an encoder to learn an inverse mapping from data to the latent space \cite{donahue2019large}. Another hot research sub-area is self-supervised learning, which learns discriminative visual representation by solving predefined pretext tasks \cite{jing2020self}. For example, the recently proposed SimCLR method defines a context-based contrasting task for self-supervised learning \cite{chen2020simple}, obtaining comparable performance as fully supervised models.

Semi-supervised learning refers to learning from both labeled and unlabeled data \cite{van2020survey}. Usually, the amount of unlabeled data is much larger than that of labeled data. Recent studies adopt a teacher-student training paradigm, i.e., pseudo-labels are generated by the teacher model on the unlabeled dataset, which is then combined with the labeled data and used to train or finetune the student model. For example, an iterative training scheme is proposed in \cite{xie2020self}, where the trained student model is used as the teacher model at the subsequent training round. The method outperforms the fully supervised counterpart on ImageNet by a large margin.

Since annotating large-scale data can be prohibitively expensive and time-consuming, USL and SSL can be useful for continually improving models in AIoT systems by harvesting the large-scale unlabeled data collected by massive numbers of sensors \cite{zhang2020detecting}. Besides, the multi-modal data from heterogeneous sensors (e.g., RGB/infrared/depth camera, IMU, Lidar, microphone) can be used to design cross modal-based pretext tasks (e.g., by leveraging audio-visual correspondence and ego-motion) and free semantic label-based pretext task (e.g., by leveraging depth estimation and semantic segmentation) for self-supervised learning \cite{saeed2020federated}.

\subsubsection{Transfer Learning and Domain Adaptation}
\label{subsubsec:tlda}
Transfer learning is a sub-field of machine learning, aiming to address the learning problem of a target task without sufficient training data by transferring the learned knowledge from a source related task \cite{zhuang2020comprehensive}. Note that different from the aforementioned semi-supervised learning where labeled and unlabeled data are usually drawn from the same distribution, transfer learning does not require the data distributions of the source and the target domains to be identical. For example, it has been almost the \textit{de facto} practice to fine-tune the models pre-trained on ImageNet in different down-stream tasks, e.g., object detection and semantic segmentation, for faster convergence and better generalization. In a recent study \cite{zamir2018taskonomy}, a computational taxonomic map is discovered for transfer learning between twenty-six visual tasks, providing valuable empirical insights, e.g., what tasks transfer well to other target tasks, and how to reuse supervision among related tasks to reduce the demand for labeled data while achieving same performance.

\begin{figure}
  \centering
  \includegraphics[width=0.9\linewidth]{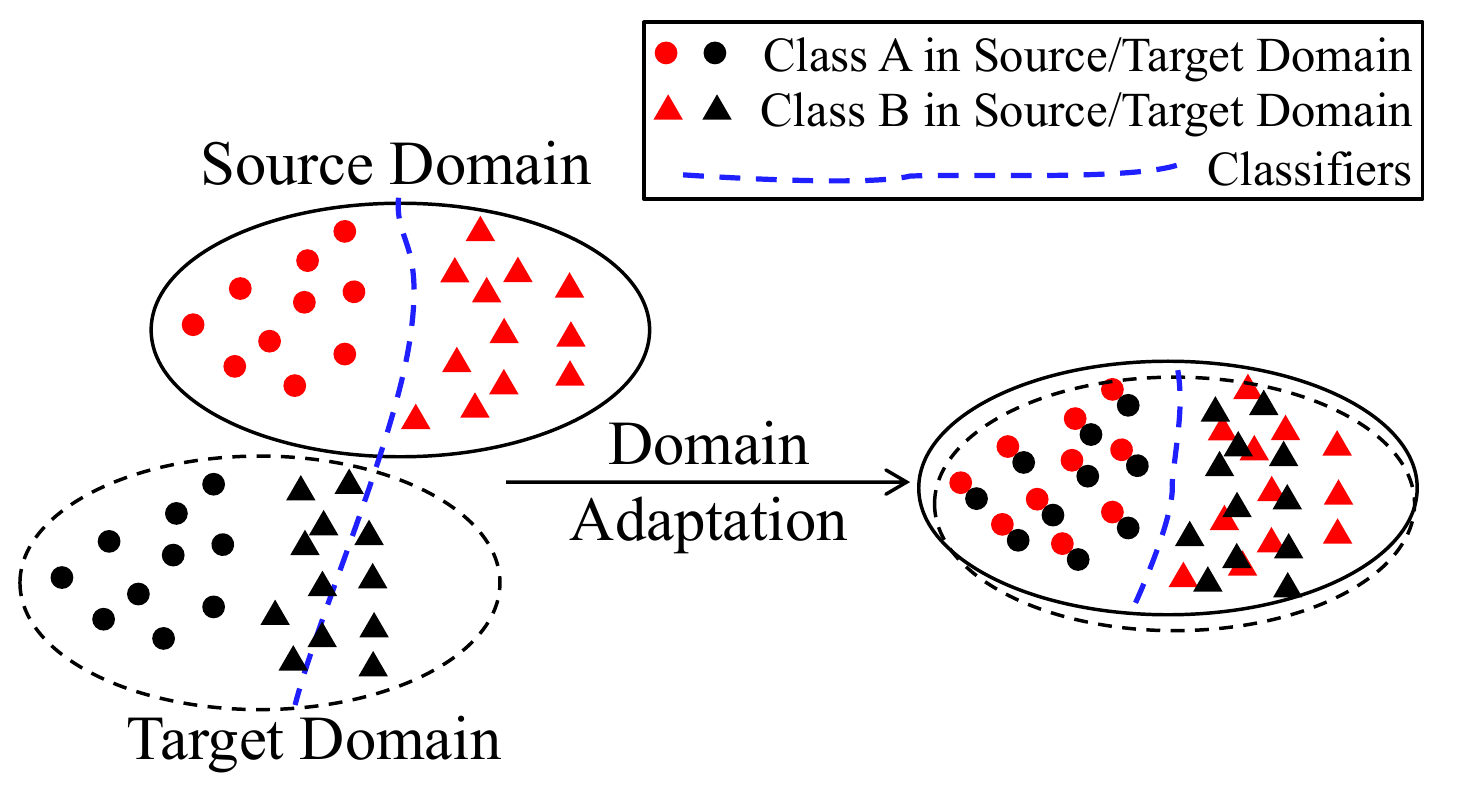}
  \caption{Illustration of domain adaptation.}
  \label{fig:da}
\end{figure}

Domain adaptation is also a long-standing research topic related to transfer learning, which aims to learn a model from one or multiple source domains that performs well on the target domain for the same task (Figure~\ref{fig:da}). When there are no annotations available in the target domain, this problem is also known as unsupervised domain adaptation (UDA). Visual domain adaptation methods try to learn domain-invariant representations by matching the distributions between source and target domains at the appearance level, feature level, or output level, thereby reducing the domain shift. Domain adaptation has been used in many computer vision tasks including classification, object detection, and especially semantic segmentation \cite{zhang2019category}, where obtaining the dense pixel-level annotations in the target domain is costly and time-consuming. Recently, a mobile domain adaptation framework is proposed for edge computing in AIoT \cite{yang2020mobileda} by knowledge distillation from the teacher model on the server to the student model on the edge device.

In real-world AIoT systems, there are always many related tasks involved, e.g., object detection and tracking, and semantic segmentation in video surveillance. Therefore, finding the transfer learning dependencies across these tasks and leveraging such prior knowledge to learn better models, are of practical value for AIoT \cite{liang2019federated,d2019transfer,bargoti2017deep,chen2017counting}. Domain adaptation could be useful for AIoT applications when deploying models to new scenarios or new working modes of machines \cite{wang2019learning,pan2017virtual,bousmalis2018using,valerio2019leaf,zhang2020cross}, e.g., ``synthetic$\rightarrow$real'', ``daytime$\rightarrow$nighttime'' or ``clear$\rightarrow$rainy''.

\subsubsection{Few-/Zero-shot Learning}
\label{subsubsec:fslzsl}
Few-shot learning, as an application of meta-learning (i.e., learning to learn), aims to learn from only a few samples with annotations \cite{wang2020generalizing}. Prior knowledge can be leveraged to facilitate addressing the unreliable empirical risk minimizer issue in FSL due to the small few-shot training set. For example, prior knowledge can be used to augment training data by transforming samples from the training set, or an extra weakly labeled/unlabeled dataset, or extra similar datasets. Besides, it can also be used to constrain hypothesis space and alter the search strategy in hypothesis space. In real-world AIoT applications, there are always some rare cases that need to be recognized by AI models, e.g., a car collision, cyber attack, machine fault. However, the collection and annotation of large-scale such cases are usually very difficult. Thereby, FSL can be used to learn suitable models in these scenarios \cite{passalis2020hypersphere}.

Zero-shot learning refers to learning a model with good generalization ability that can recognize unseen samples, whose classes have not been seen previously. Usually, auxiliary semantic information is provided to describe both seen and unseen classes, e.g., attributes-based description, text-based description. Thereby, each category can be represented as a feature vector in the attribute space or lexical space (a.k.a. semantic space). In some cases, the semantic space is a given learned semantic space, e.g., label embedding space or text-embedding space, where semantically similar classes are embedded as nearby vectors. Therefore, ZSL can be formulated as learning a mapping from the data space to the semantic space \cite{wang2019survey}. ZSL can be useful in some AIoT application scenarios. For example, in fault diagnosis \cite{feng2020fault}, some specific types of faults may occur in the future and should be recognized, but no training instances belonging to them have ever been collected previously. In other cases, classes may change over time or new classes may emerge, e.g., electric bicycle and tricar, where ZSL can be leveraged to adapt to the variations.  

\textit{Then, we review another two learning topics related to AIoT, i.e., reinforcement learning (RL) and federated learning (FL).}

\begin{figure}
  \centering
  \includegraphics[width=0.8\linewidth]{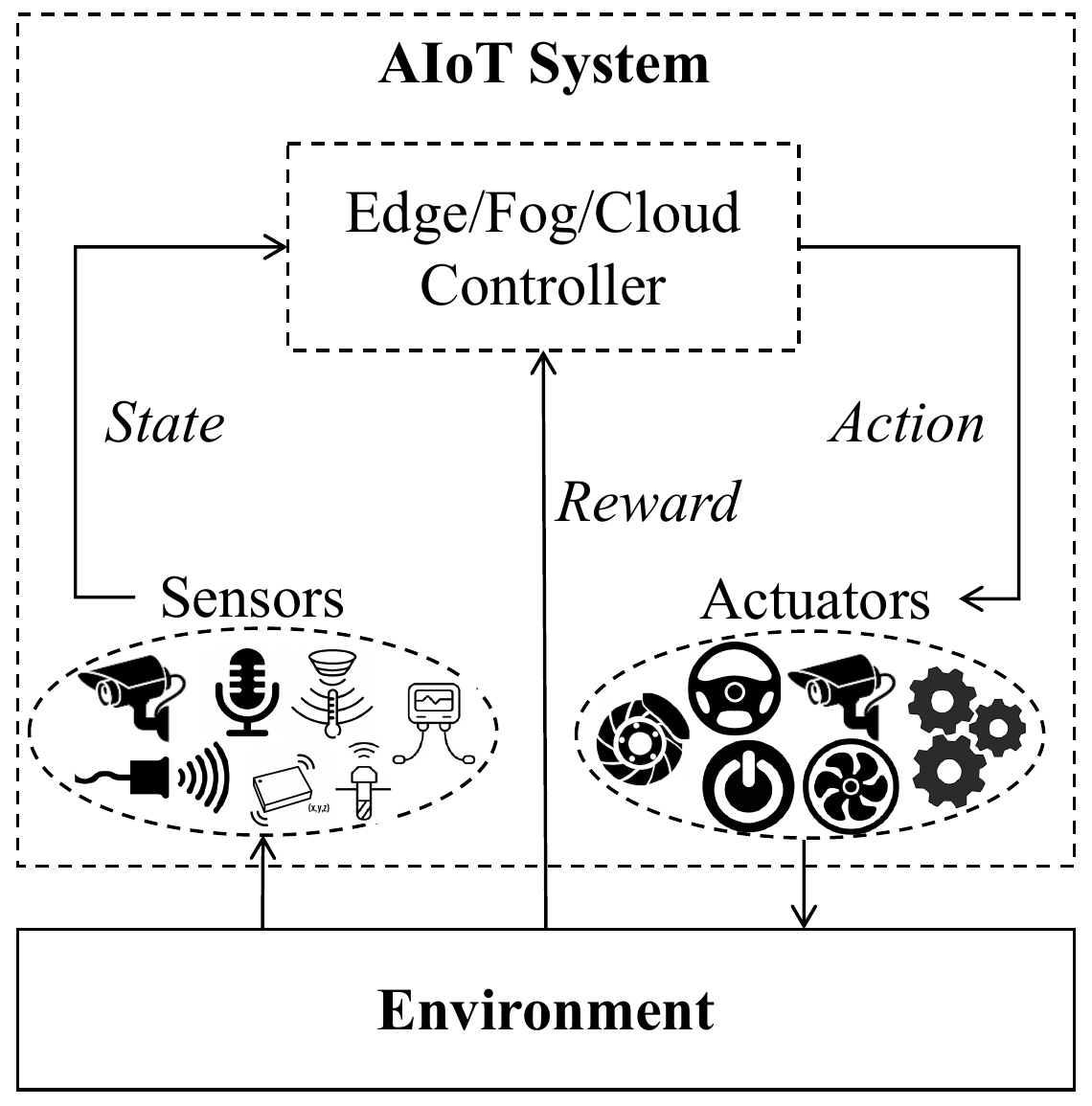}
  \caption{Illustration of reinforcement learning in AIoT.}
  \label{fig:rl}
\end{figure}

\subsubsection{Reinforcement Learning}
As one of the three basic learning paradigms along with supervised learning and unsupervised learning, reinforcement learning aims to learn a policy model for an agent through interactions with the environment, such that it can maximize the cumulative reward. Recently, rapid progress has been made by incorporating deep learning in RL (i.e., DRL) \cite{arulkumaran2017deep}. DNN has a strong representation capacity for learning compact and discriminative feature representation from the high-dimensional image and video data, which enables RL to deal with previously intractable problems by learning better policy and value function. There are two main groups of DRL methods including deep Q-network \cite{mnih2015human} and policy gradient-based methods such as asynchronous advantage actor-critic \cite{mnih2016asynchronous}. While actor-critic based methods directly optimize the cumulative reward, a surrogate objective function can be used to address the distribution shift problem in policy gradient based RL methods. 

DRL can empower things with the ability to interact with and adapt to the dynamic world, making it useful in many AIoT applications (Figure~\ref{fig:rl}), such as autonomous driving in smart transportation \cite{pan2017virtual,liang2019federated}, 3D-landmark detection of CT scans \cite{ghesu2017multi} and robot control \cite{thananjeyan2017multilateral} in smart healthcare, course recommendation in smart education \cite{zhang2019hierarchical}, real-time scheduling for smart factory \cite{shiue2018real}, load scheduling in smart grids \cite{wan2018model,chung2020distributed}, plant growth control in  smart agriculture \cite{somov2018pervasive,binas2019reinforcement}, and network management in smart cities \cite{zhao2019routing,hussain2020machine}. Moreover, DRL can be integrated into the FL framework for privacy-preserving learning \cite{liang2019federated,chung2020distributed}.

\subsubsection{Federated Learning}
\label{subsubsec:federatedlearning}
\begin{figure}
  \centering
  \includegraphics[width=0.9\linewidth]{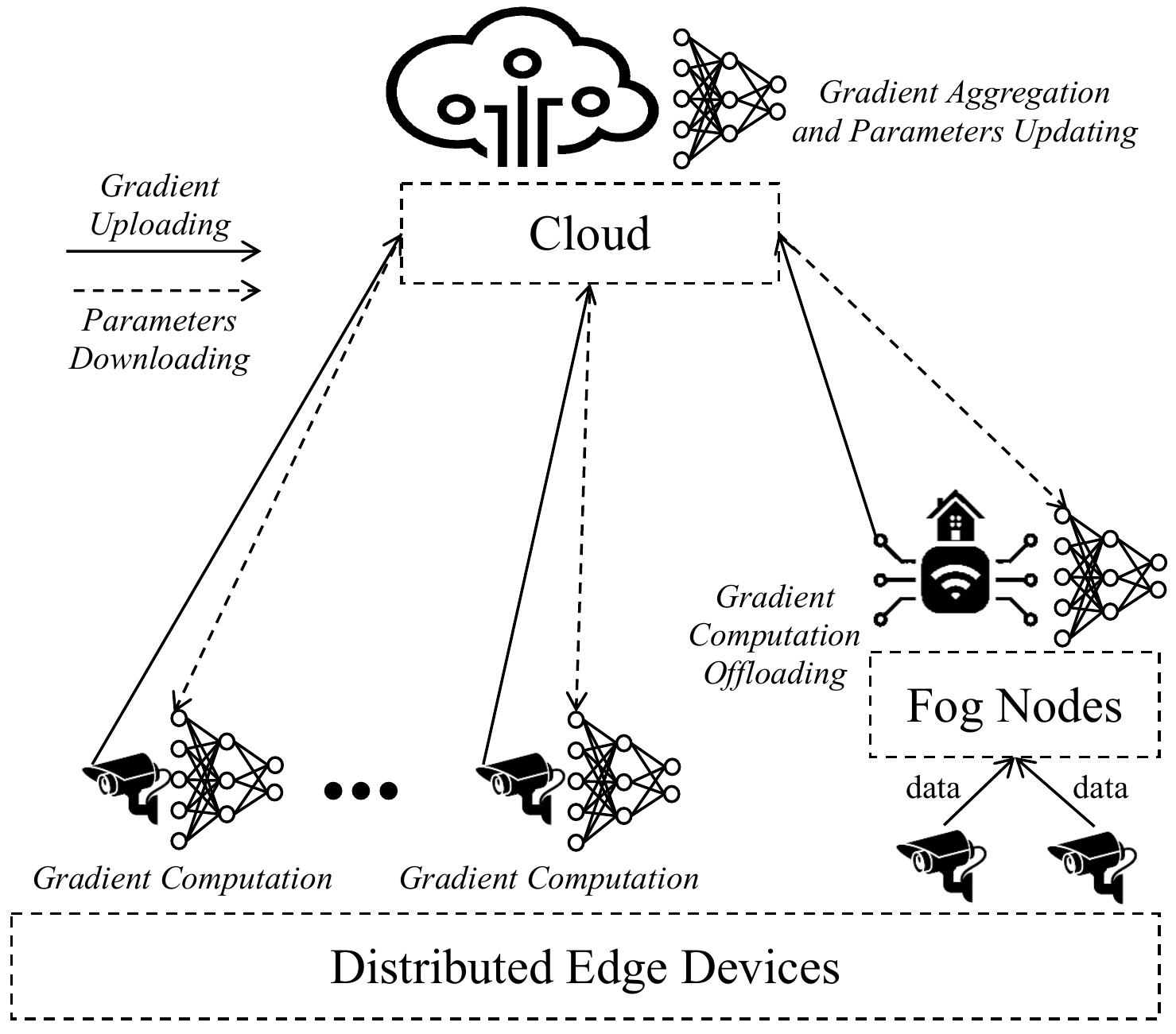}
  \caption{\rev{Illustration of federated learning in AIoT. Some edge devices offload gradient computation to fog nodes.}}
  \label{fig:fl}
\end{figure}

Federated learning was initially proposed to address the learning problem in which datasets are distributed across multiple devices and not allowed to be leaked to others \cite{konevcny2016federated}. Different data owners collaboratively train a model, whose performance is expected to match the performance of the model directly trained on the union of all data \cite{yang2019federated}. The architecture of FL usually consists of a central server (or collaborator) and many distributed client devices. Gradients are computed in each client using their own data and aggregated (or concatenated) in the server, and then sent back to clients to update their models (Figure~\ref{fig:fl}). FL offers a general learning framework to use massive distributed and isolated data while preserving data security and privacy, which is particularly appealing to AIoT applications in the context of edge computing \cite{wu2020personalized,saeed2020federated}. For example, FL can be used to improve the perceiving and learning ability of AIoT devices such as connected vehicles for autonomous driving \cite{liang2019federated} and wearable devices for health monitoring \cite{brisimi2018federated,saeed2020federated}.


\subsection{Reasoning}
\label{subsec:reansoning}
In our real-world, there is much knowledge carried by web information, medical records, financial transactions, etc., which can be used to reason the answer to a question or infer patient cohorts. We humans have the ability of causal reasoning such as causal inference and causal discovery. Empowering AIoT with such reasoning abilities is important for making smart and explainable decisions. In this part, we present the review on two related topics, i.e., knowledge graph and reasoning, as well as causal reasoning.

\subsubsection{Knowledge Graph and Reasoning}
\label{subsubsec:kg}
Knowledge graph (KG) is an efficient structured way to represent knowledge in a graph, where nodes represent entities, and edges represent relations (a.k.a. facts) in the forms of triples, i.e., (head entity, relation, tail entity). Some well-known KGs such as WordNet, Freebase, YAGO, NELL have been constructed and used in many applications via knowledge reasoning. Knowledge reasoning refers to inferring new knowledge based on the existing knowledge, e.g., identification and removal of erroneous knowledge, adding missing knowledge, answering questions, and drawing conclusions. Classical knowledge reasoning methods are based on rules, including first-order predicate logic rules, probability rules, ontology languages, and path rules. Recently, knowledge graph embedding based methods have attracted significant attention, aiming to embed the entities and relations in a KG into continuous vector spaces such that the reasoning can be done by leveraging translational distance models and semantic matching models \cite{wang2017knowledge}. 

Knowledge reasoning is useful for many down-stream tasks which can be categorized into in-KG and out-of-KG. In-KG applications include graph refinement (e.g., completion and error detection), triple/entity classification, and entity resolution. Out-of-KG applications include relation extraction, question answering, and recommendation, which are related to specific AIoT scenarios, such as network forensics analysis in smart security \cite{wang2005building}, smart assistants in smart healthcare and agriculture \cite{zhang2019multi,chen2019agrikg}, explainable recommendation in smart education and e-commerce \cite{bendakir2006using,wang2019explainable}, digital twins in smart factory \cite{banerjee2017generating,longo2019ubiquitous}, fault diagnosis in smart grids \cite{wang2015knowledge}. Note that one crucial feature of knowledge reasoning is explainable, which has the potential to be incorporated with deep learning and mitigate the interpretability issue in AIoT applications.

\subsubsection{Causal Reasoning}
Causality refers to the generic relationship between an effect and the cause, where the cause partly gives rise to the effect and the effect partly depends on the cause. Causal reasoning includes the topics of causal inference that aims to estimate the causal effect and causal discovery that aims to find causal relations. One classical way of reasoning causality is via randomized controlled trial by evaluating the outcomes from the treatment group and control group, which is however costly and time-consuming. Recently, learning causality from observational data has attracted much attention \cite{guo2020survey}. There are two well-known causal models used for learning causality, i.e., the structural causal models and the potential outcome framework (a.k.a. Rubin Causal Model). Different representation learning, multi-task learning, and meta-learning methods have been proposed for causal inference under the potential outcome framework.

Causal reasoning is useful in many AIoT applications such as online recommendation in smart e-learning \cite{bendakir2006using}, fault analysis in smart grids \cite{jiang2016spatial}, and driving safety in smart transportation \cite{may2009driver}. Moreover, causal inference can be used to unfold the ``black-box'' decision process of DNNs to address the interpretability issue, including model-based interpretation methods \cite{chattopadhyay2019neural} and example-based interpretation methods \cite{goyal2019counterfactual}. They are crucial for building explainable AIoT systems based DNNs, which have been widely used in many perceiving tasks as described in Section \ref{subsec:perceiving}.

\subsection{Behaving}
\label{subsec:behaving}
The behaving ability is also very important for AIoT systems passively responding to environment variations and raised requests, or actively exploring the unknown. Thereby, in this part, we present a brief review of two topics related to behaving in AIoT, i.e., control and interaction.

\subsubsection{Control}
\label{subsubsec:control}
The term of control here refers to controlling sensors and actuators in an AIoT system to transform the current system state to the target. The system state can be measured by sensors or calculated based on the perceiving methods described in Section \ref{subsec:perceiving}. The target state can be calculated based on predefined rules or determined by a decision model. The control algorithms are task-oriented and specifically designed in different AIoT systems regarding their physical structures and sensors. For example, there are two kinds of control in autonomous driving systems, i.e., lateral control (steering) for autonomic turning and lane-keeping, and longitudinal control (brake and throttle) for autonomic braking and forward collision avoidance. Multi-robot systems have been used in many AIoT applications such as smart logistics and precision agriculture, where decentralized control methods and coordination strategies are proposed for achieving and maintaining formations. Recently, deep reinforcement learning has been used for autonomous robot control, e.g., self-driving vehicle \cite{pan2017virtual,liang2019federated}, medical robot \cite{thananjeyan2017multilateral}, mobile service robot. In the public safety or traffic video surveillance scenarios, the active cameras can change its orientation and focal length by pan-tilt-zoom control to track and focus on specific targets \cite{denzler2003information,sommerlade2008information}.

\subsubsection{Interaction}
\label{subsubsec:interaction}
Real-world AIoT systems may interact with humans and the environment in different ways, which can be categorized into three groups: 1) using input and output devices for communication, e.g., keyboard, mouse, touch screen, microphone, and headset. 2) using mechanical arms for gripping and moving items, e.g., humanoid robots and industrial robots; 3) using gearing for moving and transportation, e.g., wheels of mobile robots. In the first category, AIoT systems may communicate with humans via multimedia, such as text, speech, image, and video. For example, AIoT systems can send a message to user-end or edge devices and display it on the screen as a reminder or a response to the user query. The message may contain text, audio, and video, which can be pre-recorded or searched from the database. Besides, the content can also be automatically created according to the user request. For example, in the dialogue system of a smart home assistant, the user may send 1) a picture or video to the assistant for generating a text description about it, which is related to image and video captioning \cite{hossain2019comprehensive}; 2) a keyword to the assistant for writing a poem \cite{zhang2014chinese}; 3) a piece of text description or a sketch to the assistant for creating an image \cite{qiao2019mirrorgan,qiao2019learn}; 4) lyrics and score to the assistant for composing a piece of music \cite{briot2017deep}. Speech is another way for AIoT systems communicating with humans, which is enabled by speech synthesis \cite{arik2017deep}, e.g., in question-answering system \cite{longo2019ubiquitous} and multilingual translator \cite{angelov2014speech}. Moreover, AIoT systems can create virtual 3D objects or scenes and render them on the AR/VR glasses or headsets. Human can interact with the virtual objects in the virtual environment or real objects in the augmented real environment.

In the second group, a service robot in the smart home scenario may use its arms with flexible joints to grip a cup and fill it with water for elderly people. Besides, robot arms in the scenario of the industrial internet of things (IIoT), can be used for assembling in industrial product lines \cite{krug2016next}. Note that human-robot interaction is a long-standing research topic that special efforts have been made in task planning and programming \cite{tsarouchi2016human}, safe interaction \cite{krug2016next}, and imitation learning \cite{zhang2018deep}. Recently, the idea of a digital twin has been proposed for the smart factory, various human interactions are enabled such as question-answering and voice control \cite{banerjee2017generating,longo2019ubiquitous}. In the third group, due to the simple kinematics, wheel-based mobile robots are widely used for SLAM and navigation to explore and interact with the unknown environment \cite{cheein2011optimized,shao2020tightly}. Note that there may be several interactive ways in an AIoT system upon the usage scenarios and devices.

\section{AIoT Applications}
\label{sec:applications}
\begin{figure}
  \centering
  \includegraphics[width=1\linewidth]{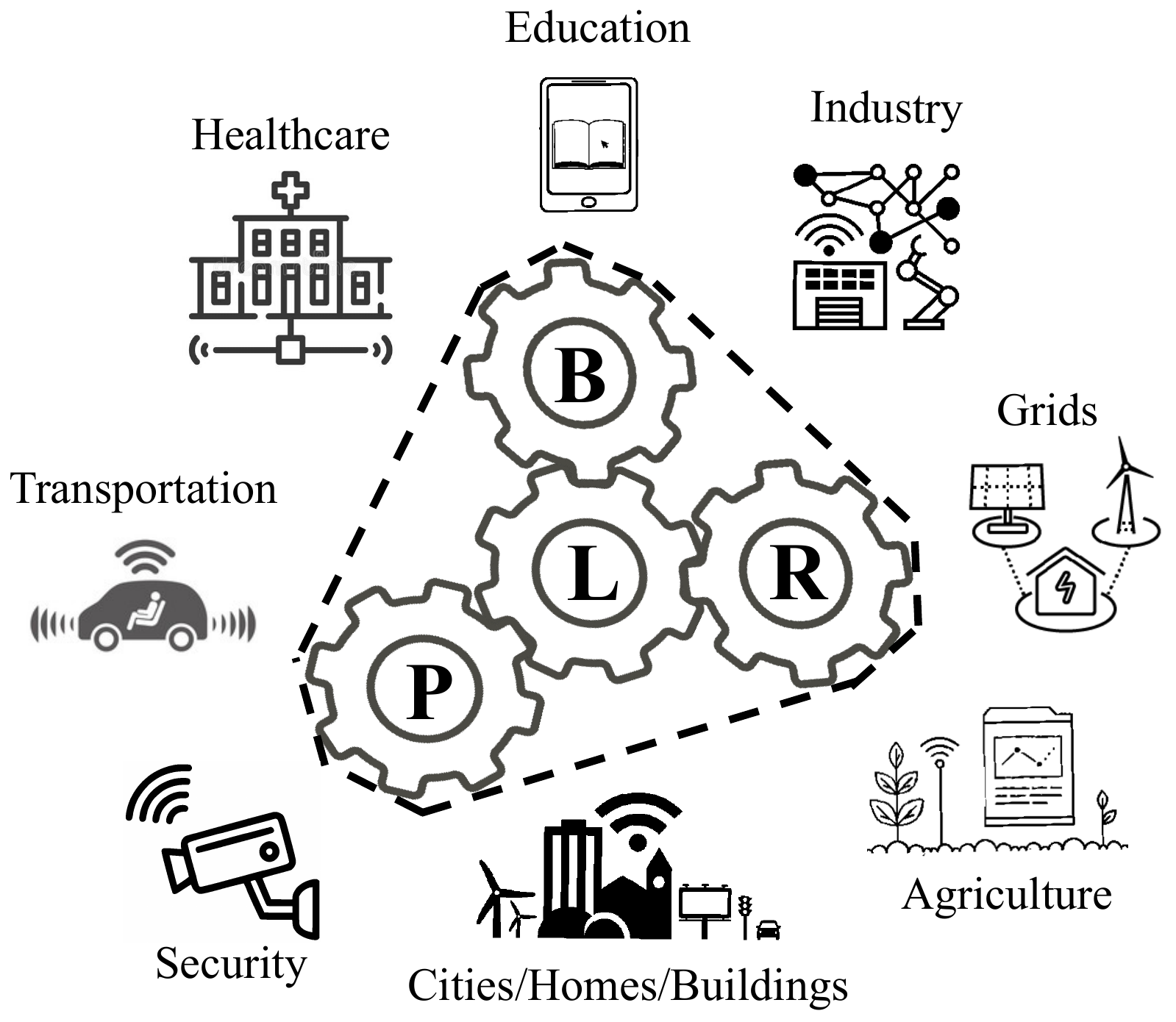}
  \caption{AI empowers things with the ability of perceiving (P), learning (L), reasoning (R), and behaving (B) in many AIoT domains. }
  \label{fig:application}
\end{figure}

As reviewed in Section \ref{sec:progress}, the progress of AI shows a great potential to empower the connected things in AIoT systems with the ability for perceiving, learning, reasoning, and behaving. The resulting AIoT systems will have a huge impact on the economic sectors and our living environments, such as security, transportation, healthcare, education, industry, energy, agriculture, as well as our homes and cities. In this part, we showcase some promising applications of AIoT in these areas (Figure~\ref{fig:application}) and demonstrate how will AI enable the AIoT systems to be faster, smarter, greener, and safer.

\subsection{Smart Security}
\label{subsec:security}
The goal of smart security is to ensure the security of our physical world and cyberspace, which can be achieved with the help of various AIoT systems. One of the most important features of them is \textit{human-centric perceiving}, which can recognize the identities of individuals and analyze their behaviors to prevent illegal activities. For example, face recognition systems have been deployed in building entrance, railway station, and airport, enabled by cloud/fog computing \cite{soyata2012cloud} or edge computing \cite{amos2016openface}. Despite their utility, one major concern is data security and privacy preservation. Recently, a lightweight solution is proposed in \cite{yang2019privacy} by block-based logic transformation, which not only reduces feature size but also preserves the original feature, therefore appealing to resource-limited AIoT devices and resistant to potential attacks. Beyond the biometric features from the face, fingerprint, and iris for recognition, spatial and temporal human body features (e.g., shape and gait) are leveraged for person re-identification, which aims to recognize individuals and trace their trajectories in multiple cameras. However, deploying the technique in a real-world scenario faces the domain shift challenge arisen from camera view differences (e.g., viewpoint, illumination, resolution). To address it, Zhang et al. propose a style translation-based method for cross-domain person re-identification in camera sensor networks \cite{zhang2020cross}, which can reduce domain shift and learn domain-invariant features. Besides, recognizing identities of human-related objects, e.g., vehicle license plate recognition  \cite{bulan2017segmentation}, is also useful for tracing human trajectory. Moreover, to capture salient and high-resolution humans and vehicles, active cameras can be used in AIoT systems by adjusting the orientation and focal length according to target location \cite{denzler2003information,sommerlade2008information}. Beyond the aforementioned techniques for individual analysis, estimating the crowd density and monitoring the crowd flow is also very important for public safety \cite{wang2019learning}, e.g., avoid deadly accidents of trampling and crushing of pedestrians. Some AIoT applications in different domains are summarized in Table \ref{tab:aiot_applications} according to the enabling AI technologies.

\begin{table*}[htbp]
  \centering
  \caption{AIoT applications empowered by different AI technologies.}
    \begin{tabular}{p{1.8cm}<{\raggedjustifyalign}p{6.7cm}<{\raggedjustifyalign}p{3.2cm}<{\raggedjustifyalign}p{1.8cm}<{\raggedjustifyalign}p{2.5cm}<{\raggedjustifyalign}}
    \toprule
     & Perceiving  & Learning  & Reasoning & Behaving \\
     \midrule
     Smart Security & object detection/tracking \cite{denzler2003information,sommerlade2008information,bulan2017segmentation}, text-spotting \cite{bulan2017segmentation}, biometric recognition \cite{soyata2012cloud,feng2017continuous,kumar2019intelligent,yang2019privacy,ross2020security}, person re-identification \cite{zhang2020cross}, pose/gesture/action recognition \cite{wu2020personalized}, crow counting \cite{schauer2014estimating,wang2019learning}, speech/speaker recognition \cite{feng2017continuous,ross2020security}  & TL \cite{wang2019learning}, DA \cite{wang2019learning,zhang2020cross}, FL \cite{wu2020personalized}    & KG reasoning \cite{wang2005building}  & control \cite{denzler2003information,sommerlade2008information}, interaction \cite{kumar2019intelligent}\\
     \midrule
    Smart Transportation & pose/gesture/action recognition \cite{ma2019trafficpredict,rasouli2019pie,kashevnik2019methodology}, object detection/tracking \cite{rasouli2019pie,kashevnik2019methodology,shao2020tightly}, semantic segmentation \cite{pan2017virtual,chen2019progressive}, 3D \cite{rasouli2019pie,kashevnik2019methodology,chen2019progressive,shao2020tightly}, text-spotting \cite{bulan2017segmentation}, speech/speaker recognition \cite{owens2010road,tashev2009commute,kashevnik2019methodology}, multimedia \cite{tashev2009commute}  & TL \cite{pan2017virtual,liang2019federated}, DA \cite{pan2017virtual}, RL \cite{pan2017virtual,liang2019federated}, FL \cite{liang2019federated} & causal reasoning \cite{may2009driver}  & control \cite{pan2017virtual,liang2019federated}, interaction \cite{may2009driver,tashev2009commute,owens2010road,ma2019trafficpredict,rasouli2019pie,kashevnik2019methodology,shao2020tightly} \\
     \midrule
    Smart Healthcare & object detection \cite{gonzalez2014kinect,ghesu2017multi,thananjeyan2017multilateral}, pose/gesture/action recognition \cite{gonzalez2014kinect}, 3D \cite{thananjeyan2017multilateral}, multimedia \cite{zhang2019multi}  & SSL \cite{saeed2020federated}, RL \cite{ghesu2017multi,thananjeyan2017multilateral}, FL \cite{saeed2020federated,brisimi2018federated} & KG reasoning \cite{zhang2019multi}  & control \cite{thananjeyan2017multilateral}, interaction \cite{zhang2019multi} \\
     \midrule
    Smart Education & image classification \cite{passalis2020hypersphere}, pose/gesture/action recognition \cite{starner1998real,he2020converting}, text-spotting \cite{he2020converting}, speech/speaker recognition \cite{angelov2014speech}, machine translation \cite{angelov2014speech}  & FSL \cite{passalis2020hypersphere}, RL \cite{zhang2019hierarchical}  & KG reasoning \cite{bendakir2006using}  & interaction \cite{angelov2014speech,brule2016mapsense,zhang2019hierarchical,he2020converting} \\
     \midrule
    Smart Industry & classification \cite{feng2020fault}, object detection/tracking \cite{krug2016next}, 3D \cite{krug2016next}, speech/speaker recognition \cite{longo2019ubiquitous}  & DA \cite{bousmalis2018using}, ZSL \cite{feng2020fault}, RL \cite{shiue2018real}  & KG reasoning \cite{banerjee2017generating,longo2019ubiquitous}  & control \cite{krug2016next}, interaction \cite{krug2016next,banerjee2017generating,bousmalis2018using,longo2019ubiquitous} \\
     \midrule
    Smart Grids & image classification \cite{hosseini2020intelligent}, object detection/tracking \cite{hosseini2020intelligent}, 3D \cite{senocak2018learning}  & USL \cite{chen2016detection,senocak2018learning}, SSL \cite{senocak2018learning,zhang2020detecting}, TL \cite{d2019transfer}, RL \cite{wan2018model,chung2020distributed}  & KG reasoning \cite{wang2015knowledge}, causal reasoning \cite{jiang2016spatial}  & control \cite{hosseini2020intelligent}, interaction \cite{senocak2018learning} \\
     \midrule
    Smart Agriculture & object detection/trakcing \cite{cheein2011optimized,bargoti2017deep,alsalam2017autonomous,liu2018robust}, counting \cite{chen2017counting,liu2018robust,valerio2019leaf}, semantic segmentation  \cite{chen2017counting,liu2018robust}, 3D \cite{cheein2011optimized,alsalam2017autonomous,liu2018robust,chebrolu2018robust}  & TL \cite{bargoti2017deep,chen2017counting}, DA \cite{valerio2019leaf}, RL \cite{somov2018pervasive,binas2019reinforcement}  & KG reasoning \cite{chen2019agrikg}  & control \cite{somov2018pervasive}, interaction \cite{cheein2011optimized,alsalam2017autonomous,chen2019agrikg} \\
     \midrule
    Smart Cities / Homes / Buildings & object detection/tracking \cite{li2011smart}, text spotting \cite{mancas2007multifunctional}, pose/gesture/action recognition \cite{amir2017low}, biometric recognition \cite{feng2017continuous}, speech/speaker recognition \cite{feng2017continuous,chang2011csr,he2019streaming}, 3D \cite{jain2015head,lee2016rgb,wang2017enabling,katzschmann2018safe} & RL \cite{mocanu2018line,zhao2019routing}  & KG reasoning \cite{le2016graph} & control \cite{pan2015internet}, interaction \cite{jain2015head,le2016graph,lee2016rgb,amir2017low,wang2017enabling,katzschmann2018safe} \\
    \bottomrule
    \end{tabular}%
  \label{tab:aiot_applications}%
\end{table*}%

\subsection{Smart Transportation}
\label{subsec:transportation}
Smart transportation enabled by AIoT covers traffic participants (e.g., smart internet of vehicles \cite{jiang2017novel}), traffic infrastructures \cite{jiang2018joint}, and industry applications (e.g., smart connected logistics \cite{lee2018design}). Among them, the self-driving car is a typical example empowered by AI, which integrates various perceiving, learning, reasoning, and behaving abilities together. The self-driving system should perceive the driving environment, such as detecting road \cite{chen2019progressive}, traffic-sign \cite{zhu2016traffic}, pedestrian \cite{chen2019shape}, and car \cite{hu2019joint}, estimating the intention of cars and pedestrians and predicting their trajectories \cite{ma2019trafficpredict,rasouli2019pie}. Besides, it should also measure the pose and location of landmarks (e.g., traffic-signs) for SLAM \cite{shao2020tightly}. Based on them, the self-driving system can determine its driving policy and interact with other traffic participants. Recently, deep reinforcement learning is leveraged to learn driving policy directly from visual input (e.g., front-view images). However, carrying out the training in the real-world is unaffordable. To address the issue, both domain adaptation \cite{pan2017virtual} and transfer learning methods \cite{liang2019federated} are proposed by leveraging virtual 3D game engines.

AIoT can also enable in-car driver monitoring and interactions with the infotainment system. Monitoring dangerous driver behaviors is crucial for preventing traffic accidents. To this end, a mobile application is developed in \cite{kashevnik2019methodology} for detecting dangerous driving behaviors, e.g., distraction and drowsiness. It leverages multi-model data for behavior detection including images for face state detection (e.g., eyes openness and head yaw angles), motion data from IMU sensors for estimating vehicle location and speed, light level for estimating lighting conditions, and speech data for speech state detection (e.g., speech rate and loudness). Moreover, the research in \cite{may2009driver} shows that identifying the causal factors of driving fatigue can be used for triggering corresponding countermeasures. For example, sleep-related fatigue is resistant to most interventions while task-related fatigue (e.g., distraction) can be counteracted effectively. Recently, AI interaction techniques such as gesture and speech recognition have been used for non-contact control in the in-car infotainment system \cite{tashev2009commute}. The study in \cite{owens2010road} shows that voice control allows drivers to keep watching out-car environment and demands lower mental load. By contrast, handheld controls may distract driver's attention and require higher mental demand, probably leading to dangerous driving behaviors.

\subsection{Smart Healthcare}
\label{subsec:healthcare}
AIoT systems for smart healthcare cover several phases including monitoring, examination, surgery, and rehabilitation. For monitoring, both wearable devices with motion sensors \cite{trung2016flexible} and cameras \cite{yang2016super} can be used for human activity recognition. Using motion sensors like accelerometer, human activities can be recognized from time-series motion data based on a CNN model. Recently, a mobile robot with a camera is used for human activity recognition, where the control policy is obtained using deep reinforcement learning by maximizing the recognition accuracy while minimizing its energy consumption. Note that the recognition module can be deployed in the edge devices or in the fog node depending on the model size and computational demand, which can be connected to the smart healthcare systems in community healthcare centers or hospitals. For examination, deep learning has been used for medical image understanding such as 3D-landmark detection in CT scans \cite{ghesu2017multi} and semantic segmentation \cite{ronneberger2015u}. Usually, these models are deployed on the private cloud of the hospital due to their high computational cost and the privacy concern. Recently, deep reinforcement learning has been used to control the surgical robot \cite{thananjeyan2017multilateral} for multilateral cutting in 2d orthotropic gauze. Furthermore, AIoT systems can be useful for various rehabilitation monitoring and assessment \cite{gonzalez2014kinect}, e.g., stroke rehabilitation and ankles rehabilitation. Via the connected 3D AR/VR devices, therapists can assess the rehabilitation and make better treatments accordingly, which is helpful for patients in rural or distant areas. Recently, online healthcare services or medical assistant robots can offer convenient information-query and auxiliary diagnosis services. For example, a hierarchical attention network is proposed to make explainable and accurate answers by exploiting the structural, linguistics, and visual information within a multi-modal medical knowledge graph \cite{zhang2019multi}. 

\subsection{Smart Education}
\label{subsec:education}
AI technologies can empower AIoT things to help children and students recognize new species, learn native or foreign languages, choose personalized learning resources, and help those with visual impairments learn through interactions. For example, a weight imprinting-based few-shot learning method is proposed in \cite{passalis2020hypersphere}, which can be used in edge devices to recognize new bird and animal species with only a few labeled samples. In \cite{he2020converting}, Raspberry Pi is used to build a smart classroom by leveraging hand gesture recognition and text recognition technologies. This AIoT system enables to control Raspberry Pi to capture the lecture notes in blackboard/whiteboard via static hand gesture, recognize the text in the image (e.g., numbers, characters, symbols), and convert them into an editable format, which is then saved in private cloud and shared to students for further editing or collaboration via the desktop application. A speech-to-speech multilingual translation system on mobile devices are proposed in \cite{angelov2014speech}, which includes three modules, i.e., speech recognition, language translation, and text-to-speech synthesis. It can work in offline mode and offers grammatical information that is useful for language learners. Recently, many mobile translator products have been released in CES 2020, which can translate tens of languages, benefiting from the advances of AI technologies like deep learning. As a complement of on-campus learning, online learning via Massive Open Online Course (MOOC) platforms (e.g., Coursera) has become very popular. Learners watch the courses via their end-devices connected to the cloud. Recommending courses from massive numbers of candidates could be helpful for personalized learning. To this end, deep reinforcement learning-based \cite{zhang2019hierarchical} and rule-based \cite{bendakir2006using} recommendation methods have been proposed. Besides, helping children with visual impairments in education also matters. For example, multi-sensory interactive maps are designed \cite{brule2016mapsense}, which can help children to acquire skills via interactions, e.g., touching, listening, tasting, and scenting.

\subsection{Smart Industry}
\label{subsec:industry}
\begin{figure}
  \centering
  \includegraphics[width=1\linewidth]{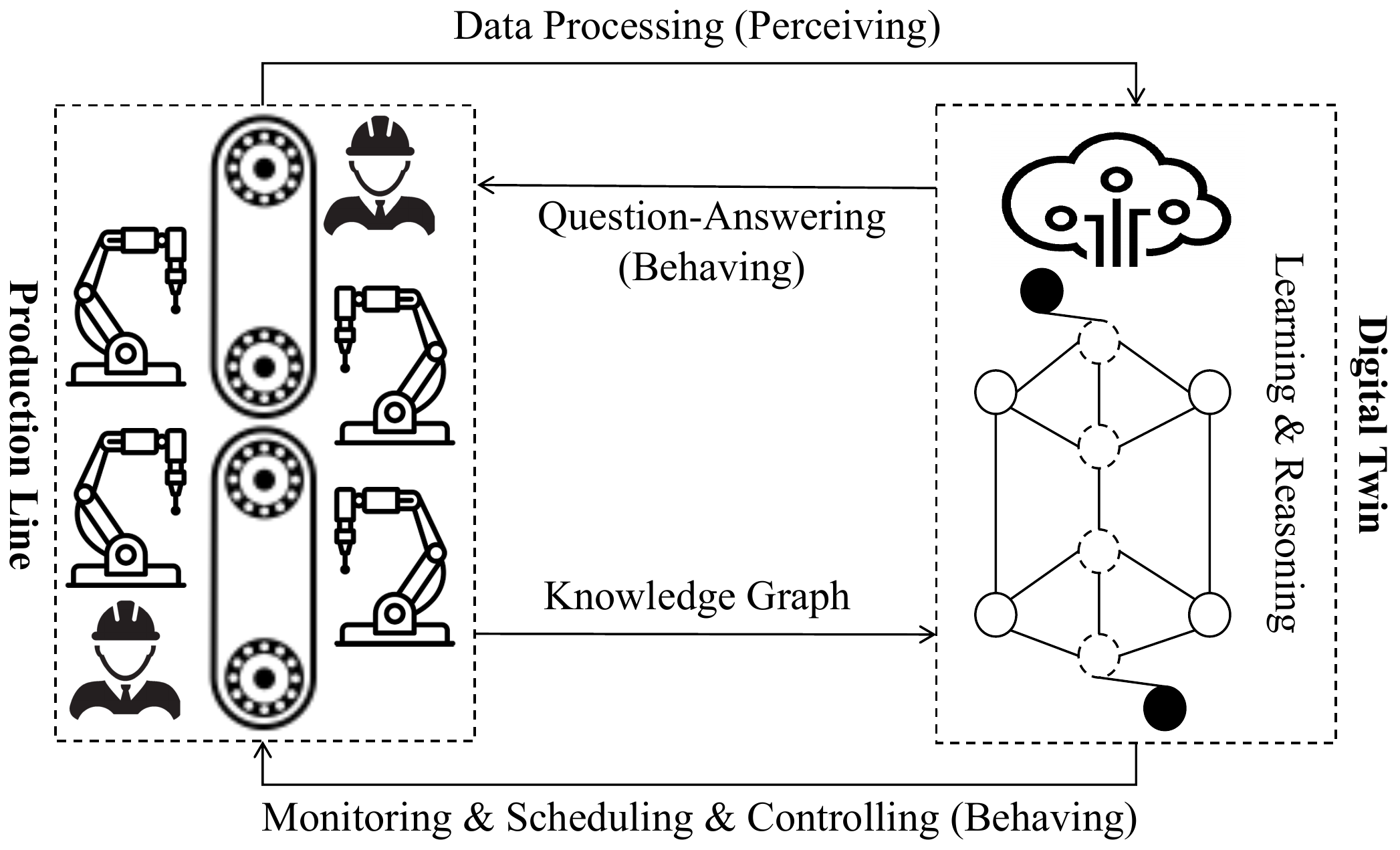}
  \caption{Diagram of the digital twin system in a smart factory.}
  \label{fig:digtaltwin}
\end{figure}

Digital Twin, i.e., a mirror digital representation of a physical system, has demonstrated great value for smart factories in Industry 4.0, e.g., monitoring the manufacturing process, diagnosing the fault, and preventing downtime. AIoT can be a critical part of implementing digital twins where the connected sensors and actuators can collect real-time data from production lines and send them to the digital twin running in the cloud (Figure~\ref{fig:digtaltwin}). Moreover, AI technologies can enable an intelligent analysis of data and help to make smart decisions. Recently, a service-oriented digital twin model is proposed in \cite{longo2019ubiquitous}, which uses an ontology-oriented knowledge structure to represent the knowledge about the manufacturing system from the sensing data. It also designs a vocal interaction system for knowledge retrieval based on speech recognition and text-to-speech synthesis. In \cite{banerjee2017generating}, a knowledge graph-based digital twin model is introduced which is composed of four parts, i.e., feature extraction, ontology creation, knowledge graph generation, and semantic relation extraction. It can extract and infer knowledge from large scale production line data and enhance manufacturing process management via semantic relation reasoning. Real-time scheduling (RTS) in the smart factory is another hot research topic. In \cite{shiue2018real}, a reinforcement learning-based RTS model is proposed, which can incrementally update and maintain the knowledge base in RTS during operations to respond to shop floor environment change.

\rev{A typical example of AIoT application in the smart industry is the Printed Circuit Board (PCB) manufacturing. There are three scenarios that are related to AIoT systems with different sensors and devices, i.e., manufacturing, visual defect inspection, and machine fault diagnosis. First, industrial robots have been widely used in the production line of smart factories, e.g., for drilling and grasping. AI technologies can be used to improve their functionalities. For example, Bousmalis et al. propose a deep robotic grasping model named GraspGAN \cite{bousmalis2018using}, which bridges the domain gap between synthetic images and real-world ones via the pixel-level image translation and a feature-level domain classifier. To increase the safety, speed, and accuracy of autonomous picking and palletizing, Krug et al. propose a novel grasp representation scheme allowing redundancy in the gripper pose placement \cite{krug2016next}. Second, PCB defect inspection carried out by workers manually is laborious and time-consuming. Recently, deep learning-based methods have been proposed for automatic real-time visual defect inspection \cite{gao2020real}. Third, it is important to predict and diagnose machine faults from sensor data to reduce PCB defects, thereby increasing production efficiency and reducing losses. Although the digital twin system provides a useful mirror virtual environment for creating and testing new equipment and models, it is still challenging to fast adapt the trained model or control policy to the physical world. Thereby, more efforts should be made in the areas of domain adaptation, transfer learning, and mete-learning. Besides, since it is difficult to collect and annotate edge samples in the industrial context, zero-/few-shot learning is also worth further study. In addition, causal analysis of the product defects based on data and knowledge is also of practical importance.}


\subsection{Smart Grids}
\label{subsec:grid}
AIoT in the smart grids can be used for grid fault diagnosis, load monitoring and scheduling, and cyber-attack detection. For example, unmanned aerial vehicles (UAVs) connected to the control center via the cellular network are used for damage classification and estimation of power distribution poles \cite{hosseini2020intelligent}. The captured image by UAVs is sent back to the cloud and processed by a CNN model to predict falling damage extent and fire damage extent. Besides, some ``industrial stethoscopes'' are designed to recognize and localize fault sound sources in visual scenes via cameras and multiple microphones, where the algorithm can run on edge devices for real-time monitoring. Recently, a two-stream network with an attention mechanism is proposed to directly localize a sound source in images \cite{senocak2018learning}. AI technologies have also advanced fault diagnosis. For example, a convolutional sparse autoencoder-based unsupervised learning method has been proposed for power transmission line fault diagnosis based on voltage and current signals \cite{chen2016detection}. It can detect faults within 7 ms latency, which is appealing to real-world applications. Besides, knowledge representation \cite{wang2015knowledge} and causal relationship discovery \cite{jiang2016spatial} are also explored for power grid fault diagnosis and impact causal analysis. For load monitoring and electric vehicle charging scheduling, transfer learning \cite{d2019transfer} and deep reinforcement learning \cite{wan2018model} methods are proposed. Furthermore, to preserve the privacy of the households while intelligently managing their load scheduling, a distributed deep reinforcement learning method is proposed inspired by the idea of federated learning \cite{chung2020distributed}, where the action networks are located at distributed households and the critic network is located at an aggregator from a trusted third party. The security concern about the smart grids under cyber-attacks has attracted much attention. Recently, a semi-supervised deep learning method is proposed based on autoencoder and generative adversarial networks, which can effectively detect false data injection attacks in smart grids. 

\subsection{Smart Agriculture}
\label{subsec:agriculture}
Recently, the concept of precision agriculture becomes popular, referring to observing, measuring, and responding to crop variability through sensors, autonomous agricultural machines, and geographic information systems, which can be achieved by AIoT in smart agriculture. For example, crop counting and yield estimation is an important topic of precision agriculture \cite{chen2017counting,liu2018robust}. UVAs are used for capturing images of crops and fruits and sending them to the cloud for further counting \cite{chen2017counting}. Note that accumulating the counting results across image frames does not lead to the total yield since fruits are double-counted in adjacent frames. To address this issue, a detection-tracking-counting based method is proposed \cite{liu2018robust}, which can reject outliers and double-counted fruits. UAVs can also be used for continuous crop monitoring by capturing crop field images over time and temporally aligning them \cite{chebrolu2018robust}. The path planning \cite{alsalam2017autonomous} and self-localization and navigation \cite{cheein2011optimized} abilities of UAVs are critical to complete the above tasks, which have been studied in the agricultural scenario. For plant grow control, an AIoT system is set up in a tomato greenhouse \cite{somov2018pervasive}, which is composed of a wireless sensor network and cloud computing center empowered by AI technologies. Deep reinforcement learning is used for obtaining the optimal control policy on the illumination, nutrition, and ventilation conditions in the greenhouse. For crop management and pest/disease control, agricultural knowledge retrieval and answering system will be helpful. To this end, an agricultural knowledge graph, namely AgriKG \cite{chen2019agrikg}, is automatically constructed by leveraging NLP and deep learning techniques.

\subsection{Smart Cities/Homes/Buildings}
Smart cities/homes/buildings AIoT is related to those in the aforementioned smart sectors and can be empowered by similar AI technologies. For example, 1) continuous speaker authentication \cite{feng2017continuous} in smart home voice assistants by integrating body-surface vibration signals with speech signal, is related to smart security (Section~\ref{subsec:security}); 2) human-machine interaction systems (e.g., control television) based on hand gesture recognition \cite{rautaray2015vision,amir2017low} in smart homes, share the similar techniques for traffic sign language recognition in smart transportation (Section~\ref{subsec:transportation}) and sign language recognition in smart education (Section~\ref{subsec:education}). Recently, an event-based gesture recognition system is proposed by using dynamic vision sensors and a neurosynaptic event-based processor \cite{amir2017low}, which can identify gestures with a low latency energy consumption; 3) visual-navigation systems for the visually impaired \cite{lee2016rgb,wang2017enabling,katzschmann2018safe}, share some basic features like localization, obstacle recognition, path planning of SLAM for autonomous driving in smart transportation and smart agriculture (Section \ref{subsec:transportation} and Section~\ref{subsec:agriculture}), but also have some differences. For example, these systems usually contain some feedback modules (e.g., via vibration or speech) specifically designed for the visually impaired; 4) sound visualization systems for the hearing impaired are related to the industrial stethoscopes for localizing fault sound sources in the grid (Section~\ref{subsec:grid}), but also offer different features. For example, the gender of speakers, sound types, loudness, sound directions are displayed as indicator icons \cite{jain2015head}, specifically designed for the hearing impaired; 5) the energy management and optimization systems in smart buildings \cite{pan2015internet,mocanu2018line} can be regarded as the extension of smart grids AIoT (Section~\ref{subsec:grid}).

\section{Challenges and Opportunities}
\label{sec:challengesOpportunities}
\subsection{Challenges}

\subsubsection{Multi-modal Heterogeneous Data Processing, Transmission, and Storage}
AIoT systems contain massive numbers of heterogeneous sensors that generate a large data stream of different formats, sizes, and timestamps, thereby significantly challenging further processing, transmission, and storage. An efficient coding scheme can be used to reduce network bandwidth and transmission latency. For example, the video coding scheme for machines \cite{duan2020video} is promising for facilitating downstream computer vision tasks and requires further study. The AI perceiving technologies mentioned in Section~\ref{subsec:perceiving} could be used to extract compact and structural representations from the data, which would be transmission and storage friendly. However, the structural representation is task-oriented and should be calculated at the network edge to minimize bandwidth and latency, which represents another challenge.

\subsubsection{Deep Learning on Edge Devices}
Deploying deep CNN models on edge devices is crucial for real-time data stream processing and low latency in AIoT systems. However, edge devices are limited by their limited computational and storage resources. Thereby, how to design or automatically search lightweight, computationally efficient, and hardware-friendly DNN architectures is of practical value but still remains challenging. Moreover, network pruning, compression, and quantization are also worth further exploring.

\subsubsection{Computational Scheduling in the AIoT Architecture}
As described in Section~\ref{subsec:tritierArch}, a typical AIoT architecture contains heterogeneous computing resources including cloud centers, fog nodes, and edge devices. In real-world AIoT systems, some intense computation may be needed to offload to the fog node or cloud center from the edge devices, thereby creating a computational scheduling challenge. Specifically, when scheduling computation across different resources, the following factors should be taken into account: data type and volume, network bandwidth, processing latency, performance accuracy, energy consumption, and data security and privacy in each specific application scenario. Moreover, a dynamic adaptive scheduling strategy would deal with unbalanced data flow and user demands over time.

\subsubsection{Big and Small Data for Deep Learning in AIoT}
Big data generated from massive numbers of sensors are ubiquitous in AIoT systems, with huge potential for deep learning. As reviewed in Section~\ref{subsec:perceiving}, deep supervised learning methods have achieved remarkable success for perceiving in different areas due to large-scale labeled data. However, most AIoT data are unlabeled, and labeling them would be both time and financially expensive. Although there has been rapid progress in unsupervised learning, especially self-supervised learning \cite{chen2020simple,he2020momentum,jing2020self}, future efforts are expected to further leverage AIoT data, especially multi-modal data. Furthermore, given the small scale of labeled data, transfer learning, semi-supervised learning, and few-shot learning in the AIoT context might provide solutions to challenges from new classes, rare cases, and state drifting of devices.

\subsubsection{Data Monopoly}
In the AI era, data provide a valuable resource for creating new products and improving services. AIoT companies collect and exploit massive data, thereby leading to new opportunities for data collection and exploitation. This positive loop could lead to a data monopoly, i.e., vast proprietary data protected by established interests that cannot be accessed by other entities. Consequently, new competitors face a \textit{de facto} barrier to market entry, and a data monopoly becomes a real threat to free-market competition.

\subsubsection{Data Security and Privacy}
Due to sensors being ubiquitous in smart homes, hospitals, and cities, vast biometric data (e.g., face image, voice, action, pulse, imaging data, etc.) of AIoT users or informed and uninformed participants may be collected. This raises important concerns with regards to data security and privacy. Who owns these data? How long will these data be retained? How will these data be used? Legislation is important in response to these concerns, a good example being the General Data Protection Regulation (GDPR)\footnote{\url{https://eur-lex.europa.eu/eli/reg/2016/679/oj}} enforced by the European Union, which gives individuals control over their personal data. Controllers and processors of personal data must take appropriate measures to protect data security and privacy.

\subsubsection{Growing Energy Consumption in Data Centers}
According to \cite{andrae2015global}, electricity use by communication technologies is expected to account for 21\% of global total usage, with data centers contributing more than 1/3 to this. Therefore, enhancing energy efficiency in data centers is required for a sustainable future. For example, some data centers are located in cold climates to take advantage of air cooling. Other solutions include water cooling and immersing servers in a non-conductive oil or mineral bath. Workload analysis, task scheduling, and virtual machine consolidation have also been studied to improve power efficiency in data centers \cite{verma2009server,yuan2020biobjective}. The rapidly growing number of cloud centers mirrors the rapid growth in AIoT applications. Consequently, more efforts should continue to be made to address energy consumption in data centers.

\subsection{Opportunities}

\subsubsection{Built-in Neural Processing Capacity for Edge Devices}
Many edge devices are equipped with specialized chips (e.g., GPUs in smartphones and intelligent cameras) to accelerate neural network processing. Consequently, building neural processing capacity into edge devices is very useful for AIoT applications. First, it reduces processing latency and network bandwidth consumption. Since the sensing data can be processed on-site, only a small amount of processed data needs to be transmitted. Second, it can protect data security and privacy. For example, for biometric verification, registered user biometric data could be stored with encryption on local hardware, with only the built-in verification capacity on the edge devices exposed to the applications, thereby reducing the risk of data leakage. Third, it enables distributed and asymmetric model training. A federated learning framework can be used to train models on distributed edge devices by leveraging their local sensor data. Moreover, some groups of devices may choose different model updating policies than others depending on their usage scenarios.

\subsubsection{Event-based Sensors and Neuromorphic Processors}
Traditional camera sensors continually generate dense data once opened, which are further fed into deep CNNs for further GPU processing. Generally, all the pixels are used in the calculation, resulting in high computational costs. Recently, event-based sensors and neuromorphic processors have been proposed \cite{amir2017low}. For example, event-based cameras only record pixels that change in brightness, thereby reducing redundant data generation and transmission. Event-based neuromorphic processors can operate on sparse and asynchronous event streams directly, avoiding dense and redundant computations on regular sensing data like GPUs. These can be used in many AIoT applications such as gesture/action recognition with low power consumption and latency.

\subsubsection{Deep Learning from the Virtual to the Real}
In embodied AI, it is difficult and/or costly to train models in the real world, e.g., autonomous driving, robot arm control, and robot navigation. 3D virtual platforms that mimic real-world scenarios have been proposed such as Voyage Deepdrive\footnote{\url{https://deepdrive.voyage.auto/}}, OpenAI gym\footnote{\url{https://gym.openai.com/}}, and Habitat\footnote{\url{https://aihabitat.org/}} which are very useful for cost-effectively training deep learning models, especially for deep reinforcement learning. Nevertheless, the critical issue of domain shift between the virtual and physical environments must be addressed before deploying the trained model into real-world scenarios. Recently, transfer learning and domain adaptation have attracted significant attention to address this issue in the setting of both unsupervised learning and reinforcement learning.

\subsubsection{Data and Knowledge Integration for Perceiving, Learning, Reasoning, and Behaving}
Deep learning model performance is largely determined by large-scale training data. However, humans learn new concepts based not only on data but on prior knowledge. Likewise, prior knowledge can be very useful for training deep learning models in a data-efficient way. For example, attribute-based class description enables zero-shot learning for new concepts via attribute transfer. Another example is knowledge graphs, which represent structural relationships between entities. Knowledge can be extracted from unstructured data to build knowledge graphs, learn knowledge-embedding representations, and for reasoning. Integrated with deep learning (e.g., graph neural networks), this is a useful approach in many areas such as question and answer systems and fault/disease diagnosis, opening up promising research avenues towards human-level cognition intelligence. Therefore, data and knowledge integration are important for improving the perceiving, learning, reasoning, and behavior of AIoT.

\subsubsection{Privacy-preserving Deep Learning}
Deep learning requires large-scale data generated by different things from different users in the context of AIoT. Individuals may worry about data security and privacy if data are transmitted to and stored in the cloud. To alleviate these concerns, privacy-preserving deep learning has attracted attention from both the deep learning and information security communities. The recently proposed federated learning framework (please refer to Section~\ref{subsubsec:federatedlearning}) is a representative and promising solution that allows data to be stored locally in distributed devices. Homomorphic encryption \cite{aono2017privacy} has been used in federated learning to prevent data leakage to the server.

\section{Conclusion}
\label{sec:conclusion}
Here we present a comprehensive survey of AIoT, covering AIoT computing architectures; AI technologies for empowering IoT with perceiving, learning, reasoning, and behaving abilities; promising AIoT applications; and the challenges and opportunities facing AIoT research. The three-tier computing architecture of AIoT provides different computing resources for deep learning whilst also posing new challenges, e.g., in the design and search of lightweight models and computation scheduling within the three-tier architecture. Deep learning has rapidly progressed in many perceiving areas and enables many AIoT applications. Nevertheless, more effort should be made to improve edge intelligence. In the context of unsupervised learning and other machine learning topics like reinforcement learning, deep learning has attracted an increasing amount of attention and is useful for further improving the intelligence of AIoT systems to handle dynamic and complex environments. Moreover, reasoning based on knowledge graphs and causal analysis is a challenging but active research area, having the potential to enable AIoT systems to approach human-level cognitive intelligence. To respond to the dynamic environment, AIoT behaves via control and interaction, where deep learning has demonstrated its value in improving control accuracy and enabling multi-modal interactions. In the future, empowered by rapidly developing AI technologies, many fast, smart, green, and safe AIoT applications are expected to deeply reshape our world.



\begin{table}[htbp]
  \centering
  \caption{List of abbreviations.}
    \begin{tabular}{ll}
    \toprule
     Abbreviation & Description \\
     \midrule
     5G & 5th Generation Mobile Networks  \\
     AI & Artificial Intelligence \\
     AIoT & Artificial Intelligence of Things  \\
     AR/VR & Augmented Reality / Virtual Reality   \\
     ASIC & Application-Specific Integrated Circuit  \\
     ASR & Automatic Speech Recognition  \\
     CNN & Convolutional Neural Networks \\
     CTC & Connectionist Temporal Classification  \\
     DA & Domain Adaptation \\
     DNN & Deep Neural Network  \\
     DRL & Deep Reinforcement Learning \\
     FL & Federated Learning \\
     FOV & Field-of-View \\
     FPGA & Field-Programmable Gate Arrays  \\
     FSL & Few-Shot Learning \\
     GAN & Generative Adversarial Network  \\
     GCN & Graph Convolutional Network  \\
     GDPR & General Data Protection Regulation \\
     GPU & Graphics Processing Unit  \\
     HMI & Human-Machine Interaction  \\
     HMM & Hidden Markov Model  \\
     IIoT & Industrial Internet of Things \\
     IMU & Inertial Measurement Unit  \\
     IoT & Internet of Things  \\
     IoV & Internet of Vehicles  \\
     IoVT & Internet of Video Things  \\
     KG & Knowledge Graph \\
     LAN & Local Area Network  \\
     LSTM & Long Short-Term Memory  \\
     LTE & Long Term Evolution  \\
     MOOC & Massive Open Online Course \\
     MT & Machine Translation  \\
     NAS & Neural Architecture Search  \\
     NMT & Neural Machine Translation  \\
     OCR & Optical Character Recognition  \\
     ONNX & Open Neural Network Exchange  \\
     PCB & Printed Circuit Board \\
     RFID & Radio-Frequency Identification  \\
     RL & Reinforcement Learning \\
     RNN & Recurrent Neural Network  \\
     ROI & Region of Interest \\
     RTS & Real-Time Scheduling \\
     SLAM &  Simultaneous Localization and Mapping  \\
     SSL & Semi-Supervised Learning \\
     TL & Transfer Learning \\
     TPU & Tensor Processing Unit  \\
     UAV & Unmanned Aerial Vehicle \\
     UDA & Unsupervised Domain Adaptation \\
     USL & Unsupervised Learning \\
     VIO & Visual Inertial Odometry  \\
     VO & Visual Odometry  \\
     ZSL & Zero-Shot Learning \\
    \bottomrule
    \end{tabular}%
  \label{tab:abbreviation}%
\end{table}%

\ifCLASSOPTIONcaptionsoff
  \newpage
\fi



%



\normalem
\bibliographystyle{IEEEtran}
\bibliography{AIoT}\ 

\begin{thebibliography}{100}
\providecommand{\url}[1]{#1}
\csname url@samestyle\endcsname
\providecommand{\newblock}{\relax}
\providecommand{\bibinfo}[2]{#2}
\providecommand{\BIBentrySTDinterwordspacing}{\spaceskip=0pt\relax}
\providecommand{\BIBentryALTinterwordstretchfactor}{4}
\providecommand{\BIBentryALTinterwordspacing}{\spaceskip=\fontdimen2\font plus
\BIBentryALTinterwordstretchfactor\fontdimen3\font minus
  \fontdimen4\font\relax}
\providecommand{\BIBforeignlanguage}[2]{{%
\expandafter\ifx\csname l@#1\endcsname\relax
\typeout{** WARNING: IEEEtran.bst: No hyphenation pattern has been}%
\typeout{** loaded for the language `#1'. Using the pattern for}%
\typeout{** the default language instead.}%
\else
\language=\csname l@#1\endcsname
\fi
#2}}
\providecommand{\BIBdecl}{\relax}
\BIBdecl

\bibitem{ashton2009internet}
K.~Ashton \emph{et~al.}, ``That ‘internet of things’ thing,'' \emph{RFID
  Journal}, vol.~22, no.~7, pp. 97--114, 2009.

\bibitem{ccsa2011}
``Terms of the ubiquitous network,'' \emph{CCSA Standard YDB 062-2011}, 2011.

\bibitem{itut2012}
``Overview of iot,'' \emph{ITU-T Standard Y.2060}, 2012.

\bibitem{da2014internet}
L.~Da~Xu, W.~He, and S.~Li, ``Internet of things in industries: A survey,''
  \emph{IEEE Transactions on Industrial Informatics}, vol.~10, no.~4, pp.
  2233--2243, 2014.

\bibitem{li2011smart}
X.~Li, R.~Lu, X.~Liang, X.~Shen, J.~Chen, and X.~Lin, ``Smart community: An
  internet of things application,'' \emph{IEEE Communications Magazine},
  vol.~49, no.~11, pp. 68--75, 2011.

\bibitem{stojkoska2017review}
B.~L.~R. Stojkoska and K.~V. Trivodaliev, ``A review of internet of things for
  smart home: Challenges and solutions,'' \emph{Journal of Cleaner Production},
  vol. 140, pp. 1454--1464, 2017.

\bibitem{contreras2017internet}
J.~Contreras-Castillo, S.~Zeadally, and J.~A. Guerrero-Iba{\~n}ez, ``Internet
  of vehicles: Architecture, protocols, and security,'' \emph{IEEE Internet of
  Things Journal}, vol.~5, no.~5, pp. 3701--3709, 2017.

\bibitem{aslam2020internet}
S.~Aslam, M.~P. Michaelides, and H.~Herodotou, ``Internet of ships: A survey on
  architectures, emerging applications, and challenges,'' \emph{IEEE Internet
  of Things Journal}, 2020.

\bibitem{guan2017achieving}
Z.~Guan, J.~Li, L.~Wu, Y.~Zhang, J.~Wu, and X.~Du, ``Achieving efficient and
  secure data acquisition for cloud-supported internet of things in smart
  grid,'' \emph{IEEE Internet of Things Journal}, vol.~4, no.~6, pp.
  1934--1944, 2017.

\bibitem{catarinucci2015iot}
L.~Catarinucci, D.~De~Donno, L.~Mainetti, L.~Palano, L.~Patrono, M.~L.
  Stefanizzi, and L.~Tarricone, ``An iot-aware architecture for smart
  healthcare systems,'' \emph{IEEE Internet of Things Journal}, vol.~2, no.~6,
  pp. 515--526, 2015.

\bibitem{rahmani2018exploiting}
A.~M. Rahmani, T.~N. Gia, B.~Negash, A.~Anzanpour, I.~Azimi, M.~Jiang, and
  P.~Liljeberg, ``Exploiting smart e-health gateways at the edge of healthcare
  internet-of-things: A fog computing approach,'' \emph{Future Generation
  Computer Systems}, vol.~78, pp. 641--658, 2018.

\bibitem{lamarre2019ten}
E.~Lamarre and B.~May, ``Ten trends shaping the internet of things business
  landscape,'' \emph{McKinsey Digital}, 2019.

\bibitem{lin2017survey}
J.~Lin, W.~Yu, N.~Zhang, X.~Yang, H.~Zhang, and W.~Zhao, ``A survey on internet
  of things: Architecture, enabling technologies, security and privacy, and
  applications,'' \emph{IEEE Internet of Things Journal}, vol.~4, no.~5, pp.
  1125--1142, 2017.

\bibitem{macaulay2015internet}
J.~Macaulay, L.~Buckalew, and G.~Chung, ``Internet of things in logistics: A
  collaborative report by dhl and cisco on implications and use cases for the
  logistics industry,'' \emph{DHL Trend Research and Cisco Consulting
  Services}, 2015.

\bibitem{shirer2019growth}
M.~Shirer and C.~MacGillivray, ``The growth in connected iot devices is
  expected to generate 79.4 zb of data in 2025, according to a new idc
  forecast,'' 2019.

\bibitem{bonomi2011connected}
F.~Bonomi, ``Connected vehicles, the internet of things, and fog computing,''
  in \emph{The Eighth ACM International Workshop on Vehicular
  Inter-Networking}, 2011, pp. 13--15.

\bibitem{chiang2016fog}
M.~Chiang and T.~Zhang, ``Fog and iot: An overview of research opportunities,''
  \emph{IEEE Internet of Things Journal}, vol.~3, no.~6, pp. 854--864, 2016.

\bibitem{ni2017securing}
J.~Ni, K.~Zhang, X.~Lin, and X.~S. Shen, ``Securing fog computing for internet
  of things applications: Challenges and solutions,'' \emph{IEEE Communications
  Surveys \& Tutorials}, vol.~20, no.~1, pp. 601--628, 2017.

\bibitem{abbas2017mobile}
N.~Abbas, Y.~Zhang, A.~Taherkordi, and T.~Skeie, ``Mobile edge computing: A
  survey,'' \emph{IEEE Internet of Things Journal}, vol.~5, no.~1, pp.
  450--465, 2017.

\bibitem{pan2017future}
J.~Pan and J.~McElhannon, ``Future edge cloud and edge computing for internet
  of things applications,'' \emph{IEEE Internet of Things Journal}, vol.~5,
  no.~1, pp. 439--449, 2017.

\bibitem{shi2016edge}
W.~Shi, J.~Cao, Q.~Zhang, Y.~Li, and L.~Xu, ``Edge computing: Vision and
  challenges,'' \emph{IEEE Internet of Things Journal}, vol.~3, no.~5, pp.
  637--646, 2016.

\bibitem{tordera2016fog}
E.~M. Tordera, X.~Masip-Bruin, J.~Garcia-Alminana, A.~Jukan, G.-J. Ren, J.~Zhu,
  and J.~Farr{\'e}, ``What is a fog node a tutorial on current concepts towards
  a common definition,'' \emph{arXiv preprint arXiv:1611.09193}, 2016.

\bibitem{chen2020internet}
C.~W. Chen, ``Internet of video things: Next generation iot with visual
  sensors,'' \emph{IEEE Internet of Things Journal}, 2020.

\bibitem{sak2014long}
H.~Sak, A.~Senior, and F.~Beaufays, ``Long short-term memory recurrent neural
  network architectures for large scale acoustic modeling,'' in
  \emph{Proceedings of the Fifteenth Annual Conference of the International
  Speech Communication Association}, 2014.

\bibitem{taigman2014deepface}
Y.~Taigman, M.~Yang, M.~Ranzato, and L.~Wolf, ``Deepface: Closing the gap to
  human-level performance in face verification,'' in \emph{Proceedings of the
  IEEE Conference on Computer Vision and Pattern Recognition}, 2014, pp.
  1701--1708.

\bibitem{krizhevsky2012imagenet}
A.~Krizhevsky, I.~Sutskever, and G.~E. Hinton, ``Imagenet classification with
  deep convolutional neural networks,'' in \emph{Advances in Neural Information
  Processing Systems}, 2012, pp. 1097--1105.

\bibitem{ren2015faster}
S.~Ren, K.~He, R.~Girshick, and J.~Sun, ``Faster r-cnn: Towards real-time
  object detection with region proposal networks,'' in \emph{Advances in Neural
  Information Processing Systems}, 2015, pp. 91--99.

\bibitem{chen2017deeplab}
L.-C. Chen, G.~Papandreou, I.~Kokkinos, K.~Murphy, and A.~L. Yuille, ``Deeplab:
  Semantic image segmentation with deep convolutional nets, atrous convolution,
  and fully connected crfs,'' \emph{IEEE Transactions on Pattern Analysis and
  Machine Intelligence}, vol.~40, no.~4, pp. 834--848, 2017.

\bibitem{devlin2019bert}
J.~Devlin, M.-W. Chang, K.~Lee, and K.~Toutanova, ``Bert: Pre-training of deep
  bidirectional transformers for language understanding,'' in \emph{Proceedings
  of the 2019 Conference of the North American Chapter of the Association for
  Computational Linguistics: Human Language Technologies}, 2019, pp.
  4171--4186.

\bibitem{deng2009imagenet}
J.~Deng, W.~Dong, R.~Socher, L.-J. Li, K.~Li, and L.~Fei-Fei, ``Imagenet: A
  large-scale hierarchical image database,'' in \emph{Proceedings of the IEEE
  Conference on Computer Vision and Pattern Recognition}, 2009, pp. 248--255.

\bibitem{wang2020generalizing}
Y.~Wang, Q.~Yao, J.~T. Kwok, and L.~M. Ni, ``Generalizing from a few examples:
  A survey on few-shot learning,'' \emph{ACM Computing Surveys}, vol.~53,
  no.~3, pp. 1--34, 2020.

\bibitem{wang2019survey}
W.~Wang, V.~W. Zheng, H.~Yu, and C.~Miao, ``A survey of zero-shot learning:
  Settings, methods, and applications,'' \emph{ACM Transactions on Intelligent
  Systems and Technology}, vol.~10, no.~2, pp. 1--37, 2019.

\bibitem{finn2017model}
C.~Finn, P.~Abbeel, and S.~Levine, ``Model-agnostic meta-learning for fast
  adaptation of deep networks,'' in \emph{Proceedings of the 34th International
  Conference on Machine Learning}, 2017.

\bibitem{chen2020simple}
T.~Chen, S.~Kornblith, M.~Norouzi, and G.~Hinton, ``A simple framework for
  contrastive learning of visual representations,'' \emph{arXiv preprint
  arXiv:2002.05709}, 2020.

\bibitem{xie2020self}
Q.~Xie, M.-T. Luong, E.~Hovy, and Q.~V. Le, ``Self-training with noisy student
  improves imagenet classification,'' in \emph{Proceedings of the IEEE
  Conference on Computer Vision and Pattern Recognition}, 2020, pp.
  10\,687--10\,698.

\bibitem{zamir2018taskonomy}
A.~R. Zamir, A.~Sax, W.~Shen, L.~J. Guibas, J.~Malik, and S.~Savarese,
  ``Taskonomy: Disentangling task transfer learning,'' in \emph{Proceedings of
  the IEEE Conference on Computer Vision and Pattern Recognition}, 2018, pp.
  3712--3722.

\bibitem{zhang2019category}
Q.~Zhang, J.~Zhang, W.~Liu, and D.~Tao, ``Category anchor-guided unsupervised
  domain adaptation for semantic segmentation,'' in \emph{Advances in Neural
  Information Processing Systems}, 2019, pp. 435--445.

\bibitem{yang2020mobileda}
J.~Yang, H.~Zou, S.~Cao, Z.~Chen, and L.~Xie, ``Mobileda: Towards edge domain
  adaptation,'' \emph{IEEE Internet of Things Journal}, 2020.

\bibitem{zhang2019multi}
Y.~Zhang, S.~Qian, Q.~Fang, and C.~Xu, ``Multi-modal knowledge-aware
  hierarchical attention network for explainable medical question answering,''
  in \emph{Proceedings of the 27th ACM International Conference on Multimedia},
  2019, pp. 1089--1097.

\bibitem{wang2019explainable}
X.~Wang, D.~Wang, C.~Xu, X.~He, Y.~Cao, and T.-S. Chua, ``Explainable reasoning
  over knowledge graphs for recommendation,'' in \emph{Proceedings of the AAAI
  Conference on Artificial Intelligence}, vol.~33, 2019, pp. 5329--5336.

\bibitem{zhang2019hierarchical}
J.~Zhang, B.~Hao, B.~Chen, C.~Li, H.~Chen, and J.~Sun, ``Hierarchical
  reinforcement learning for course recommendation in moocs,'' in
  \emph{Proceedings of the AAAI Conference on Artificial Intelligence},
  vol.~33, 2019, pp. 435--442.

\bibitem{arik2017deep}
S.~{\"O}. Ar{\i}k, M.~Chrzanowski, A.~Coates, G.~Diamos, A.~Gibiansky, Y.~Kang,
  X.~Li, J.~Miller, A.~Ng, J.~Raiman \emph{et~al.}, ``Deep voice: Real-time
  neural text-to-speech,'' in \emph{Proceedings of the 34th International
  Conference on Machine Learning}, 2017, pp. 195--204.

\bibitem{bradley2013embracing}
J.~Bradley, J.~Barbier, and D.~Handler, ``Embracing the internet of everything
  to capture your share of \$14.4 trillion,'' \emph{White Paper, Cisco}, vol.
  318, 2013.

\bibitem{atzori2010internet}
L.~Atzori, A.~Iera, and G.~Morabito, ``The internet of things: A survey,''
  \emph{Computer networks}, vol.~54, no.~15, pp. 2787--2805, 2010.

\bibitem{whitmore2015internet}
A.~Whitmore, A.~Agarwal, and L.~Da~Xu, ``The internet of things—a survey of
  topics and trends,'' \emph{Information Systems Frontiers}, vol.~17, no.~2,
  pp. 261--274, 2015.

\bibitem{yaqoob2017internet}
I.~Yaqoob, E.~Ahmed, I.~A.~T. Hashem, A.~I.~A. Ahmed, A.~Gani, M.~Imran, and
  M.~Guizani, ``Internet of things architecture: Recent advances, taxonomy,
  requirements, and open challenges,'' \emph{IEEE Wireless Communications},
  vol.~24, no.~3, pp. 10--16, 2017.

\bibitem{gubbi2013internet}
J.~Gubbi, R.~Buyya, S.~Marusic, and M.~Palaniswami, ``Internet of things (iot):
  A vision, architectural elements, and future directions,'' \emph{Future
  Generation Computer Systems}, vol.~29, no.~7, pp. 1645--1660, 2013.

\bibitem{verma2017survey}
S.~Verma, Y.~Kawamoto, Z.~M. Fadlullah, H.~Nishiyama, and N.~Kato, ``A survey
  on network methodologies for real-time analytics of massive iot data and open
  research issues,'' \emph{IEEE Communications Surveys \& Tutorials}, vol.~19,
  no.~3, pp. 1457--1477, 2017.

\bibitem{mainetti2011evolution}
L.~Mainetti, L.~Patrono, and A.~Vilei, ``Evolution of wireless sensor networks
  towards the internet of things: A survey,'' in \emph{Proceedings of the 19th
  International Conference on Software, Telecommunications and Computer
  Networks}.\hskip 1em plus 0.5em minus 0.4em\relax IEEE, 2011, pp. 1--6.

\bibitem{chettri2019comprehensive}
L.~Chettri and R.~Bera, ``A comprehensive survey on internet of things (iot)
  toward 5g wireless systems,'' \emph{IEEE Internet of Things Journal}, vol.~7,
  no.~1, pp. 16--32, 2019.

\bibitem{hussain2020machine}
F.~Hussain, S.~A. Hassan, R.~Hussain, and E.~Hossain, ``Machine learning for
  resource management in cellular and iot networks: Potentials, current
  solutions, and open challenges,'' \emph{IEEE Communications Surveys \&
  Tutorials}, vol.~22, no.~2, pp. 1251--1275, 2020.

\bibitem{tsai2013data}
C.-W. Tsai, C.-F. Lai, M.-C. Chiang, and L.~T. Yang, ``Data mining for internet
  of things: A survey,'' \emph{IEEE Communications Surveys \& Tutorials},
  vol.~16, no.~1, pp. 77--97, 2013.

\bibitem{mahdavinejad2018machine}
M.~S. Mahdavinejad, M.~Rezvan, M.~Barekatain, P.~Adibi, P.~Barnaghi, and A.~P.
  Sheth, ``Machine learning for internet of things data analysis: A survey,''
  \emph{Digital Communications and Networks}, vol.~4, no.~3, pp. 161--175,
  2018.

\bibitem{perera2013context}
C.~Perera, A.~Zaslavsky, P.~Christen, and D.~Georgakopoulos, ``Context aware
  computing for the internet of things: A survey,'' \emph{IEEE Communications
  Surveys \& Tutorials}, vol.~16, no.~1, pp. 414--454, 2013.

\bibitem{mohammadi2018deep}
M.~Mohammadi, A.~Al-Fuqaha, S.~Sorour, and M.~Guizani, ``Deep learning for iot
  big data and streaming analytics: A survey,'' \emph{IEEE Communications
  Surveys \& Tutorials}, vol.~20, no.~4, pp. 2923--2960, 2018.

\bibitem{ota2017deep}
K.~Ota, M.~S. Dao, V.~Mezaris, and F.~G.~D. Natale, ``Deep learning for mobile
  multimedia: A survey,'' \emph{ACM Transactions on Multimedia Computing,
  Communications, and Applications}, vol.~13, no.~3s, pp. 1--22, 2017.

\bibitem{li2018deep}
L.~Li, K.~Ota, and M.~Dong, ``Deep learning for smart industry: Efficient
  manufacture inspection system with fog computing,'' \emph{IEEE Transactions
  on Industrial Informatics}, vol.~14, no.~10, pp. 4665--4673, 2018.

\bibitem{qiu2020survey}
J.~Qiu, Z.~Tian, C.~Du, Q.~Zuo, S.~Su, and B.~Fang, ``A survey on access
  control in the age of internet of things,'' \emph{IEEE Internet of Things
  Journal}, 2020.

\bibitem{yan2014survey}
Z.~Yan, P.~Zhang, and A.~V. Vasilakos, ``A survey on trust management for
  internet of things,'' \emph{Journal of Network and Computer Applications},
  vol.~42, pp. 120--134, 2014.

\bibitem{yang2019federated}
Q.~Yang, Y.~Liu, T.~Chen, and Y.~Tong, ``Federated machine learning: Concept
  and applications,'' \emph{ACM Transactions on Intelligent Systems and
  Technology}, vol.~10, no.~2, pp. 1--19, 2019.

\bibitem{zanella2014internet}
A.~Zanella, N.~Bui, A.~Castellani, L.~Vangelista, and M.~Zorzi, ``Internet of
  things for smart cities,'' \emph{IEEE Internet of Things Journal}, vol.~1,
  no.~1, pp. 22--32, 2014.

\bibitem{tokognon2017structural}
C.~A. Tokognon, B.~Gao, G.~Y. Tian, and Y.~Yan, ``Structural health monitoring
  framework based on internet of things: A survey,'' \emph{IEEE Internet of
  Things Journal}, vol.~4, no.~3, pp. 619--635, 2017.

\bibitem{elijah2018overview}
O.~Elijah, T.~A. Rahman, I.~Orikumhi, C.~Y. Leow, and M.~N. Hindia, ``An
  overview of internet of things (iot) and data analytics in agriculture:
  Benefits and challenges,'' \emph{IEEE Internet of Things Journal}, vol.~5,
  no.~5, pp. 3758--3773, 2018.

\bibitem{he2016deep}
K.~He, X.~Zhang, S.~Ren, and J.~Sun, ``Deep residual learning for image
  recognition,'' in \emph{Proceedings of the IEEE Conference on Computer Vision
  and Pattern Recognition}, 2016, pp. 770--778.

\bibitem{howard2017mobilenets}
A.~G. Howard, M.~Zhu, B.~Chen, D.~Kalenichenko, W.~Wang, T.~Weyand,
  M.~Andreetto, and H.~Adam, ``Mobilenets: Efficient convolutional neural
  networks for mobile vision applications,'' \emph{arXiv preprint
  arXiv:1704.04861}, 2017.

\bibitem{zhang2018fully}
J.~Zhang, Y.~Cao, Y.~Wang, C.~Wen, and C.~W. Chen, ``Fully point-wise
  convolutional neural network for modeling statistical regularities in natural
  images,'' in \emph{Proceedings of the 26th ACM international conference on
  Multimedia}, 2018, pp. 984--992.

\bibitem{rastegari2016xnor}
M.~Rastegari, V.~Ordonez, J.~Redmon, and A.~Farhadi, ``Xnor-net: Imagenet
  classification using binary convolutional neural networks,'' in
  \emph{Proceedings of the European Conference on Computer Vision}.\hskip 1em
  plus 0.5em minus 0.4em\relax Springer, 2016, pp. 525--542.

\bibitem{redmon2016you}
J.~Redmon, S.~Divvala, R.~Girshick, and A.~Farhadi, ``You only look once:
  Unified, real-time object detection,'' in \emph{Proceedings of the IEEE
  Conference on Computer Vision and Pattern Recognition}, 2016, pp. 779--788.

\bibitem{law2018cornernet}
H.~Law and J.~Deng, ``Cornernet: Detecting objects as paired keypoints,'' in
  \emph{Proceedings of the European Conference on Computer Vision}, 2018, pp.
  734--750.

\bibitem{chen2015deepdriving}
C.~Chen, A.~Seff, A.~Kornhauser, and J.~Xiao, ``Deepdriving: Learning
  affordance for direct perception in autonomous driving,'' in
  \emph{Proceedings of the IEEE International Conference on Computer Vision},
  2015, pp. 2722--2730.

\bibitem{li2013survey}
X.~Li, W.~Hu, C.~Shen, Z.~Zhang, A.~Dick, and A.~V.~D. Hengel, ``A survey of
  appearance models in visual object tracking,'' \emph{ACM transactions on
  Intelligent Systems and Technology}, vol.~4, no.~4, pp. 1--48, 2013.

\bibitem{danelljan2016beyond}
M.~Danelljan, A.~Robinson, F.~S. Khan, and M.~Felsberg, ``Beyond correlation
  filters: Learning continuous convolution operators for visual tracking,'' in
  \emph{Proceedings of the European Conference on Computer Vision}, 2016, pp.
  472--488.

\bibitem{valmadre2017end}
J.~Valmadre, L.~Bertinetto, J.~Henriques, A.~Vedaldi, and P.~H. Torr,
  ``End-to-end representation learning for correlation filter based tracking,''
  in \emph{Proceedings of the IEEE Conference on Computer Vision and Pattern
  Recognition}, 2017, pp. 2805--2813.

\bibitem{li2019siamrpn}
B.~Li, W.~Wu, Q.~Wang, F.~Zhang, J.~Xing, and J.~Yan, ``Siamrpn++: Evolution of
  siamese visual tracking with very deep networks,'' in \emph{Proceedings of
  the IEEE Conference on Computer Vision and Pattern Recognition}, 2019, pp.
  4282--4291.

\bibitem{ronneberger2015u}
O.~Ronneberger, P.~Fischer, and T.~Brox, ``U-net: Convolutional networks for
  biomedical image segmentation,'' in \emph{Proceedings of the International
  Conference on Medical Image Computing and Computer-Assisted Intervention},
  2015, pp. 234--241.

\bibitem{long2015fully}
J.~Long, E.~Shelhamer, and T.~Darrell, ``Fully convolutional networks for
  semantic segmentation,'' in \emph{Proceedings of the IEEE Conference on
  Computer Vision and Pattern Recognition}, 2015, pp. 3431--3440.

\bibitem{liu2015parsenet}
W.~Liu, A.~Rabinovich, and A.~C. Berg, ``Parsenet: Looking wider to see
  better,'' in \emph{Proceedings of the International Conference on Learning
  Representations}, 2016.

\bibitem{zhao2017pyramid}
H.~Zhao, J.~Shi, X.~Qi, X.~Wang, and J.~Jia, ``Pyramid scene parsing network,''
  in \emph{Proceedings of the IEEE Conference on Computer Vision and Pattern
  Recognition}, 2017, pp. 2881--2890.

\bibitem{he2017mask}
K.~He, G.~Gkioxari, P.~Doll{\'a}r, and R.~Girshick, ``Mask r-cnn,'' in
  \emph{Proceedings of the IEEE International Conference on Computer Vision},
  2017, pp. 2961--2969.

\bibitem{lin2017feature}
T.-Y. Lin, P.~Doll{\'a}r, R.~Girshick, K.~He, B.~Hariharan, and S.~Belongie,
  ``Feature pyramid networks for object detection,'' in \emph{Proceedings of
  the IEEE Conference on Computer Vision and Pattern Recognition}, 2017, pp.
  2117--2125.

\bibitem{wang2018non}
X.~Wang, R.~Girshick, A.~Gupta, and K.~He, ``Non-local neural networks,'' in
  \emph{Proceedings of the IEEE Conference on Computer Vision and Pattern
  Recognition}, 2018, pp. 7794--7803.

\bibitem{chen2019hybrid}
K.~Chen, J.~Pang, J.~Wang, Y.~Xiong, X.~Li, S.~Sun, W.~Feng, Z.~Liu, J.~Shi,
  W.~Ouyang \emph{et~al.}, ``Hybrid task cascade for instance segmentation,''
  in \emph{Proceedings of the IEEE Conference on Computer Vision and Pattern
  Recognition}, 2019, pp. 4974--4983.

\bibitem{kirillov2019panoptic}
A.~Kirillov, K.~He, R.~Girshick, C.~Rother, and P.~Doll{\'a}r, ``Panoptic
  segmentation,'' in \emph{Proceedings of the IEEE Conference on Computer
  Vision and Pattern Recognition}, 2019, pp. 9404--9413.

\bibitem{kirillov2019panopticfeature}
A.~Kirillov, R.~Girshick, K.~He, and P.~Doll{\'a}r, ``Panoptic feature pyramid
  networks,'' in \emph{Proceedings of the IEEE Conference on Computer Vision
  and Pattern Recognition}, 2019, pp. 6399--6408.

\bibitem{chen2019progressive}
Z.~Chen, J.~Zhang, and D.~Tao, ``Progressive lidar adaptation for road
  detection,'' \emph{IEEE/CAA Journal of Automatica Sinica}, vol.~6, no.~3, pp.
  693--702, 2019.

\bibitem{he2020grapy}
H.~He, J.~Zhang, Q.~Zhang, and D.~Tao, ``Grapy-ml: Graph pyramid mutual
  learning for cross-dataset human parsing,'' \emph{Proceedings of the
  Thirty-Fourth AAAI Conference on Artificial Intelligence}, vol.~34, no.~7,
  pp. 10\,949--10\,956, 2020.

\bibitem{jaderberg2014deep}
M.~Jaderberg, A.~Vedaldi, and A.~Zisserman, ``Deep features for text
  spotting,'' in \emph{Proceedings of the European Conference on Computer
  Vision}, 2014, pp. 512--528.

\bibitem{liu2018fots}
X.~Liu, D.~Liang, S.~Yan, D.~Chen, Y.~Qiao, and J.~Yan, ``Fots: Fast oriented
  text spotting with a unified network,'' in \emph{Proceedings of the IEEE
  Conference on Computer Vision and Pattern Recognition}, 2018, pp. 5676--5685.

\bibitem{gupta2016synthetic}
A.~Gupta, A.~Vedaldi, and A.~Zisserman, ``Synthetic data for text localisation
  in natural images,'' in \emph{Proceedings of the IEEE Conference on Computer
  Vision and Pattern Recognition}, 2016, pp. 2315--2324.

\bibitem{li2017towards}
H.~Li, P.~Wang, and C.~Shen, ``Towards end-to-end text spotting with
  convolutional recurrent neural networks,'' in \emph{Proceedings of the IEEE
  International Conference on Computer Vision}, 2017, pp. 5238--5246.

\bibitem{liu2020asts}
J.~Liu, Z.~Chen, B.~Du, and D.~Tao, ``Asts: A unified framework for arbitrary
  shape text spotting,'' \emph{IEEE Transactions on Image Processing}, vol.~29,
  pp. 5924--5936, 2020.

\bibitem{graves2006connectionist}
A.~Graves, S.~Fern{\'a}ndez, F.~Gomez, and J.~Schmidhuber, ``Connectionist
  temporal classification: Labelling unsegmented sequence data with recurrent
  neural networks,'' in \emph{Proceedings of the 23rd International Conference
  on Machine Learning}, 2006, pp. 369--376.

\bibitem{mancas2007multifunctional}
C.~Mancas-Thillou, S.~Ferreira, J.~Demeyer, C.~Minetti, and B.~Gosselin, ``A
  multifunctional reading assistant for the visually impaired,'' \emph{EURASIP
  Journal on Image and Video Processing}, vol. 2007, no.~1, p. 064295, 2007.

\bibitem{viola2001rapid}
P.~Viola and M.~Jones, ``Rapid object detection using a boosted cascade of
  simple features,'' in \emph{Proceedings of the IEEE Conference on Computer
  Vision and Pattern Recognition}, 2001.

\bibitem{li2015convolutional}
H.~Li, Z.~Lin, X.~Shen, J.~Brandt, and G.~Hua, ``A convolutional neural network
  cascade for face detection,'' in \emph{Proceedings of the IEEE Conference on
  Computer Vision and Pattern Recognition}, 2015, pp. 5325--5334.

\bibitem{zhang2016joint}
K.~Zhang, Z.~Zhang, Z.~Li, and Y.~Qiao, ``Joint face detection and alignment
  using multitask cascaded convolutional networks,'' \emph{IEEE Signal
  Processing Letters}, vol.~23, no.~10, pp. 1499--1503, 2016.

\bibitem{ranjan2017hyperface}
R.~Ranjan, V.~M. Patel, and R.~Chellappa, ``Hyperface: A deep multi-task
  learning framework for face detection, landmark localization, pose
  estimation, and gender recognition,'' \emph{IEEE Transactions on Pattern
  Analysis and Machine Intelligence}, vol.~41, no.~1, pp. 121--135, 2017.

\bibitem{ren2014face}
S.~Ren, X.~Cao, Y.~Wei, and J.~Sun, ``Face alignment at 3000 fps via regressing
  local binary features,'' in \emph{Proceedings of the IEEE Conference on
  Computer Vision and Pattern Recognition}, 2014, pp. 1685--1692.

\bibitem{schroff2015facenet}
F.~Schroff, D.~Kalenichenko, and J.~Philbin, ``Facenet: A unified embedding for
  face recognition and clustering,'' in \emph{Proceedings of the IEEE
  Conference on Computer Vision and Pattern Recognition}, 2015, pp. 815--823.

\bibitem{soyata2012cloud}
T.~Soyata, R.~Muraleedharan, C.~Funai, M.~Kwon, and W.~Heinzelman,
  ``Cloud-vision: Real-time face recognition using a mobile-cloudlet-cloud
  acceleration architecture,'' in \emph{Proceedings of the IEEE Symposium on
  Computers and Communications}, 2012.

\bibitem{amos2016openface}
B.~Amos, B.~Ludwiczuk, M.~Satyanarayanan \emph{et~al.}, ``Openface: A
  general-purpose face recognition library with mobile applications,''
  \emph{CMU School of Computer Science}, vol.~6, no.~2, 2016.

\bibitem{ramachandra2017presentation}
R.~Ramachandra and C.~Busch, ``Presentation attack detection methods for face
  recognition systems: A comprehensive survey,'' \emph{ACM Computing Surveys},
  vol.~50, no.~1, pp. 1--37, 2017.

\bibitem{fei2018feature}
L.~Fei, G.~Lu, W.~Jia, S.~Teng, and D.~Zhang, ``Feature extraction methods for
  palmprint recognition: A survey and evaluation,'' \emph{IEEE Transactions on
  Systems, Man, and Cybernetics: Systems}, vol.~49, no.~2, pp. 346--363, 2018.

\bibitem{zhang2019pay}
Y.~Zhang, L.~Zhang, X.~Liu, S.~Zhao, Y.~Shen, and Y.~Yang, ``Pay by showing
  your palm: A study of palmprint verification on mobile platforms,'' in
  \emph{Proceedings of the IEEE International Conference on Multimedia and
  Expo}.\hskip 1em plus 0.5em minus 0.4em\relax IEEE, 2019, pp. 862--867.

\bibitem{zhang2017towards}
L.~Zhang, L.~Li, A.~Yang, Y.~Shen, and M.~Yang, ``Towards contactless palmprint
  recognition: A novel device, a new benchmark, and a collaborative
  representation based identification approach,'' \emph{Pattern Recognition},
  vol.~69, pp. 199--212, 2017.

\bibitem{chen2020recursive}
Z.~Chen, J.~Zhang, and D.~Tao, ``Recursive context routing for object
  detection,'' \emph{International Journal of Computer Vision}, pp. 1--19,
  2020.

\bibitem{zhang2020towards}
J.~Zhang, Z.~Chen, and D.~Tao, ``Towards high performance human keypoint
  detection,'' \emph{arXiv preprint arXiv:2002.00537}, 2020.

\bibitem{zeng2019soft}
X.~Zeng, C.~Ding, Y.~Wen, and D.~Tao, ``Soft-ranking label encoding for robust
  facial age estimation,'' \emph{arXiv preprint arXiv:1906.03625}, 2019.

\bibitem{fu2018deep}
H.~Fu, M.~Gong, C.~Wang, K.~Batmanghelich, and D.~Tao, ``Deep ordinal
  regression network for monocular depth estimation,'' in \emph{Proceedings of
  the IEEE Conference on Computer Vision and Pattern Recognition}, 2018, pp.
  2002--2011.

\bibitem{ye2020deep}
M.~Ye, J.~Shen, G.~Lin, T.~Xiang, L.~Shao, and S.~C. Hoi, ``Deep learning for
  person re-identification: A survey and outlook,'' \emph{arXiv preprint
  arXiv:2001.04193}, 2020.

\bibitem{wei2018person}
L.~Wei, S.~Zhang, W.~Gao, and Q.~Tian, ``Person transfer gan to bridge domain
  gap for person re-identification,'' in \emph{Proceedings of the IEEE
  Conference on Computer Vision and Pattern Recognition}, 2018, pp. 79--88.

\bibitem{cao2017realtime}
Z.~Cao, T.~Simon, S.-E. Wei, and Y.~Sheikh, ``Realtime multi-person 2d pose
  estimation using part affinity fields,'' in \emph{Proceedings of the IEEE
  Conference on Computer Vision and Pattern Recognition}, 2017, pp. 7291--7299.

\bibitem{sun2019deep}
K.~Sun, B.~Xiao, D.~Liu, and J.~Wang, ``Deep high-resolution representation
  learning for human pose estimation,'' in \emph{Proceedings of the IEEE
  Conference on Computer Vision and Pattern Recognition}, 2019, pp. 5693--5703.

\bibitem{zhang2020distribution}
F.~Zhang, X.~Zhu, H.~Dai, M.~Ye, and C.~Zhu, ``Distribution-aware coordinate
  representation for human pose estimation,'' in \emph{Proceedings of the IEEE
  Conference on Computer Vision and Pattern Recognition}, 2020, pp. 7093--7102.

\bibitem{wang2019not}
J.~Wang, S.~Huang, X.~Wang, and D.~Tao, ``Not all parts are created equal: 3d
  pose estimation by modeling bi-directional dependencies of body parts,'' in
  \emph{Proceedings of the IEEE International Conference on Computer Vision},
  2019, pp. 7771--7780.

\bibitem{du2015hierarchical}
Y.~Du, W.~Wang, and L.~Wang, ``Hierarchical recurrent neural network for
  skeleton based action recognition,'' in \emph{Proceedings of the IEEE
  Conference on Computer Vision and Pattern Recognition}, 2015, pp. 1110--1118.

\bibitem{ke2017new}
Q.~Ke, M.~Bennamoun, S.~An, F.~Sohel, and F.~Boussaid, ``A new representation
  of skeleton sequences for 3d action recognition,'' in \emph{Proceedings of
  the IEEE Conference on Computer Vision and Pattern Recognition}, 2017, pp.
  3288--3297.

\bibitem{yan2018spatial}
S.~Yan, Y.~Xiong, and D.~Lin, ``Spatial temporal graph convolutional networks
  for skeleton-based action recognition,'' \emph{arXiv preprint
  arXiv:1801.07455}, 2018.

\bibitem{song2017end}
S.~Song, C.~Lan, J.~Xing, W.~Zeng, and J.~Liu, ``An end-to-end spatio-temporal
  attention model for human action recognition from skeleton data,'' in
  \emph{Proceedings of the Thirty-First AAAI Conference on Artificial
  Intelligence}, 2017.

\bibitem{gonzalez2014kinect}
D.~Gonz{\'a}lez-Ortega, F.~D{\'\i}az-Pernas, M.~Mart{\'\i}nez-Zarzuela, and
  M.~Ant{\'o}n-Rodr{\'\i}guez, ``A kinect-based system for cognitive
  rehabilitation exercises monitoring,'' \emph{Computer Methods and Programs in
  Biomedicine}, vol. 113, no.~2, pp. 620--631, 2014.

\bibitem{kashevnik2019methodology}
A.~Kashevnik, I.~Lashkov, and A.~Gurtov, ``Methodology and mobile application
  for driver behavior analysis and accident prevention,'' \emph{IEEE
  Transactions on Intelligent Transportation Systems}, vol.~21, no.~6, pp.
  2427--2436, 2019.

\bibitem{rautaray2015vision}
S.~S. Rautaray and A.~Agrawal, ``Vision based hand gesture recognition for
  human computer interaction: A survey,'' \emph{Artificial Intelligence
  Review}, vol.~43, no.~1, pp. 1--54, 2015.

\bibitem{lien2016soli}
J.~Lien, N.~Gillian, M.~E. Karagozler, P.~Amihood, C.~Schwesig, E.~Olson,
  H.~Raja, and I.~Poupyrev, ``Soli: Ubiquitous gesture sensing with millimeter
  wave radar,'' \emph{ACM Transactions on Graphics}, vol.~35, no.~4, pp. 1--19,
  2016.

\bibitem{amir2017low}
A.~Amir, B.~Taba, D.~Berg, T.~Melano, J.~McKinstry, C.~Di~Nolfo, T.~Nayak,
  A.~Andreopoulos, G.~Garreau, M.~Mendoza \emph{et~al.}, ``A low power, fully
  event-based gesture recognition system,'' in \emph{Proceedings of the IEEE
  Conference on Computer Vision and Pattern Recognition}, 2017, pp. 7243--7252.

\bibitem{starner1998real}
T.~Starner, J.~Weaver, and A.~Pentland, ``Real-time american sign language
  recognition using desk and wearable computer based video,'' \emph{IEEE
  Transactions on Pattern Analysis and Machine Intelligence}, vol.~20, no.~12,
  pp. 1371--1375, 1998.

\bibitem{schauer2014estimating}
L.~Schauer, M.~Werner, and P.~Marcus, ``Estimating crowd densities and
  pedestrian flows using wi-fi and bluetooth,'' in \emph{Proceedings of the
  11th International Conference on Mobile and Ubiquitous Systems: Computing,
  Networking and Services}, 2014, pp. 171--177.

\bibitem{wang2020nwpu}
Q.~Wang, J.~Gao, W.~Lin, and X.~Li, ``Nwpu-crowd: A large-scale benchmark for
  crowd counting and localization,'' \emph{IEEE Transactions on Pattern
  Analysis and Machine Intelligence}, 2020.

\bibitem{wang2019learning}
Q.~Wang, J.~Gao, W.~Lin, and Y.~Yuan, ``Learning from synthetic data for crowd
  counting in the wild,'' in \emph{Proceedings of the IEEE Conference on
  Computer Vision and Pattern Recognition}, 2019, pp. 8198--8207.

\bibitem{hoiem2007recovering}
D.~Hoiem, A.~N. Stein, A.~A. Efros, and M.~Hebert, ``Recovering occlusion
  boundaries from a single image,'' in \emph{Proceedings of the IEEE
  International Conference on Computer Vision}.\hskip 1em plus 0.5em minus
  0.4em\relax IEEE, 2007, pp. 1--8.

\bibitem{wang2008region}
Z.-F. Wang and Z.-G. Zheng, ``A region based stereo matching algorithm using
  cooperative optimization,'' in \emph{Proceedings of the IEEE Conference on
  Computer Vision and Pattern Recognition}, 2008, pp. 1--8.

\bibitem{mahjourian2018unsupervised}
R.~Mahjourian, M.~Wicke, and A.~Angelova, ``Unsupervised learning of depth and
  ego-motion from monocular video using 3d geometric constraints,'' in
  \emph{Proceedings of the IEEE Conference on Computer Vision and Pattern
  Recognition}, 2018, pp. 5667--5675.

\bibitem{gordon2019depth}
A.~Gordon, H.~Li, R.~Jonschkowski, and A.~Angelova, ``Depth from videos in the
  wild: Unsupervised monocular depth learning from unknown cameras,'' in
  \emph{Proceedings of the IEEE International Conference on Computer Vision},
  2019, pp. 8977--8986.

\bibitem{yang2020d3vo}
N.~Yang, L.~v. Stumberg, R.~Wang, and D.~Cremers, ``D3vo: Deep depth, deep pose
  and deep uncertainty for monocular visual odometry,'' in \emph{Proceedings of
  the IEEE Conference on Computer Vision and Pattern Recognition}, 2020, pp.
  1281--1292.

\bibitem{han2019deepvio}
L.~Han, Y.~Lin, G.~Du, and S.~Lian, ``Deepvio: Self-supervised deep learning of
  monocular visual inertial odometry using 3d geometric constraints,'' in
  \emph{Proceedings of the IEEE/RSJ International Conference on Intelligent
  Robots and Systems}.\hskip 1em plus 0.5em minus 0.4em\relax IEEE, 2019, pp.
  6906--6913.

\bibitem{mur2015orb}
R.~Mur-Artal, J.~M.~M. Montiel, and J.~D. Tardos, ``Orb-slam: a versatile and
  accurate monocular slam system,'' \emph{IEEE Transactions on Robotics},
  vol.~31, no.~5, pp. 1147--1163, 2015.

\bibitem{parisotto2018global}
E.~Parisotto, D.~Singh~Chaplot, J.~Zhang, and R.~Salakhutdinov, ``Global pose
  estimation with an attention-based recurrent network,'' in \emph{Proceedings
  of the IEEE Conference on Computer Vision and Pattern Recognition Workshops},
  2018, pp. 237--246.

\bibitem{shao2020tightly}
X.~Shao, L.~Zhang, T.~Zhang, Y.~Shen, H.~Li, and Y.~Zhou, ``A tightly-coupled
  semantic slam system with visual, inertial and surround-view sensors for
  autonomous indoor parking,'' in \emph{Proceedings of the 28th ACM
  International Conference on Multimedia}, 2020.

\bibitem{krug2016next}
R.~Krug, T.~Stoyanov, V.~Tincani, H.~Andreasson, R.~Mosberger, G.~Fantoni, and
  A.~J. Lilienthal, ``The next step in robot commissioning: Autonomous picking
  and palletizing,'' \emph{IEEE Robotics and Automation Letters}, vol.~1,
  no.~1, pp. 546--553, 2016.

\bibitem{cheein2011optimized}
F.~A. Cheein, G.~Steiner, G.~P. Paina, and R.~Carelli, ``Optimized eif-slam
  algorithm for precision agriculture mapping based on stems detection,''
  \emph{Computers and Electronics in Agriculture}, vol.~78, no.~2, pp.
  195--207, 2011.

\bibitem{alsalam2017autonomous}
B.~H.~Y. Alsalam, K.~Morton, D.~Campbell, and F.~Gonzalez, ``Autonomous uav
  with vision based on-board decision making for remote sensing and precision
  agriculture,'' in \emph{Proceedings of the IEEE Aerospace Conference}, 2017,
  pp. 1--12.

\bibitem{liu2018robust}
X.~Liu, S.~W. Chen, S.~Aditya, N.~Sivakumar, S.~Dcunha, C.~Qu, C.~J. Taylor,
  J.~Das, and V.~Kumar, ``Robust fruit counting: Combining deep learning,
  tracking, and structure from motion,'' in \emph{Proceedings of the IEEE/RSJ
  International Conference on Intelligent Robots and Systems}, 2018, pp.
  1045--1052.

\bibitem{lee2016rgb}
Y.~H. Lee and G.~Medioni, ``Rgb-d camera based wearable navigation system for
  the visually impaired,'' \emph{Computer Vision and Image Understanding}, vol.
  149, pp. 3--20, 2016.

\bibitem{wang2017enabling}
H.-C. Wang, R.~K. Katzschmann, S.~Teng, B.~Araki, L.~Giarr{\'e}, and D.~Rus,
  ``Enabling independent navigation for visually impaired people through a
  wearable vision-based feedback system,'' in \emph{Proceedings of the IEEE
  International Conference on Robotics and Automation}, 2017, pp. 6533--6540.

\bibitem{katzschmann2018safe}
R.~K. Katzschmann, B.~Araki, and D.~Rus, ``Safe local navigation for visually
  impaired users with a time-of-flight and haptic feedback device,'' \emph{IEEE
  Transactions on Neural Systems and Rehabilitation Engineering}, vol.~26,
  no.~3, pp. 583--593, 2018.

\bibitem{land1977retinex}
E.~H. Land, ``The retinex theory of color vision,'' \emph{Scientific american},
  vol. 237, no.~6, pp. 108--129, 1977.

\bibitem{guo2016lime}
X.~Guo, Y.~Li, and H.~Ling, ``Lime: Low-light image enhancement via
  illumination map estimation,'' \emph{IEEE Transactions on Image Processing},
  vol.~26, no.~2, pp. 982--993, 2016.

\bibitem{li2018structure}
M.~Li, J.~Liu, W.~Yang, X.~Sun, and Z.~Guo, ``Structure-revealing low-light
  image enhancement via robust retinex model,'' \emph{IEEE Transactions on
  Image Processing}, vol.~27, no.~6, pp. 2828--2841, 2018.

\bibitem{zhu2020zero}
A.~Zhu, L.~Zhang, Y.~Shen, Y.~Ma, S.~Zhao, and Y.~Zhou, ``Zero-shot restoration
  of underexposed images via robust retinex decomposition,'' in
  \emph{Proceedings of 2020 IEEE International Conference on Multimedia and
  Expo}.\hskip 1em plus 0.5em minus 0.4em\relax IEEE, 2020, pp. 1--6.

\bibitem{he2010single}
K.~He, J.~Sun, and X.~Tang, ``Single image haze removal using dark channel
  prior,'' \emph{IEEE Transactions on Pattern Analysis and Machine
  Intelligence}, vol.~33, no.~12, pp. 2341--2353, 2010.

\bibitem{cai2016dehazenet}
B.~Cai, X.~Xu, K.~Jia, C.~Qing, and D.~Tao, ``Dehazenet: An end-to-end system
  for single image haze removal,'' \emph{IEEE Transactions on Image
  Processing}, vol.~25, no.~11, pp. 5187--5198, 2016.

\bibitem{zhang2019famed}
J.~Zhang and D.~Tao, ``Famed-net: A fast and accurate multi-scale end-to-end
  dehazing network,'' \emph{IEEE Transactions on Image Processing}, vol.~29,
  pp. 72--84, 2019.

\bibitem{zhao2020dehazing}
S.~Zhao, L.~Zhang, S.~Huang, Y.~Shen, and S.~Zhao, ``Dehazing evaluation:
  Real-world benchmark datasets, criteria and baselines,'' \emph{IEEE
  Transactions on Image Processing}, 2020.

\bibitem{zhang2017fast}
J.~Zhang, Y.~Cao, S.~Fang, Y.~Kang, and C.~Wen~Chen, ``Fast haze removal for
  nighttime image using maximum reflectance prior,'' in \emph{Proceedings of
  the IEEE Conference on Computer Vision and Pattern Recognition}, 2017, pp.
  7418--7426.

\bibitem{li2015nighttime}
Y.~Li, R.~T. Tan, and M.~S. Brown, ``Nighttime haze removal with glow and
  multiple light colors,'' in \emph{Proceedings of the IEEE International
  Conference on Computer Vision}, 2015, pp. 226--234.

\bibitem{zhang2020nighttime}
J.~Zhang, Y.~Cao, Z.-J. Zha, and D.~Tao, ``Nighttime dehazing with a synthetic
  benchmark,'' in \emph{Proceedings of the 28th ACM International Conference on
  Multimedia}, 2020, pp. 2355--2363.

\bibitem{zhang2000flexible}
Z.~Zhang, ``A flexible new technique for camera calibration,'' \emph{IEEE
  Transactions on Pattern Analysis and Machine Intelligence}, vol.~22, no.~11,
  pp. 1330--1334, 2000.

\bibitem{xue2019learning}
Z.~Xue, N.~Xue, G.-S. Xia, and W.~Shen, ``Learning to calibrate straight lines
  for fisheye image rectification,'' in \emph{Proceedings of the IEEE
  Conference on Computer Vision and Pattern Recognition}, 2019, pp. 1643--1651.

\bibitem{liu2019online}
X.~Liu, L.~Zhang, Y.~Shen, S.~Zhang, and S.~Zhao, ``Online camera pose
  optimization for the surround-view system,'' in \emph{Proceedings of the 27th
  ACM International Conference on Multimedia}, 2019, pp. 383--391.

\bibitem{bahl1986maximum}
L.~Bahl, P.~Brown, P.~De~Souza, and R.~Mercer, ``Maximum mutual information
  estimation of hidden markov model parameters for speech recognition,'' in
  \emph{Proceedings of the IEEE International Conference on Acoustics, Speech,
  and Signal Processing}, vol.~11, 1986, pp. 49--52.

\bibitem{dong2018speech}
L.~Dong, S.~Xu, and B.~Xu, ``Speech-transformer: A no-recurrence
  sequence-to-sequence model for speech recognition,'' in \emph{Proceedings of
  the IEEE International Conference on Acoustics, Speech and Signal
  Processing}, 2018, pp. 5884--5888.

\bibitem{chang2011csr}
Y.-S. Chang, S.-H. Hung, N.~J. Wang, and B.-S. Lin, ``Csr: A cloud-assisted
  speech recognition service for personal mobile device,'' in \emph{Proceedings
  of the International Conference on Parallel Processing}, 2011, pp. 305--314.

\bibitem{he2019streaming}
Y.~He, T.~N. Sainath, R.~Prabhavalkar, I.~McGraw, R.~Alvarez, D.~Zhao,
  D.~Rybach, A.~Kannan, Y.~Wu, R.~Pang \emph{et~al.}, ``Streaming end-to-end
  speech recognition for mobile devices,'' in \emph{Proceedings of the IEEE
  International Conference on Acoustics, Speech and Signal Processing}, 2019,
  pp. 6381--6385.

\bibitem{longo2019ubiquitous}
F.~Longo, L.~Nicoletti, and A.~Padovano, ``Ubiquitous knowledge empowers the
  smart factory: The impacts of a service-oriented digital twin on enterprises'
  performance,'' \emph{Annual Reviews in Control}, vol.~47, pp. 221--236, 2019.

\bibitem{owens2010road}
J.~M. Owens, S.~B. McLaughlin, and J.~Sudweeks, ``On-road comparison of driving
  performance measures when using handheld and voice-control interfaces for
  mobile phones and portable music players,'' \emph{SAE International Journal
  of Passenger Cars-Mechanical Systems}, vol.~3, no. 2010-01-1036, pp.
  734--743, 2010.

\bibitem{tashev2009commute}
I.~Tashev, M.~Seltzer, Y.-C. Ju, Y.-Y. Wang, and A.~Acero, ``Commute ux: Voice
  enabled in-car infotainment system,'' Microsoft Research, Tech. Rep., 2009.

\bibitem{angelov2014speech}
K.~Angelov, B.~Bringert, and A.~Ranta, ``Speech-enabled hybrid multilingual
  translation for mobile devices,'' in \emph{Proceedings of the Demonstrations
  at the 14th Conference of the European Chapter of the Association for
  Computational Linguistics}, 2014, pp. 41--44.

\bibitem{campbell1997speaker}
J.~P. Campbell, ``Speaker recognition: A tutorial,'' \emph{Proceedings of the
  IEEE}, vol.~85, no.~9, pp. 1437--1462, 1997.

\bibitem{dehak2010front}
N.~Dehak, P.~J. Kenny, R.~Dehak, P.~Dumouchel, and P.~Ouellet, ``Front-end
  factor analysis for speaker verification,'' \emph{IEEE Transactions on Audio,
  Speech, and Language Processing}, vol.~19, no.~4, pp. 788--798, 2010.

\bibitem{chung2018voxceleb2}
J.~S. Chung, A.~Nagrani, and A.~Zisserman, ``Voxceleb2: Deep speaker
  recognition,'' \emph{Proc. Interspeech 2018}, pp. 1086--1090, 2018.

\bibitem{feng2017continuous}
H.~Feng, K.~Fawaz, and K.~G. Shin, ``Continuous authentication for voice
  assistants,'' in \emph{Proceedings of the 23rd Annual International
  Conference on Mobile Computing and Networking}, 2017, pp. 343--355.

\bibitem{ross2020security}
A.~Ross, S.~Banerjee, and A.~Chowdhury, ``Security in smart cities: A brief
  review of digital forensic schemes for biometric data,'' \emph{Pattern
  Recognition Letters}, 2020.

\bibitem{cho2014properties}
K.~Cho, B.~van Merrienboer, D.~Bahdanau, and Y.~Bengio, ``On the properties of
  neural machine translation: Encoder-decoder approaches,'' in
  \emph{Proceedings of the 8th Workshop on Syntax, Semantics and Structure in
  Statistical Translation}, 2014.

\bibitem{bahdanau2015neural}
D.~Bahdanau, K.~Cho, and Y.~Bengio, ``Neural machine translation by jointly
  learning to align and translate,'' in \emph{Proceedings of the 3rd
  International Conference on Learning Representations}, 2015.

\bibitem{wu2016google}
Y.~Wu, M.~Schuster, Z.~Chen, Q.~V. Le, M.~Norouzi, W.~Macherey, M.~Krikun,
  Y.~Cao, Q.~Gao, K.~Macherey \emph{et~al.}, ``Google's neural machine
  translation system: Bridging the gap between human and machine translation,''
  \emph{arXiv preprint arXiv:1609.08144}, 2016.

\bibitem{vaswani2017attention}
A.~Vaswani, N.~Shazeer, N.~Parmar, J.~Uszkoreit, L.~Jones, A.~N. Gomez,
  {\L}.~Kaiser, and I.~Polosukhin, ``Attention is all you need,'' in
  \emph{Advances in Neural Information Processing Systems}, 2017, pp.
  5998--6008.

\bibitem{zhu2019incorporating}
J.~Zhu, Y.~Xia, L.~Wu, D.~He, T.~Qin, W.~Zhou, H.~Li, and T.~Liu,
  ``Incorporating bert into neural machine translation,'' in \emph{Proceedings
  of the International Conference on Learning Representations}, 2019.

\bibitem{wang2017adversarial}
B.~Wang, Y.~Yang, X.~Xu, A.~Hanjalic, and H.~T. Shen, ``Adversarial cross-modal
  retrieval,'' in \emph{Proceedings of the 25th ACM International Conference on
  Multimedia}, 2017, pp. 154--162.

\bibitem{guo2019learning}
W.~Guo, H.~Huang, X.~Kong, and R.~He, ``Learning disentangled representation
  for cross-modal retrieval with deep mutual information estimation,'' in
  \emph{Proceedings of the 27th ACM International Conference on Multimedia},
  2019, pp. 1712--1720.

\bibitem{hossain2019comprehensive}
M.~Z. Hossain, F.~Sohel, M.~F. Shiratuddin, and H.~Laga, ``A comprehensive
  survey of deep learning for image captioning,'' \emph{ACM Computing Surveys},
  vol.~51, no.~6, pp. 1--36, 2019.

\bibitem{qiao2019mirrorgan}
T.~Qiao, J.~Zhang, D.~Xu, and D.~Tao, ``Mirrorgan: Learning text-to-image
  generation by redescription,'' in \emph{Proceedings of the IEEE Conference on
  Computer Vision and Pattern Recognition}, 2019, pp. 1505--1514.

\bibitem{li2017traffic}
L.~Li, B.~Qian, J.~Lian, W.~Zheng, and Y.~Zhou, ``Traffic scene segmentation
  based on rgb-d image and deep learning,'' \emph{IEEE Transactions on
  Intelligent Transportation Systems}, vol.~19, no.~5, pp. 1664--1669, 2017.

\bibitem{zhang2019rgb}
Q.~Zhang, N.~Huang, L.~Yao, D.~Zhang, C.~Shan, and J.~Han, ``Rgb-t salient
  object detection via fusing multi-level cnn features,'' \emph{IEEE
  Transactions on Image Processing}, vol.~29, pp. 3321--3335, 2019.

\bibitem{wang2020cross}
C.~Wang, C.~Xu, Z.~Cui, L.~Zhou, T.~Zhang, X.~Zhang, and J.~Yang, ``Cross-modal
  pattern-propagation for rgb-t tracking,'' in \emph{Proceedings of the
  IEEE/CVF Conference on Computer Vision and Pattern Recognition}, 2020, pp.
  7064--7073.

\bibitem{liu2018multi}
J.~Liu, Y.~Li, S.~Song, J.~Xing, C.~Lan, and W.~Zeng, ``Multi-modality
  multi-task recurrent neural network for online action detection,'' \emph{IEEE
  Transactions on Circuits and Systems for Video Technology}, vol.~29, no.~9,
  pp. 2667--2682, 2018.

\bibitem{wang2015large}
A.~Wang, J.~Lu, J.~Cai, T.-J. Cham, and G.~Wang, ``Large-margin multi-modal
  deep learning for rgb-d object recognition,'' \emph{IEEE Transactions on
  Multimedia}, vol.~17, no.~11, pp. 1887--1898, 2015.

\bibitem{xu2020outdoor}
Z.~Xu, S.~Liu, J.~Shi, and C.~Lu, ``Outdoor rgbd instance segmentation with
  residual regretting learning,'' \emph{IEEE Transactions on Image Processing},
  vol.~29, pp. 5301--5309, 2020.

\bibitem{liu2020deep}
D.~Liu, Y.~Li, J.~Lin, H.~Li, and F.~Wu, ``Deep learning-based video coding: A
  review and a case study,'' \emph{ACM Computing Surveys}, vol.~53, no.~1, pp.
  1--35, 2020.

\bibitem{duan2020video}
L.-Y. Duan, J.~Liu, W.~Yang, T.~Huang, and W.~Gao, ``Video coding for machines:
  A paradigm of collaborative compression and intelligent analytics,''
  \emph{arXiv preprint arXiv:2001.03569}, 2020.

\bibitem{he2017channel}
Y.~He, X.~Zhang, and J.~Sun, ``Channel pruning for accelerating very deep
  neural networks,'' in \emph{Proceedings of the IEEE International Conference
  on Computer Vision}, 2017, pp. 1389--1397.

\bibitem{cai2020rethinking}
Z.~Cai and N.~Vasconcelos, ``Rethinking differentiable search for
  mixed-precision neural networks,'' in \emph{Proceedings of the IEEE
  Conference on Computer Vision and Pattern Recognition}, 2020, pp. 2349--2358.

\bibitem{cheng2018model}
Y.~Cheng, D.~Wang, P.~Zhou, and T.~Zhang, ``Model compression and acceleration
  for deep neural networks: The principles, progress, and challenges,''
  \emph{IEEE Signal Processing Magazine}, vol.~35, no.~1, pp. 126--136, 2018.

\bibitem{elsken2019neural}
T.~Elsken, J.~H. Metzen, and F.~Hutter, ``Neural architecture search: A
  survey,'' \emph{Journal of Machine Learning Research}, vol.~20, pp. 1--21,
  2019.

\bibitem{donahue2019large}
J.~Donahue and K.~Simonyan, ``Large scale adversarial representation
  learning,'' in \emph{Advances in Neural Information Processing Systems},
  2019, pp. 10\,542--10\,552.

\bibitem{jing2020self}
L.~Jing and Y.~Tian, ``Self-supervised visual feature learning with deep neural
  networks: A survey,'' \emph{IEEE Transactions on Pattern Analysis and Machine
  Intelligence}, 2020.

\bibitem{van2020survey}
J.~E. Van~Engelen and H.~H. Hoos, ``A survey on semi-supervised learning,''
  \emph{Machine Learning}, vol. 109, no.~2, pp. 373--440, 2020.

\bibitem{zhang2020detecting}
Y.~Zhang, J.~Wang, and B.~Chen, ``Detecting false data injection attacks in
  smart grids: A semi-supervised deep learning approach,'' \emph{IEEE
  Transactions on Smart Grid}, 2020.

\bibitem{saeed2020federated}
A.~Saeed, F.~D. Salim, T.~Ozcelebi, and J.~Lukkien, ``Federated self-supervised
  learning of multi-sensor representations for embedded intelligence,''
  \emph{IEEE Internet of Things Journal}, 2020.

\bibitem{zhuang2020comprehensive}
F.~Zhuang, Z.~Qi, K.~Duan, D.~Xi, Y.~Zhu, H.~Zhu, H.~Xiong, and Q.~He, ``A
  comprehensive survey on transfer learning,'' \emph{Proceedings of the IEEE},
  2020.

\bibitem{liang2019federated}
X.~Liang, Y.~Liu, T.~Chen, M.~Liu, and Q.~Yang, ``Federated transfer
  reinforcement learning for autonomous driving,'' \emph{arXiv preprint
  arXiv:1910.06001}, 2019.

\bibitem{d2019transfer}
M.~D’Incecco, S.~Squartini, and M.~Zhong, ``Transfer learning for
  non-intrusive load monitoring,'' \emph{IEEE Transactions on Smart Grid},
  vol.~11, no.~2, pp. 1419--1429, 2019.

\bibitem{bargoti2017deep}
S.~Bargoti and J.~Underwood, ``Deep fruit detection in orchards,'' in
  \emph{Proceedings of the IEEE International Conference on Robotics and
  Automation}, 2017, pp. 3626--3633.

\bibitem{chen2017counting}
S.~W. Chen, S.~S. Shivakumar, S.~Dcunha, J.~Das, E.~Okon, C.~Qu, C.~J. Taylor,
  and V.~Kumar, ``Counting apples and oranges with deep learning: A data-driven
  approach,'' \emph{IEEE Robotics and Automation Letters}, vol.~2, no.~2, pp.
  781--788, 2017.

\bibitem{pan2017virtual}
X.~Pan, Y.~You, Z.~Wang, and C.~Lu, ``Virtual to real reinforcement learning
  for autonomous driving,'' \emph{arXiv preprint arXiv:1704.03952}, 2017.

\bibitem{bousmalis2018using}
K.~Bousmalis, A.~Irpan, P.~Wohlhart, Y.~Bai, M.~Kelcey, M.~Kalakrishnan,
  L.~Downs, J.~Ibarz, P.~Pastor, K.~Konolige \emph{et~al.}, ``Using simulation
  and domain adaptation to improve efficiency of deep robotic grasping,'' in
  \emph{Proceedings of the IEEE International Conference on Robotics and
  Automation}, 2018, pp. 4243--4250.

\bibitem{valerio2019leaf}
M.~Valerio~Giuffrida, A.~Dobrescu, P.~Doerner, and S.~A. Tsaftaris, ``Leaf
  counting without annotations using adversarial unsupervised domain
  adaptation,'' in \emph{Proceedings of the IEEE Conference on Computer Vision
  and Pattern Recognition Workshops}, 2019.

\bibitem{zhang2020cross}
Z.~Zhang, Y.~Wang, and S.~Liu, ``Cross-domain person re-identification using
  dual generation learning in camera sensor networks,'' \emph{Ad Hoc Networks},
  vol.~97, p. 102019, 2020.

\bibitem{passalis2020hypersphere}
N.~Passalis, A.~Iosifidis, M.~Gabbouj, and A.~Tefas, ``Hypersphere-based weight
  imprinting for few-shot learning on embedded devices,'' \emph{IEEE
  Transactions on Neural Networks and Learning Systems}, 2020.

\bibitem{feng2020fault}
L.~Feng and C.~Zhao, ``Fault description based attribute transfer for
  zero-sample industrial fault diagnosis,'' \emph{IEEE Transactions on
  Industrial Informatics}, 2020.

\bibitem{arulkumaran2017deep}
K.~Arulkumaran, M.~P. Deisenroth, M.~Brundage, and A.~A. Bharath, ``Deep
  reinforcement learning: A brief survey,'' \emph{IEEE Signal Processing
  Magazine}, vol.~34, no.~6, pp. 26--38, 2017.

\bibitem{mnih2015human}
V.~Mnih, K.~Kavukcuoglu, D.~Silver, A.~A. Rusu, J.~Veness, M.~G. Bellemare,
  A.~Graves, M.~Riedmiller, A.~K. Fidjeland, G.~Ostrovski \emph{et~al.},
  ``Human-level control through deep reinforcement learning,'' \emph{Nature},
  vol. 518, no. 7540, pp. 529--533, 2015.

\bibitem{mnih2016asynchronous}
V.~Mnih, A.~P. Badia, M.~Mirza, A.~Graves, T.~Lillicrap, T.~Harley, D.~Silver,
  and K.~Kavukcuoglu, ``Asynchronous methods for deep reinforcement learning,''
  in \emph{Proceedings of the International Conference on Machine Learning},
  2016, pp. 1928--1937.

\bibitem{ghesu2017multi}
F.-C. Ghesu, B.~Georgescu, Y.~Zheng, S.~Grbic, A.~Maier, J.~Hornegger, and
  D.~Comaniciu, ``Multi-scale deep reinforcement learning for real-time
  3d-landmark detection in ct scans,'' \emph{IEEE Transactions on Pattern
  Analysis and Machine Intelligence}, vol.~41, no.~1, pp. 176--189, 2017.

\bibitem{thananjeyan2017multilateral}
B.~Thananjeyan, A.~Garg, S.~Krishnan, C.~Chen, L.~Miller, and K.~Goldberg,
  ``Multilateral surgical pattern cutting in 2d orthotropic gauze with deep
  reinforcement learning policies for tensioning,'' in \emph{Proceedings of the
  IEEE International Conference on Robotics and Automation}, 2017, pp.
  2371--2378.

\bibitem{shiue2018real}
Y.-R. Shiue, K.-C. Lee, and C.-T. Su, ``Real-time scheduling for a smart
  factory using a reinforcement learning approach,'' \emph{Computers \&
  Industrial Engineering}, vol. 125, pp. 604--614, 2018.

\bibitem{wan2018model}
Z.~Wan, H.~Li, H.~He, and D.~Prokhorov, ``Model-free real-time ev charging
  scheduling based on deep reinforcement learning,'' \emph{IEEE Transactions on
  Smart Grid}, vol.~10, no.~5, pp. 5246--5257, 2018.

\bibitem{chung2020distributed}
H.-M. Chung, S.~Maharjan, Y.~Zhang, and F.~Eliassen, ``Distributed deep
  reinforcement learning for intelligent load scheduling in residential smart
  grid,'' \emph{IEEE Transactions on Industrial Informatics}, 2020.

\bibitem{somov2018pervasive}
A.~Somov, D.~Shadrin, I.~Fastovets, A.~Nikitin, S.~Matveev, O.~Hrinchuk
  \emph{et~al.}, ``Pervasive agriculture: Iot-enabled greenhouse for plant
  growth control,'' \emph{IEEE Pervasive Computing}, vol.~17, no.~4, pp.
  65--75, 2018.

\bibitem{binas2019reinforcement}
J.~Binas, L.~Luginbuehl, and Y.~Bengio, ``Reinforcement learning for
  sustainable agriculture,'' in \emph{Proceedings of the International
  Conference on Machine Learning Workshop}, 2019.

\bibitem{zhao2019routing}
L.~Zhao, J.~Wang, J.~Liu, and N.~Kato, ``Routing for crowd management in smart
  cities: A deep reinforcement learning perspective,'' \emph{IEEE
  Communications Magazine}, vol.~57, no.~4, pp. 88--93, 2019.

\bibitem{konevcny2016federated}
J.~Kone{\v{c}}n{\`y}, H.~B. McMahan, F.~X. Yu, P.~Richt{\'a}rik, A.~T. Suresh,
  and D.~Bacon, ``Federated learning: Strategies for improving communication
  efficiency,'' \emph{arXiv preprint arXiv:1610.05492}, 2016.

\bibitem{wu2020personalized}
Q.~Wu, K.~He, and X.~Chen, ``Personalized federated learning for intelligent
  iot applications: A cloud-edge based framework,'' \emph{IEEE Computer
  Graphics and Applications}, 2020.

\bibitem{brisimi2018federated}
T.~S. Brisimi, R.~Chen, T.~Mela, A.~Olshevsky, I.~C. Paschalidis, and W.~Shi,
  ``Federated learning of predictive models from federated electronic health
  records,'' \emph{International Journal of Medical Informatics}, vol. 112, pp.
  59--67, 2018.

\bibitem{wang2017knowledge}
Q.~Wang, Z.~Mao, B.~Wang, and L.~Guo, ``Knowledge graph embedding: A survey of
  approaches and applications,'' \emph{IEEE Transactions on Knowledge and Data
  Engineering}, vol.~29, no.~12, pp. 2724--2743, 2017.

\bibitem{wang2005building}
W.~Wang and T.~E. Daniels, ``Building evidence graphs for network forensics
  analysis,'' in \emph{Proceedings of the 21st Annual Computer Security
  Applications Conference}, 2005, pp. 11--pp.

\bibitem{chen2019agrikg}
Y.~Chen, J.~Kuang, D.~Cheng, J.~Zheng, M.~Gao, and A.~Zhou, ``Agrikg: an
  agricultural knowledge graph and its applications,'' in \emph{Proceedings of
  the International Conference on Database Systems for Advanced Applications},
  2019, pp. 533--537.

\bibitem{bendakir2006using}
N.~Bendakir and E.~A{\"\i}meur, ``Using association rules for course
  recommendation,'' in \emph{Proceedings of the AAAI Workshop on Educational
  Data Mining}, vol.~3, 2006, pp. 1--10.

\bibitem{banerjee2017generating}
A.~Banerjee, R.~Dalal, S.~Mittal, and K.~P. Joshi, ``Generating digital twin
  models using knowledge graphs for industrial production lines,'' \emph{UMBC
  Information Systems Department}, 2017.

\bibitem{wang2015knowledge}
L.~Wang, Q.~Chen, Z.~Gao, L.~Niu, Y.~Zhao, Z.~Ma, and D.~Wu, ``Knowledge
  representation and general petri net models for power grid fault diagnosis,''
  \emph{IET Generation, Transmission \& Distribution}, vol.~9, no.~9, pp.
  866--873, 2015.

\bibitem{guo2020survey}
R.~Guo, L.~Cheng, J.~Li, P.~R. Hahn, and H.~Liu, ``A survey of learning
  causality with data: Problems and methods,'' \emph{ACM Computing Surveys},
  vol.~53, no.~4, pp. 1--37, 2020.

\bibitem{jiang2016spatial}
H.~Jiang, X.~Dai, D.~W. Gao, J.~J. Zhang, Y.~Zhang, and E.~Muljadi,
  ``Spatial-temporal synchrophasor data characterization and analytics in smart
  grid fault detection, identification, and impact causal analysis,''
  \emph{IEEE Transactions on Smart Grid}, vol.~7, no.~5, pp. 2525--2536, 2016.

\bibitem{may2009driver}
J.~F. May and C.~L. Baldwin, ``Driver fatigue: The importance of identifying
  causal factors of fatigue when considering detection and countermeasure
  technologies,'' \emph{Transportation Research Part F: Traffic Psychology and
  Behaviour}, vol.~12, no.~3, pp. 218--224, 2009.

\bibitem{chattopadhyay2019neural}
A.~Chattopadhyay, P.~Manupriya, A.~Sarkar, and V.~N. Balasubramanian, ``Neural
  network attributions: A causal perspective,'' in \emph{Proceedings of the
  International Conference on Machine Learning}, 2019, pp. 981--990.

\bibitem{goyal2019counterfactual}
Y.~Goyal, Z.~Wu, J.~Ernst, D.~Batra, D.~Parikh, and S.~Lee, ``Counterfactual
  visual explanations,'' in \emph{Proceedings of the International Conference
  on Machine Learning}, 2019, pp. 2376--2384.

\bibitem{denzler2003information}
J.~Denzler, M.~Zobel, and H.~Niemann, ``Information theoretic focal length
  selection for real-time active 3-d object tracking,'' in \emph{Proceedings of
  the IEEE International Conference on Computer Vision}, vol.~1, 2003, pp.
  400--400.

\bibitem{sommerlade2008information}
E.~Sommerlade and I.~Reid, ``Information-theoretic active scene exploration,''
  in \emph{Proceedings of the IEEE Conference on Computer Vision and Pattern
  Recognition}, 2008, pp. 1--7.

\bibitem{zhang2014chinese}
X.~Zhang and M.~Lapata, ``Chinese poetry generation with recurrent neural
  networks,'' in \emph{Proceedings of the 2014 Conference on Empirical Methods
  in Natural Language Processing}, 2014, pp. 670--680.

\bibitem{qiao2019learn}
T.~Qiao, J.~Zhang, D.~Xu, and D.~Tao, ``Learn, imagine and create:
  Text-to-image generation from prior knowledge,'' in \emph{Advances in Neural
  Information Processing Systems}, 2019, pp. 887--897.

\bibitem{briot2017deep}
J.-P. Briot, G.~Hadjeres, and F.-D. Pachet, ``Deep learning techniques for
  music generation--a survey,'' \emph{arXiv preprint arXiv:1709.01620}, 2017.

\bibitem{tsarouchi2016human}
P.~Tsarouchi, S.~Makris, and G.~Chryssolouris, ``Human--robot interaction
  review and challenges on task planning and programming,'' \emph{International
  Journal of Computer Integrated Manufacturing}, vol.~29, no.~8, pp. 916--931,
  2016.

\bibitem{zhang2018deep}
T.~Zhang, Z.~McCarthy, O.~Jow, D.~Lee, X.~Chen, K.~Goldberg, and P.~Abbeel,
  ``Deep imitation learning for complex manipulation tasks from virtual reality
  teleoperation,'' in \emph{Proceedings of the IEEE International Conference on
  Robotics and Automation}, 2018, pp. 1--8.

\bibitem{yang2019privacy}
W.~Yang, S.~Wang, G.~Zheng, J.~Yang, and C.~Valli, ``A privacy-preserving
  lightweight biometric system for internet of things security,'' \emph{IEEE
  Communications Magazine}, vol.~57, no.~3, pp. 84--89, 2019.

\bibitem{bulan2017segmentation}
O.~Bulan, V.~Kozitsky, P.~Ramesh, and M.~Shreve, ``Segmentation-and
  annotation-free license plate recognition with deep localization and failure
  identification,'' \emph{IEEE Transactions on Intelligent Transportation
  Systems}, vol.~18, no.~9, pp. 2351--2363, 2017.

\bibitem{kumar2019intelligent}
P.~M. Kumar, U.~Gandhi, R.~Varatharajan, G.~Manogaran, R.~Jidhesh, and
  T.~Vadivel, ``Intelligent face recognition and navigation system using neural
  learning for smart security in internet of things,'' \emph{Cluster
  Computing}, vol.~22, no.~4, pp. 7733--7744, 2019.

\bibitem{ma2019trafficpredict}
Y.~Ma, X.~Zhu, S.~Zhang, R.~Yang, W.~Wang, and D.~Manocha, ``Trafficpredict:
  Trajectory prediction for heterogeneous traffic-agents,'' in
  \emph{Proceedings of the AAAI Conference on Artificial Intelligence},
  vol.~33, 2019, pp. 6120--6127.

\bibitem{rasouli2019pie}
A.~Rasouli, I.~Kotseruba, T.~Kunic, and J.~K. Tsotsos, ``Pie: A large-scale
  dataset and models for pedestrian intention estimation and trajectory
  prediction,'' in \emph{Proceedings of the IEEE International Conference on
  Computer Vision}, 2019, pp. 6262--6271.

\bibitem{he2020converting}
S.~He, M.~Han, N.~Patel, and Z.~Li, ``Converting handwritten text to editable
  format via gesture recognition for education,'' in \emph{Proceedings of the
  51st ACM Technical Symposium on Computer Science Education}, 2020, pp.
  1369--1369.

\bibitem{brule2016mapsense}
E.~Brule, G.~Bailly, A.~Brock, F.~Valentin, G.~Denis, and C.~Jouffrais,
  ``Mapsense: multi-sensory interactive maps for children living with visual
  impairments,'' in \emph{Proceedings of the 2016 CHI Conference on Human
  Factors in Computing Systems}, 2016, pp. 445--457.

\bibitem{hosseini2020intelligent}
M.~M. Hosseini, A.~Umunnakwe, M.~Parvania, and T.~Tasdizen, ``Intelligent
  damage classification and estimation in power distribution poles using
  unmanned aerial vehicles and convolutional neural networks,'' \emph{IEEE
  Transactions on Smart Grid}, 2020.

\bibitem{senocak2018learning}
A.~Senocak, T.-H. Oh, J.~Kim, M.-H. Yang, and I.~So~Kweon, ``Learning to
  localize sound source in visual scenes,'' in \emph{Proceedings of the IEEE
  Conference on Computer Vision and Pattern Recognition}, 2018, pp. 4358--4366.

\bibitem{chen2016detection}
K.~Chen, J.~Hu, and J.~He, ``Detection and classification of transmission line
  faults based on unsupervised feature learning and convolutional sparse
  autoencoder,'' \emph{IEEE Transactions on Smart Grid}, vol.~9, no.~3, pp.
  1748--1758, 2016.

\bibitem{chebrolu2018robust}
N.~Chebrolu, T.~L{\"a}be, and C.~Stachniss, ``Robust long-term registration of
  uav images of crop fields for precision agriculture,'' \emph{IEEE Robotics
  and Automation Letters}, vol.~3, no.~4, pp. 3097--3104, 2018.

\bibitem{jain2015head}
D.~Jain, L.~Findlater, J.~Gilkeson, B.~Holland, R.~Duraiswami, D.~Zotkin,
  C.~Vogler, and J.~E. Froehlich, ``Head-mounted display visualizations to
  support sound awareness for the deaf and hard of hearing,'' in
  \emph{Proceedings of the 33rd Annual ACM Conference on Human Factors in
  Computing Systems}, 2015, pp. 241--250.

\bibitem{mocanu2018line}
E.~Mocanu, D.~C. Mocanu, P.~H. Nguyen, A.~Liotta, M.~E. Webber, M.~Gibescu, and
  J.~G. Slootweg, ``On-line building energy optimization using deep
  reinforcement learning,'' \emph{IEEE Transactions on Smart Grid}, vol.~10,
  no.~4, pp. 3698--3708, 2018.

\bibitem{le2016graph}
D.~Le-Phuoc, H.~N.~M. Quoc, H.~N. Quoc, T.~T. Nhat, and M.~Hauswirth, ``The
  graph of things: A step towards the live knowledge graph of connected
  things,'' \emph{Journal of Web Semantics}, vol.~37, pp. 25--35, 2016.

\bibitem{pan2015internet}
J.~Pan, R.~Jain, S.~Paul, T.~Vu, A.~Saifullah, and M.~Sha, ``An internet of
  things framework for smart energy in buildings: designs, prototype, and
  experiments,'' \emph{IEEE Internet of Things Journal}, vol.~2, no.~6, pp.
  527--537, 2015.

\bibitem{jiang2017novel}
H.~Jiang, Z.~Zhang, J.~Dang, and L.~Wu, ``A novel 3-d massive mimo channel
  model for vehicle-to-vehicle communication environments,'' \emph{IEEE
  Transactions on Communications}, vol.~66, no.~1, pp. 79--90, 2017.

\bibitem{jiang2018joint}
D.~Jiang, L.~Huo, Z.~Lv, H.~Song, and W.~Qin, ``A joint multi-criteria
  utility-based network selection approach for vehicle-to-infrastructure
  networking,'' \emph{IEEE Transactions on Intelligent Transportation Systems},
  vol.~19, no.~10, pp. 3305--3319, 2018.

\bibitem{lee2018design}
C.~Lee, Y.~Lv, K.~Ng, W.~Ho, and K.~Choy, ``Design and application of internet
  of things-based warehouse management system for smart logistics,''
  \emph{International Journal of Production Research}, vol.~56, no.~8, pp.
  2753--2768, 2018.

\bibitem{zhu2016traffic}
Z.~Zhu, D.~Liang, S.~Zhang, X.~Huang, B.~Li, and S.~Hu, ``Traffic-sign
  detection and classification in the wild,'' in \emph{Proceedings of the IEEE
  Conference on Computer Vision and Pattern Recognition}, 2016, pp. 2110--2118.

\bibitem{chen2019shape}
Z.~Chen, W.~Ouyang, T.~Liu, and D.~Tao, ``A shape transformation-based dataset
  augmentation framework for pedestrian detection,'' \emph{arXiv preprint
  arXiv:1912.07010}, 2019.

\bibitem{hu2019joint}
H.-N. Hu, Q.-Z. Cai, D.~Wang, J.~Lin, M.~Sun, P.~Krahenbuhl, T.~Darrell, and
  F.~Yu, ``Joint monocular 3d vehicle detection and tracking,'' in
  \emph{Proceedings of the IEEE International Conference on Computer Vision},
  2019, pp. 5390--5399.

\bibitem{trung2016flexible}
T.~Q. Trung and N.-E. Lee, ``Flexible and stretchable physical sensor
  integrated platforms for wearable human-activity monitoring and personal
  healthcare,'' \emph{Advanced materials}, vol.~28, no.~22, pp. 4338--4372,
  2016.

\bibitem{yang2016super}
X.~Yang and Y.~Tian, ``Super normal vector for human activity recognition with
  depth cameras,'' \emph{IEEE Transactions on Pattern Analysis and Machine
  Intelligence}, vol.~39, no.~5, pp. 1028--1039, 2016.

\bibitem{gao2020real}
Y.~Gao, J.~Lin, J.~Xie, and Z.~Ning, ``A real-time defect detection method for
  digital signal processing of industrial inspection applications,'' \emph{IEEE
  Transactions on Industrial Informatics}, 2020.

\bibitem{he2020momentum}
K.~He, H.~Fan, Y.~Wu, S.~Xie, and R.~Girshick, ``Momentum contrast for
  unsupervised visual representation learning,'' in \emph{Proceedings of the
  IEEE Conference on Computer Vision and Pattern Recognition}, 2020, pp.
  9729--9738.

\bibitem{andrae2015global}
A.~S. Andrae and T.~Edler, ``On global electricity usage of communication
  technology: trends to 2030,'' \emph{Challenges}, vol.~6, no.~1, pp. 117--157,
  2015.

\bibitem{verma2009server}
A.~Verma, G.~Dasgupta, T.~K. Nayak, P.~De, and R.~Kothari, ``Server workload
  analysis for power minimization using consolidation,'' in \emph{Proceedings
  of the 2009 conference on USENIX Annual technical conference}, 2009, pp.
  28--28.

\bibitem{yuan2020biobjective}
H.~Yuan, J.~Bi, M.~Zhou, Q.~Liu, and A.~C. Ammari, ``Biobjective task
  scheduling for distributed green data centers,'' \emph{IEEE Transactions on
  Automation Science and Engineering}, 2020.

\bibitem{aono2017privacy}
Y.~Aono, T.~Hayashi, L.~Wang, S.~Moriai \emph{et~al.}, ``Privacy-preserving
  deep learning via additively homomorphic encryption,'' \emph{IEEE
  Transactions on Information Forensics and Security}, vol.~13, no.~5, pp.
  1333--1345, 2017.

\end{thebibliography}

%


\begin{IEEEbiography}[{\includegraphics[width=1in,height=1.25in,clip,keepaspectratio]{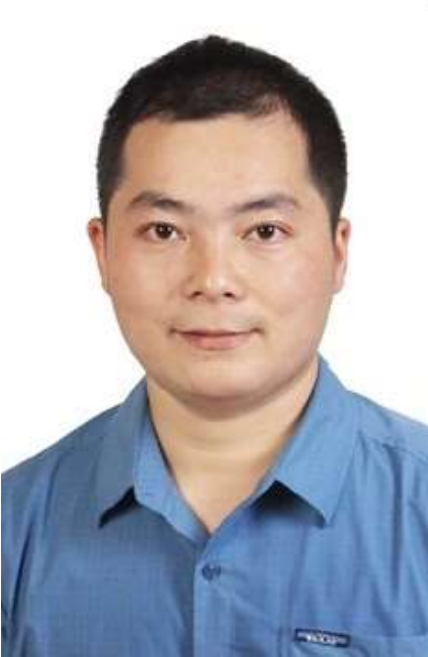}}]{Jing Zhang} is a research fellow of computer vision in the School of Computer Science and the Faculty of Engineering at the University of Sydney. His research interests include computer vision and deep learning. He has published several papers on top tier conferences and journals including CVPR, ICCV, NeurIPS, AAAI, Multimedia, SIGIR, IJCV, IEEE TIP, and TCSVT. He serves as a reviewer for a number of journals and conferences such as IEEE TIP, TCYB, TCSVT, IJCV, CVPR, ECCV, NeurIPS, ICLR, AAAI, IJCAI, and ACM Multimedia.\end{IEEEbiography}

\begin{IEEEbiography}[{\includegraphics[width=1in,height=1.25in,clip,keepaspectratio]{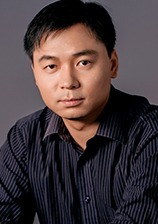}}]{Dacheng Tao} (F'15) is currently a Professor of Computer Science and an ARC Laureate Fellow in the School of Computer Science and the Faculty of Engineering at The University of Sydney. He mainly applies statistics and mathematics to artificial intelligence and data science, and his research is detailed in one monograph and over 200 publications in prestigious journals and proceedings at leading conferences. He received the 2015 Australian Scopus-Eureka Prize and the 2018 IEEE ICDM Research Contributions Award. He is a fellow of the Australian Academy of Science, AAAS, ACM and IEEE.\end{IEEEbiography}







\end{document}